%% file: main.tex
\documentclass[10pt,journal,compsoc]{IEEEtran}
\usepackage{graphics}
\usepackage{color}
\usepackage{mathtools}
\usepackage{multirow}
\usepackage{subfigure}
\usepackage{tabularx}
\usepackage{amsfonts}
\usepackage{wrapfig,lipsum,booktabs}
\usepackage{xcolor}
\usepackage{floatflt}
\usepackage{caption}
\usepackage{hyperref}

\usepackage[cal=cm]{mathalfa} 
\usepackage[ruled, lined, linesnumbered, commentsnumbered, longend]{algorithm2e}

\ifCLASSOPTIONcompsoc
  \usepackage[nocompress]{cite}
\else
  \usepackage{cite}
\fi
\ifCLASSINFOpdf
\else
\fi
\newcommand{\method}{UCD}
\begin{document}
\setcounter{page}{1}

%
\title{Uncertainty-aware Contrastive Distillation\\for Incremental Semantic Segmentation}
%
%
%
%

\author{Guanglei Yang,
        Enrico Fini,
        Dan Xu,
        Paolo Rota,
        Mingli Ding$^*$,
        Moin Nabi,\\
        Xavier Alameda-Pineda,~\IEEEmembership{Senior Member,~IEEE,}
        Elisa Ricci$^*$,~\IEEEmembership{Member,~IEEE}
\IEEEcompsocitemizethanks{
\IEEEcompsocthanksitem Guanglei Yang and Mingli Ding are with School of Instrument Science
and Engineering, Harbin Institute of Technology (HIT), Harbin, China. E-mail: \{yangguanglei,dingml\}@hit.edu.cn. $^*$Corresponding author.
\IEEEcompsocthanksitem Enrico Fini, Paolo Rota and Elisa Ricci are with the Department of Information
Engineering and Computer Science, University of Trento, Italy. E-mail: \{enrico.fini, paolo.rota, elisa.ricci\}@unitn.it.$^*$Corresponding author.
\IEEEcompsocthanksitem Dan Xu is with the Department of Computer Science and Engineering,
Hong Kong University of Science and Technology. E-mail: danxu@cse.ust.hk.
\IEEEcompsocthanksitem Xavier Alameda-Pineda is with the Perception Group, INRIA. E-mail:
xavier.alameda-pineda@inria.fr.
\IEEEcompsocthanksitem Moin Nabi is with SAP AI Research, Berlin, Germany. E-mail: m.nabi@sap.com.
\IEEEcompsocthanksitem Elisa Ricci is with Deep Visual Learning group at Fondazione Bruno
Kessler, Trento, Italy.
}
}


%
%

\markboth{Journal of \LaTeX\ Class Files,~Vol.~14, No.~8, August~2015}%
{Shell \MakeLowercase{\textit{et al.}}: Bare Demo of IEEEtran.cls for Computer Society Journals}
%



\IEEEtitleabstractindextext{%
\begin{abstract}
A fundamental and challenging problem in deep learning is catastrophic forgetting, \textit{i.e.} the tendency of neural networks to fail to preserve the knowledge acquired from old tasks when
learning new tasks. This problem has been widely investigated in the research community and several Incremental Learning (IL) approaches have been proposed in the past years. While earlier works in computer vision have mostly focused on image classification and object detection, more recently some IL approaches for semantic segmentation have been introduced. These previous works showed that, despite its simplicity, knowledge distillation can be effectively employed to alleviate catastrophic forgetting. In this paper, we follow this research direction and, inspired by recent literature on contrastive learning, we propose a novel distillation framework, Uncertainty-aware Contrastive Distillation (\method). In a nutshell, \method~is operated by introducing a novel distillation loss that takes into account all the images in a mini-batch, enforcing similarity between features associated to all the pixels from the same classes, and pulling apart those corresponding to pixels from different classes. In order to mitigate catastrophic forgetting, we contrast features of the new model with features extracted by a frozen model learned at the previous incremental step. Our experimental results demonstrate the advantage of the proposed distillation technique, which can be used in synergy with previous IL approaches, and leads to state-of-art performance on three commonly adopted benchmarks for incremental semantic segmentation.
The code is available at \url{https://github.com/ygjwd12345/UCD}.
\end{abstract}

}

\maketitle

\IEEEdisplaynontitleabstractindextext

%
\IEEEpeerreviewmaketitle

\input{intro}

\input{related}

\input{method_clean}

\input{experiments_clean}

\section{Conclusion}
We presented a new uncertainty-aware contrastive distillation framework for incremental semantic segmentation, leveraging from the complementarity of pixel-wise contrastive learning and uncertainty estimation. Using our approach, the model captures global semantic relationships between old and new features through a contrastive formulation, significantly mitigating catastrophic forgetting with respect to previous art. In order to perform feature contrastive learning on the old classes, we used pseudo-labels, which can be noisy. To address this issue we devised an uncertainty estimation mechanism that leverages the joint probability of the pixel pairs belonging to the same class. This makes the network focus on high-agreement pixel-pairs and suppress the low-entropy pixel-pairs, leading to more effective distillation. Our extensive experimental evaluation demonstrated outstanding performance of our method in several datasets (VOC 2012, ADE20K and cityscape) and settings (14 in total). Future works will focus on exploiting other continual learning approaches departing from traditional distillation schemes, as well as considering the segmentation problem in presence of domain shift \cite{mancini2018best}.


%


\ifCLASSOPTIONcompsoc
  \section*{Acknowledgments}
\else
  \section*{Acknowledgment}
\fi

The support provided by China Scholarship Council (CSC) during a visit of Guanglei Yang to University of Trento is acknowledged. This work has been also supported by Caritro Foundation.

\ifCLASSOPTIONcaptionsoff
  \newpage
\fi



\bibliographystyle{IEEEtran}
\bibliography{bibliography}

%

\begin{IEEEbiography}[{\includegraphics[width=1in,height=1.25in,clip,keepaspectratio]{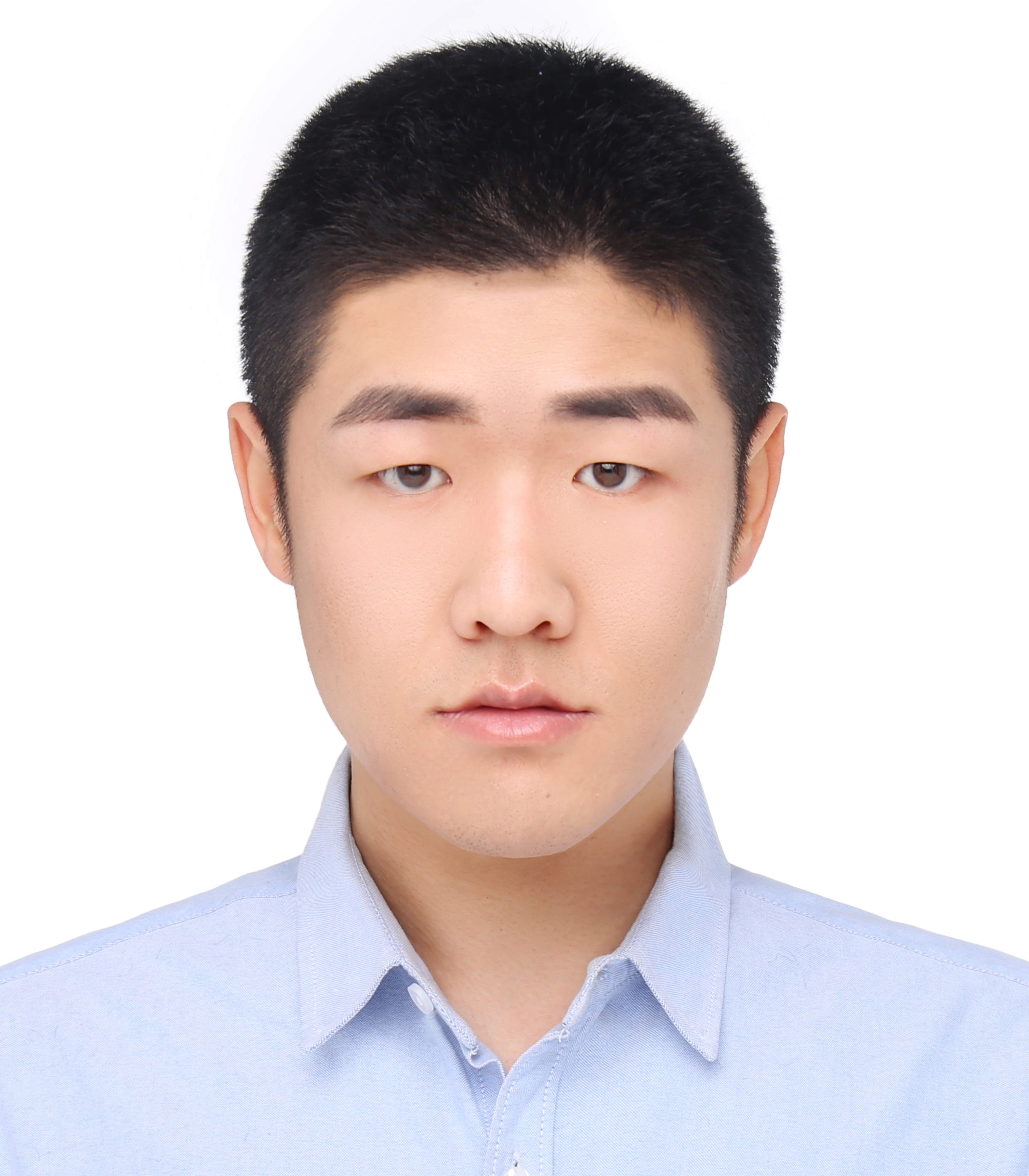}}]{Guanglei Yang}
received the B.S. degree in instrument science and technology from Harbin Institute of Technology (HIT), Harbin, China, in 2016. He is currently pursuing the Ph.D degree in the School of Instrumentation Science and Engineering, Harbin Institute of Technology(HIT), Harbin, China. He is working at University of Trento as a visiting student from 2020 to now. His research interests mainly include domain adaption, pixel-level prediction and attention gate.
\end{IEEEbiography}

\begin{IEEEbiography}[{\includegraphics[width=1in,height=1.25in,clip,keepaspectratio]{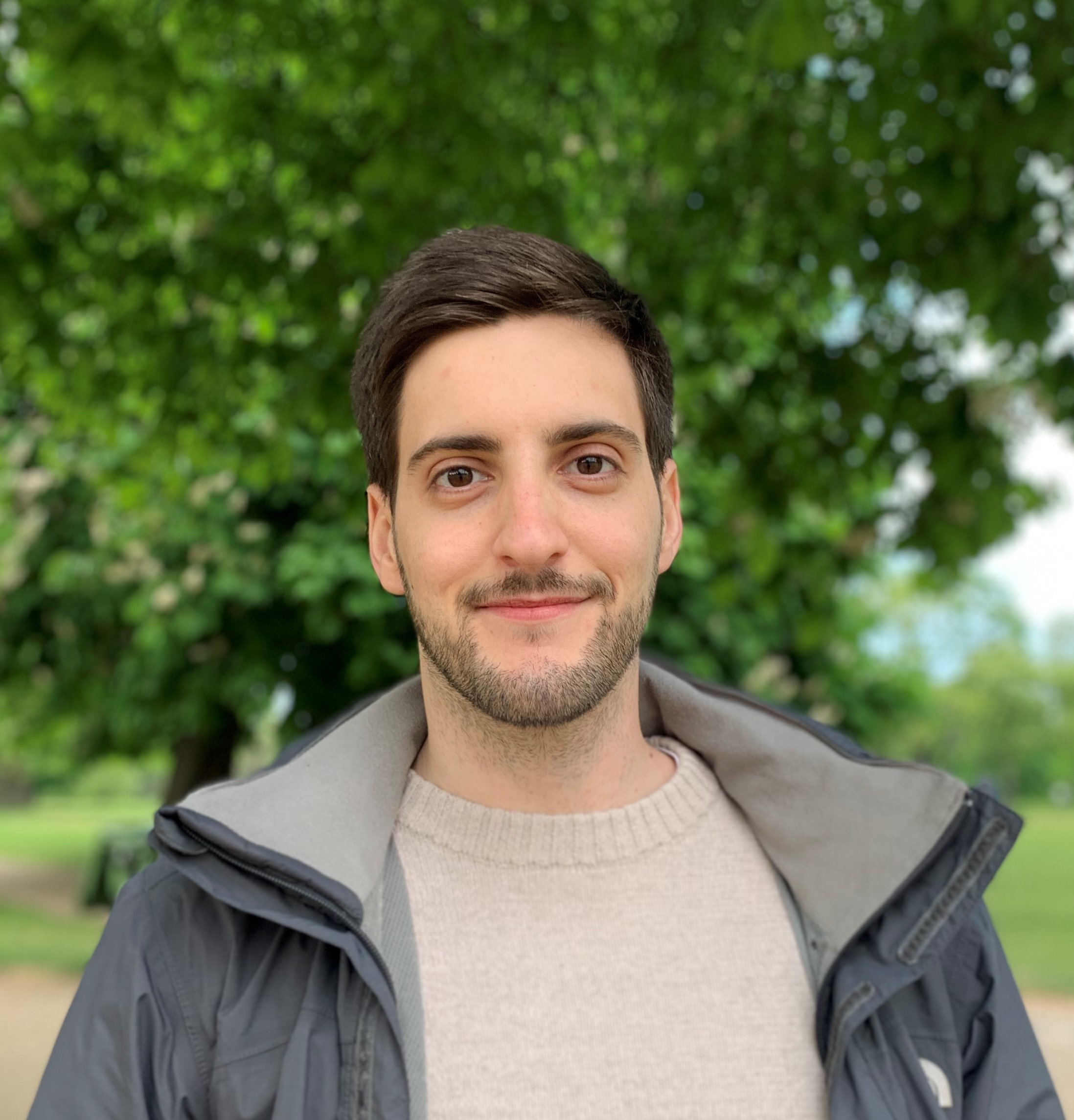}}]{Enrico Fini} is a Ph.D. student at the University of Trento. His research focuses on continual learning and self-supervised learning.
He received the B.S. degree in computer engineering from the University of Parma, Italy, in 2015 and the M.S. degree in computer science and engineering from Politecnico di Milano, Italy, in 2019. In 2018, He spent one year at the European Space Astronomy Centre of the European Space Agency in Madrid, Spain, working on machine learning for automatic sunspot detection.
\end{IEEEbiography}

\begin{IEEEbiography}[{\includegraphics[width=1in,height=1.25in,clip,keepaspectratio]{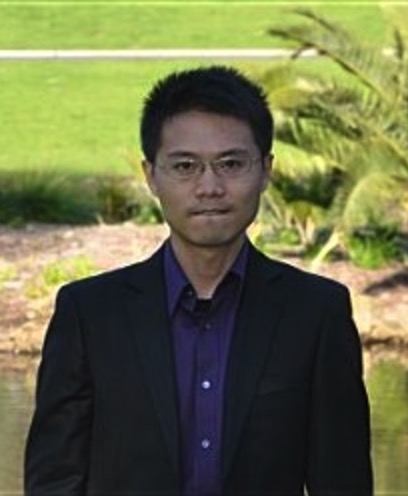}}]
{Dan Xu}
is an Assistant Professor in the Department of Computer Science and Engineering at HKUST. He was a Postdoctoral Research Fellow in VGG at the University of Oxford. He was a Ph.D. in the Department of Computer Science at the University of Trento. He was also a research assistant of MM Lab at the Chinese University of Hong Kong. He received the best scientific paper award at ICPR 2016, and a Best Paper Nominee at ACM MM 2018. He served as Area Chairs of ACM MM 2020, WACV 2021 and ICPR 2020.
\end{IEEEbiography}

\begin{IEEEbiography}[{\includegraphics[width=1in,height=1.25in,clip,keepaspectratio]{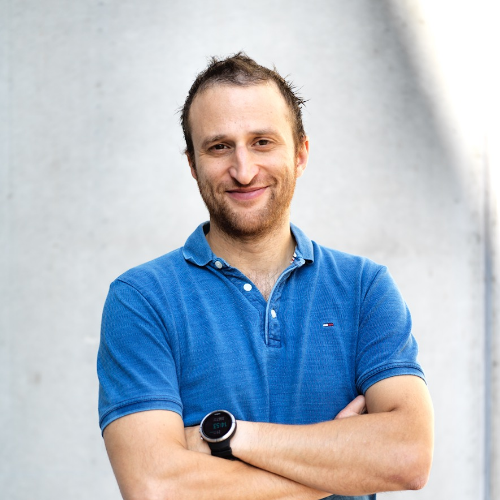}}]
{Paolo Rota}
is an assistant professor (RTDa) at University of Trento (in the MHUG group), working on computer vision and machine learning. He received his PhD in Information and Communication Technologies from the University of Trento in 2015. Prior joining UniTN he worked as Post-doc at the TU Wien and at the Italian Institute of Technology (IIT) of Genova. He is also collaborating with the ProM facility of Rovereto on assisting companies in inserting machine learning in their production chain.
\end{IEEEbiography}

\begin{IEEEbiography}[{\includegraphics[width=1in,height=1.25in,clip,keepaspectratio]{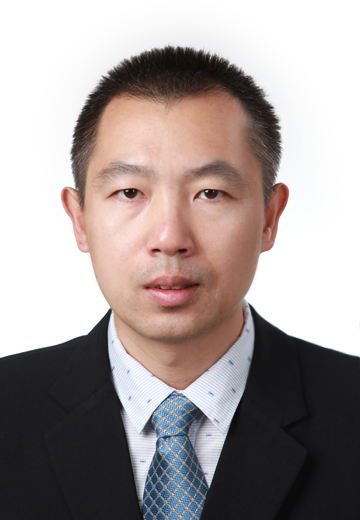}}]{Mingli Ding}
received the B.S., M.S. and Ph.D. degrees in instrument science and technology from Harbin Institute of Technology (HIT), Harbin, China, in 1996, 1997 and 2001, respectively. He worked as a visiting scholar in France from 2009 to 2010. Currently, he is a professor in the  School  of  Instrumentation Science and Engineering at Harbin Institute of Technology. Prof. Ding’s research interests are intelligence tests and information processing, automation test technology, computer vision, and machine learning. He has published over 40 papers in peer-reviewed journals and conferences.
\end{IEEEbiography}

\begin{IEEEbiography}[{\includegraphics[width=1in,height=1.25in,clip,keepaspectratio]{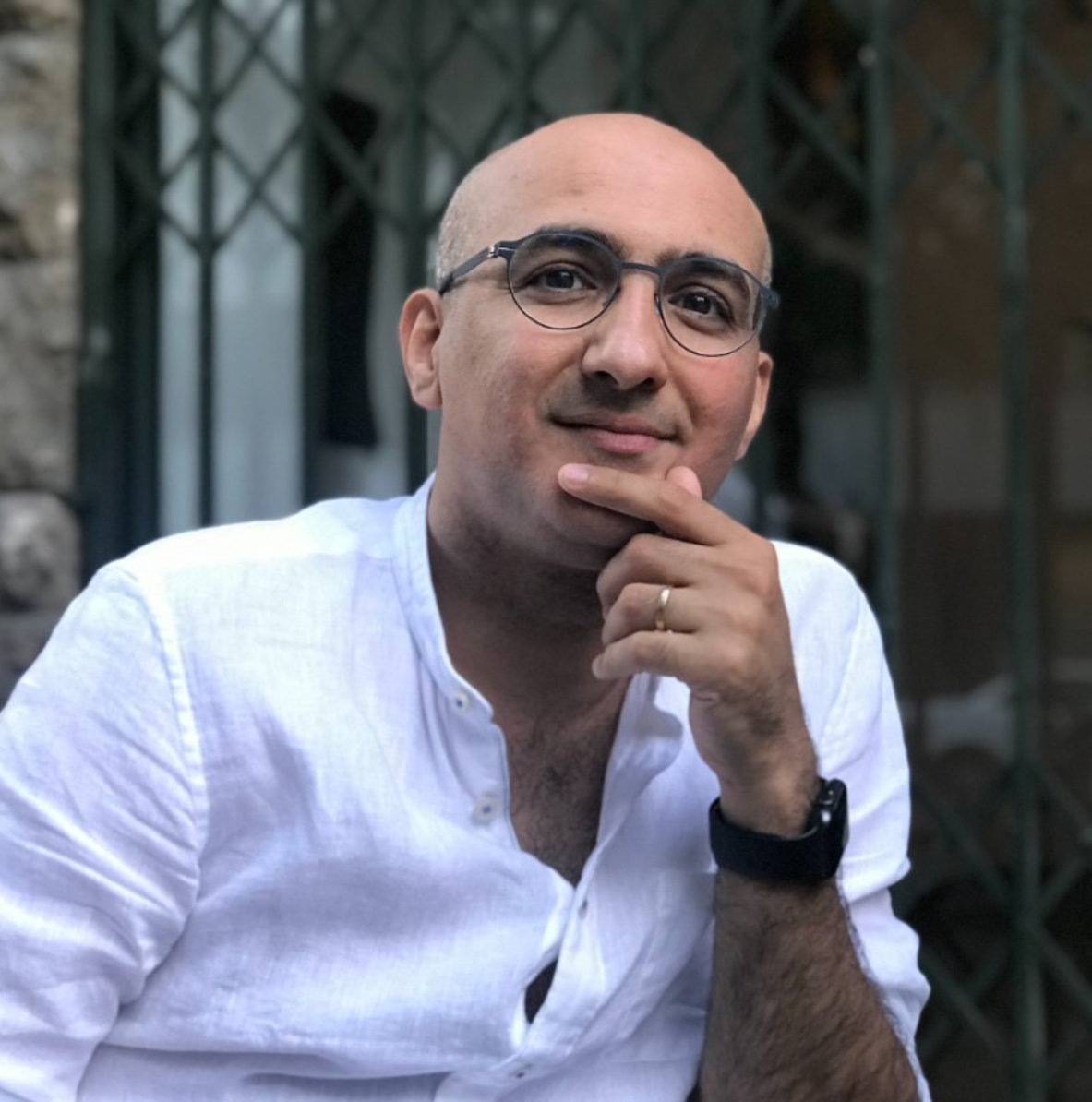}}]{Moin Nabi}
is a Principal Research Scientist at SAP AI Research in Berlin.  Before that, he was a post-doctoral research fellow at the University of Trento and a visiting researcher at the University of Washington.  He received his PhD degree from the Italian Institute of Technology in 2015. His  research  lies  at  the  intersection  of  machine  learning,  computer  vision,  and  natural  language processing with an emphasis on learning Deep Neural Networks with multimodal data.
\end{IEEEbiography}

\begin{IEEEbiography}[{\includegraphics[width=1in,height=1.25in,clip,keepaspectratio]{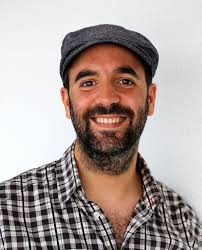}}]
{Xavier Alameda-Pineda} received M.Sc. degrees in mathematics (2008), in
telecommunications (2009) and in computer science
(2010) and a Ph.D. in mathematics and computer
science (2013) from Université Joseph Fourier. Since
2016, he is a Research Scientist at Inria Grenoble
Rhône-Alpes, with the Perception team. He served as
Area Chair at ICCV’17, of ICIAP’19 and of ACM
MM’19. He is the recipient of several paper awards
and of the ACM SIGMM Rising Star Award in 2018.
\end{IEEEbiography}

\begin{IEEEbiography}[{\includegraphics[width=1in,height=1.25in,clip,keepaspectratio]{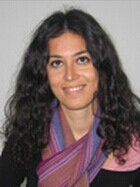}}]
{Elisa Ricci}
received the PhD degree from the
University of Perugia in 2008. She is an associate
professor at the University of Trento and a
researcher at Fondazione Bruno Kessler. She
has since been a post-doctoral researcher at
Idiap, Martigny, and Fondazione Bruno Kessler,
Trento. She was also a visiting researcher at the
University of Bristol. Her research interests are
mainly in the areas of computer vision and
machine learning. She is a member of the IEEE.
\end{IEEEbiography}




\end{document}

%% file: intro.tex
\IEEEraisesectionheading{\section{Introduction}\label{sec:introduction}}

%
%
%
%
\IEEEPARstart{A}{s} a longstanding, fundamental and challenging problem in computer vision, semantic segmentation, \textit{i.e.} the task of automatically assigning a class label to each pixel of an image, has been an active research area for several decades. In the recent years, we have witnessed substantial progress in this area \cite{long2015fully,chen2018encoder,lin2017refinenet,zhang2018exfuse,zhao2017pyramid,yu2018bisenet}, mostly thanks to the introduction of fully convolutional networks \cite{long2015fully} and to the integration of several other architectural components, such as attention models \cite{hu2018squeeze,chen2016attention,fu2019dual} or multi-scale fusion strategies \cite{chen2017deeplab,chen2017rethinking}.

\begin{figure}[t!]
    \centering
    \includegraphics[width=0.489\textwidth]{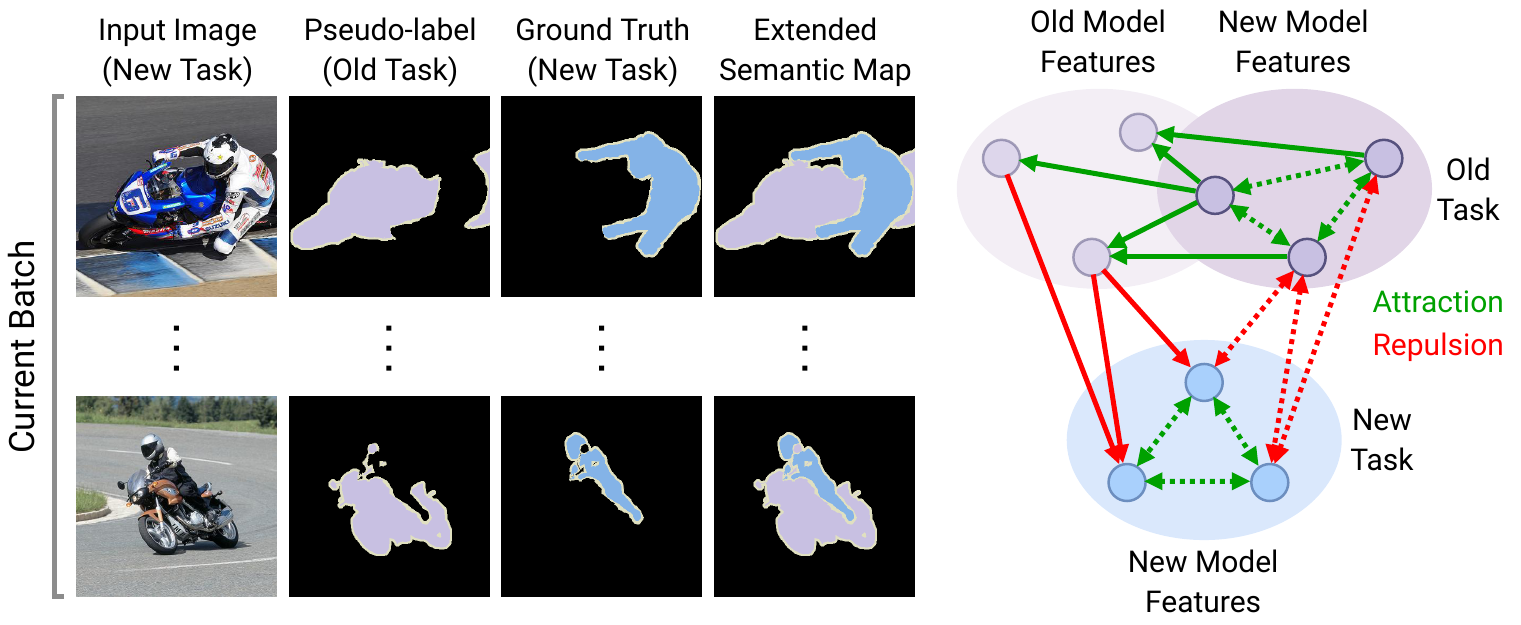}
    \caption{High-level view of our Uncertainty-aware Contrastive Distillation. (Left) Given a minibatch of data sampled from the new task we use the old model to generate pseudo-labels for the old classes. We then merge the pseudo-labels with the ground truth for the current task in order to obtain the extended semantic maps for all samples in the batch. (Right) Our contrastive distillation approach operates by contrasting the features of the new and old model based on the extended semantic map. Green arrows represent attraction, red arrows represent repulsion. The contrast between features of the new model with itself is depicted as a dashed arrow, while the contrast between new and old model features is depicted as a solid arrow.}
    \label{fig:teaser}
\end{figure}

Despite the astonishing performance, state of the art deep architectures for semantic segmentation still lack one fundamental capability: they cannot update their learned model continuously as new data are made available without losing previously learned knowledge.
This is a matter of utmost importance in several applications, such as autonomous driving or robot navigation.
The problem, also known as \textit{catastrophic forgetting} \cite{mccloskey1989catastrophic}, has been long studied
in the research community and several approaches, commonly referred as Incremental Learning (IL) methods \cite{de2019continual}, have been developed over the years. In computer vision, most research efforts in the area of IL have addressed problems such as image classification \cite{li2017learning,kirkpatrick2017overcoming,rebuffi2017icarl} and object detection~\cite{shmelkov2017incremental,liu2020multitask}. However, so far, much less attention has been devoted to incremental semantic segmentation.

Recently, due to the importance of it for several high level tasks, some approaches have proposed to continuously update a deep network for computing semantic segmentation maps \cite{michieli2019incremental,tasar2019incremental,ozdemir2018learn,ozdemir2019extending,cermelli2020modeling}. While having their own peculiarities, these methods develop from a common idea: overcoming catastrophic forgetting by resorting on knowledge distillation \cite{hinton2015distilling}, that is aiming to preserve the output of a complex network when adopting a smaller network. This idea can be transferred to IL~\cite{li2017learning} by enforcing that the predictions of the network on the old tasks do not fluctuate much whilst updating its parameters with samples of the new task. Despite their simplicity, distillation-based techniques \cite{michieli2019incremental,cermelli2020modeling} have proved effective for alleviating catastrophic forgetting, even in the challenging task of semantic segmentation. 

One prominent limitation of previous distillation-based techniques for semantic segmentation \cite{michieli2019incremental,cermelli2020modeling} is that distillation is performed on a pixel-basis, \textit{i.e.} the output of the new model for a given pixel is only compared with the output of the old model on the same pixel. This formulation is derived as a trivial extension of existing IL methods for image classification to segmentation. However, with this design choice, the crucial role of \textit{context} and the \textit{structured prediction} nature of the task is severely overlooked. Moreover, distilling on a pixel-basis does not explicitly take into account the relationships between representations of pixels belonging to different images. In this paper we propose a novel approach that allows to perform knowledge distillation for incremental semantic segmentation while contextualising the learned representation of a given pixel w.r.t.\ each and every single pixel in a minibatch. Inspired from recent literature on supervised contrastive learning~\cite{khosla2020supervised}, we introduce a constrastive distillation loss that enforces similarity on the representations of the pixels of the same class at batch level. Simultaneously, the contrastive distillation loss pushes pixels of different classes far apart. In other words, our method brings contrastive learning into distillation by considering all pixels of all images of a mini batch as samples to be pulled together or pushed apart depending on their class membership. 

The key contribution of the present manuscript is to tie together the advantages of knowledge distillation with those of contrastive learning. 
On the one side, we distill knowledge from the previous task by pulling together the representations of the current encoder to those of the past encoder for pixels belonging to classes of the previous task. 
To do so, we leverage the classifier of the previous task to provide pseudo-labels to the background of the current task, enriching the information content of the segmentation map of this latter task. 
On the other side, we use this enhanced segmentation map to contrast pixel-level representations of the new and old representations. In more detail, the new encoder is trained to pull together (push apart) pixel representations of the same class (different classes), whether these representations are extracted with the new or with the old task encoder.\footnote{Obviously the old encoder is not trained during the current task.} Additionally, we exploit the noisy nature of the pseudo-labels by modeling their uncertainty. More precisely, we re-weight the loss with the probability of two pixels belonging to the same class (whatever this class is).
We name our method Uncertainty-aware Constrastive Distillation (\method{}), and illustrate the core idea of it in Fig.~\ref{fig:teaser}.



In summary, our contributions are the following:
\begin{itemize}
    \item We introduce a novel contrastive distillation loss that accounts for semantic association among pixels of the new and old model.
    \item We introduce an uncertainty estimation strategy for our contrastive learning framework that leverages the joint probability of the pixels pairs belonging to the same class and weights the strength of the distillation signal accordingly.
    \item We conduct an extensive evaluation on three publicly available benchmarks, \textit{i.e.} Pascal-VOC2012~\cite{everingham2010pascal}, ADE20K~\cite{zhou2017scene} and Cityscapes~\cite{cordts2016cityscapes}, demonstrating that our approach outperforms state-of-the-art IL methods for semantic segmentation.
    \item We demonstrate that \method~can be integrated effortlessly on top of other state of the art methods, MiB~\cite{cermelli2020modeling} and PLOP~\cite{douillard2020plop}, boosting their performance significantly on all benchmarks.
\end{itemize}




%% file: related.tex
\section{Related Works}
\label{sec:related}

In this section, we review previous works on IL, specifically focusing on research studies addressing the catastrophic forgetting issue for semantic segmentation. As our paper also introduces a novel contrastive distillation strategy, we also review recent works on contrastive learning and knowledge distillation.

\noindent \textbf{Incremental Learning.} 
Enabling neural networks to learn continually is currently one of the most challenging and fascinating problems in computer vision. 
In the specific case of class incremental learning, the challenge is to train a neural network on subsequent tasks, each of them including a different set of categories, while avoiding catastrophic forgetting~\cite{mccloskey1989catastrophic}. The task information is assumed to be available at training time but unavailable at test time. 
Existing works in the literature can be grouped into three main categories: rehearsal-based methods~\cite{rebuffi2017icarl,shin2017continual,hou2019learning,ostapenko2019learning,wu2018memory}, parameter isolation approaches~\cite{mallya2018packnet,mallya2018piggyback} and distillation-based techniques~\cite{li2017learning,castro2018end,fini2020online}. Methods of the first category leverage from data that have been retained from the previous task or generated during the second task for counteracting forgetting. Parameter isolation strategies exploit a reduced set of task-specific parameters to incorporate knowledge about specific tasks. Distillation-based methods use predictions of the model on old tasks to transfer the knowledge from the old to the new model.

Regarding the latter category, recently, advanced distillation techniques have been explored. For instance, Dhar \textit{et al.} \cite{dhar2019learning} proposed to penalize the changes in attention maps. Differently, Douillard \textit{et al.} \cite{douillard2020podnet} distilled the pooled statistics of feature maps throughout the network, while Liu \textit{et al.} \cite{liu2020multi} developed an attentive feature distillation method in the context of object detection.
{More recently, RECALL~\cite{maracani2021recall} employs a web crawler or a pre-trained conditional GAN to replay images for the past semantic classes.
However, none of these works use contrastive learning to perform distillation.
Unlike Douillard \textit{et al.}~\cite{douillard2020podnet} and RECALL~\cite{maracani2021recall}, which use pseudo labels generated by the old model directly to supervise training new model, we propose an extended semantic map to guide uncertainty-aware contrastive distillation.
}

{\noindent \textbf{Semantic Segmentation.} 
Deep learning has enabled great advancements in semantic segmentation~\cite{long2015fully,chen2016deeplab,yu2015multi}. Long~\textit{et al.}~\cite{long2015fully} were the first to introduce fully convolutional networks (FCNs) for semantic segmentation, achieving significant improvements over previous models. Dilated convolutions~\cite{chen2016deeplab,yu2015multi} were designed in order to increase the receptive field while learning deep representations, further boosting performance. Recently, the new state-of-the-art methods use different strategies to condition pixel-level annotations on their global context,~\textit{e.g.} attention module~\cite{fu2019dual} and multiple scales~\cite{yuan2020object,wang2020deep}. On the other hand, \cite{wang2021exploring} proposes a pixel-wise contrastive algorithm for semantic segmentation in the fully supervised setting.
Most of semantic segmentation methods work in the offline setting, \textit{i.e.} they assume that
all classes are available beforehand during the training phase.
To the best of our knowledge, the problem of incremental learning in semantic segmentation has been addressed only in~\cite{michieli2019incremental,wang2020intra,tasar2019incremental,ozdemir2019extending}. Ozdemir~\textit{et al.}~\cite{ozdemir2019extending} proposed an incremental learning approach for medical images, extending a standard image-level classification method to segmentation and devising a strategy to select relevant samples of old datasets for rehearsal. Tarasr~\textit{et al.}~\cite{tasar2019incremental} introduced a similar method for segmenting remote sensing data. Differently, Michieli~\cite{michieli2019incremental} considered incremental learning for semantic segmentation in a special setting where labels are provided step by step. Moreover, they require the novel classes to be never present as background in pixels of previous learning steps. These requirements strongly limit the applicability of their method. Here we follow a more principled formulation of the
ICL problem in semantic segmentation, like~\cite{cermelli2020modeling} .
}

\noindent \textbf{Incremental Learning for Semantic Segmentation.} {Despite the flourishing literature in IL, still very few works tackled the problem of semantic segmentation~\cite{michieli2019incremental,tasar2019incremental,ozdemir2018learn,ozdemir2019extending,feng2021continual,cermelli2020modeling,yan2021framework}. For instance, in~\cite{ozdemir2018learn} Ozdemir \textit{et al.} proposed a method to extract prototypical examples for the old categories to be used as a support while learning new categories. Michieli \textit{et al.}~\cite{michieli2019incremental,michieli2021knowledge} introduced different distillation-based techniques working both on the output logits and on intermediate features. }The works in~\cite{ozdemir2019extending} and~\cite{tasar2019incremental} also considered a distillation framework to transfer knowledge between tasks, introducing specific contributions associated to the considered problems in remote sensing and medical imaging. Cermelli \textit{et al.}~\cite{cermelli2020modeling} introduced a distillation-based method which specifically accounts for the semantic distribution shift of the background class. Very recently, Douillard \textit{et al.} \cite{douillard2020plop} proposed an alternative mechanism to mitigate the background distribution shift using pseudo-labels, together with multi-scale pooling distillation adapted from~\cite{douillard2020podnet}. Finally, Michieli \textit{et al.} \cite{michieli2021continual} proposed a framework based on three components: prototypes matching, feature sparsity and an attraction-repulsion mechanism at prototype-level.

In this work, we also resort to distillation to mitigate catastrophic forgetting. However, we introduce a novel contrastive distillation loss which successfully accounts for semantic dependencies among pixels and empirically demonstrate that our contribution provides significantly improved performances over state of the art methods. Importantly, our approach can be combined with existing techniques \cite{cermelli2020modeling,douillard2020plop}.

\noindent \textbf{Contrastive Learning.} Self-supervised learning has recently attracted considerable attention in the computer vision community for its ability to learn discriminative features using a contrastive objective~\cite{kang2019contrastive,wang2021exploring}. This training scheme has been successfully employed in different fields such as image and video classification~\cite{chen2020simple,khosla2020supervised,kim2020adversarial}, as well as natural language processing~\cite{qu2020coda,klein2020contrastive} and audio representation learning~\cite{saeed2020contrastive,xiong2020approximate,zhang2020contrastive}. Contrastive learning consists in forcing the network to learn feature representations by pushing apart different samples (negatives) or pulling close similar ones (positives). The notion of similar and dissimilar samples varies according to the task. For instance, in~\cite{chen2020simple} an image and a simple transformation of such image are considered positives, while in \cite{he2020momentum} positives and negatives are generated using a momentum encoder. Contrastive learning has also proved itself to be extremely competitive in supervised settings~\cite{khosla2020supervised}. In \cite{khosla2020supervised} the class labels are exploited to encourage proximity not only to transformed versions of the same images but also to different images of the same class, leading to competitive performance with respect to traditional cross-entropy based classification methods. Very recently, in the context of semantic segmentation, the advantage of contrastive pre-training was demonstrated in \cite{chen2020simple} and a pixel-wise contrastive loss has been proposed \cite{wang2021exploring} showing promising results.
{SDR~\cite{michieli2021continual} firstly uses contrastive learning for incremental semantic segmentation while it does not use the contrastive loss in~\cite{wang2021exploring}. There is two main difference between our method and SDR: (i) we compare feature vectors between themselves in the feature space, not with prototypes, (ii) we use an uncertainty estimation mechanism to guide contrastive learning.}

%% file: method_clean.tex
\section{Proposed Approach}
\label{sec:method}
In this section, we first provide a formalization of the problem of IL for semantic segmentation. Then, we describe the proposed Uncertainty-aware Contrastive Distillation (\method). 

\subsection{Problem Definition and Notation}
\label{sec:problem}

Let $\mathcal{I}$ denote the input image space, $I\in\mathcal{I}$ an image and $\mathcal{Y}$ the label space. In semantic segmentation, given $I$ we want to assign a label in $\mathcal{Y}$, representing its semantic category, to each pixel of $I$. Pixels that are not assigned to other categories are assigned to the background class $\textsc{B}$. The problem can be addressed within a supervised framework defining a training set $\mathcal{T}=\{I_n,M_n\}_{n=1}^{N}$ of pairs of images $I_n$ and associated segmentation maps $M_n\in\mathcal{Y}^{H\times W}$, where $H$ and $W$ denote the image height and width respectively.

In the general IL setting, training is realized over multiple learning steps, and this is also the case in IL for semantic segmentation. In the following $k$ will denote the index of the incremental learning step. Each step is associated to a different training set $\mathcal{T}^k=\{I^k_n,M^k_n\}_{n=1}^{N^k}$. At each learning step $k$ we are interested in learning a model $\Phi^k$ able to segment the classes of the label space of the $k$-th step, $\mathcal{Y}^k$. Importantly, the training set with $N^k$ image-segmentation pairs is available only during the training of the corresponding incremental step. Moreover, $\Phi^k$ should be trained to mitigate catastrophic forgetting, that is to be able to also recognise the semantic classes of the previous learning steps $\cup_{k'=1}^{k-1}\mathcal{Y}^{k'}$ {without having access to $\cup_{k'=1}^{k-1}(\mathcal{I}^{k'}, \mathcal{Y}^{k'})$}. Following \cite{cermelli2020modeling}, we explore two settings for the splitting the datasets into different tasks: (i) the overlapped setting, which only ensures label separation across tasks $\mathcal{Y}^k\cap\mathcal{Y}^{k'}=\emptyset, \forall k'\in\{1,..., k-1\}$; and (ii) the disjoint setting, where in addition to the label separation the same training image must not belong to more than one task, formally: $\forall I_n^k\in \mathcal{T}^k,  I_n^k\notin \cup_{k'=1}^{k-1}\mathcal{T}^{k'}$.

In this paper we assume that the model $\Phi^k$ is implemented by the composition of a generic backbone $\psi_{\theta^k}$ with parameters $\theta^k$ and a classifier $\phi_{\omega^k}$ with parameters $\omega^k$. The features extracted with the backbone are denoted by $f\in\mathbb{R}^{\frac{H}{16}\times \frac{W}{16}\times D}$, \textit{i.e.} $f^k=\psi_{\theta^k}(I)$, meaning that $f^k$ are extracted with the backbone at the $k$-th step.
The output segmentation map can be computed from our pixel-wise predictor $Y^k_p = \{  \arg\max (\operatorname{upsample}( \phi_{\omega^k}(f^k_p)[p,c]))\}_{p\in I}$, where $\phi_{\omega^k}(f^k_p)[p,c]$ is the probability for class $c$ in pixel $p$ (which stands for a tuple $(H, W)$ where $H$ and $W$ are the height and width of an image).

\begin{figure}[t]
    \centering
    \includegraphics[width=0.95\linewidth]{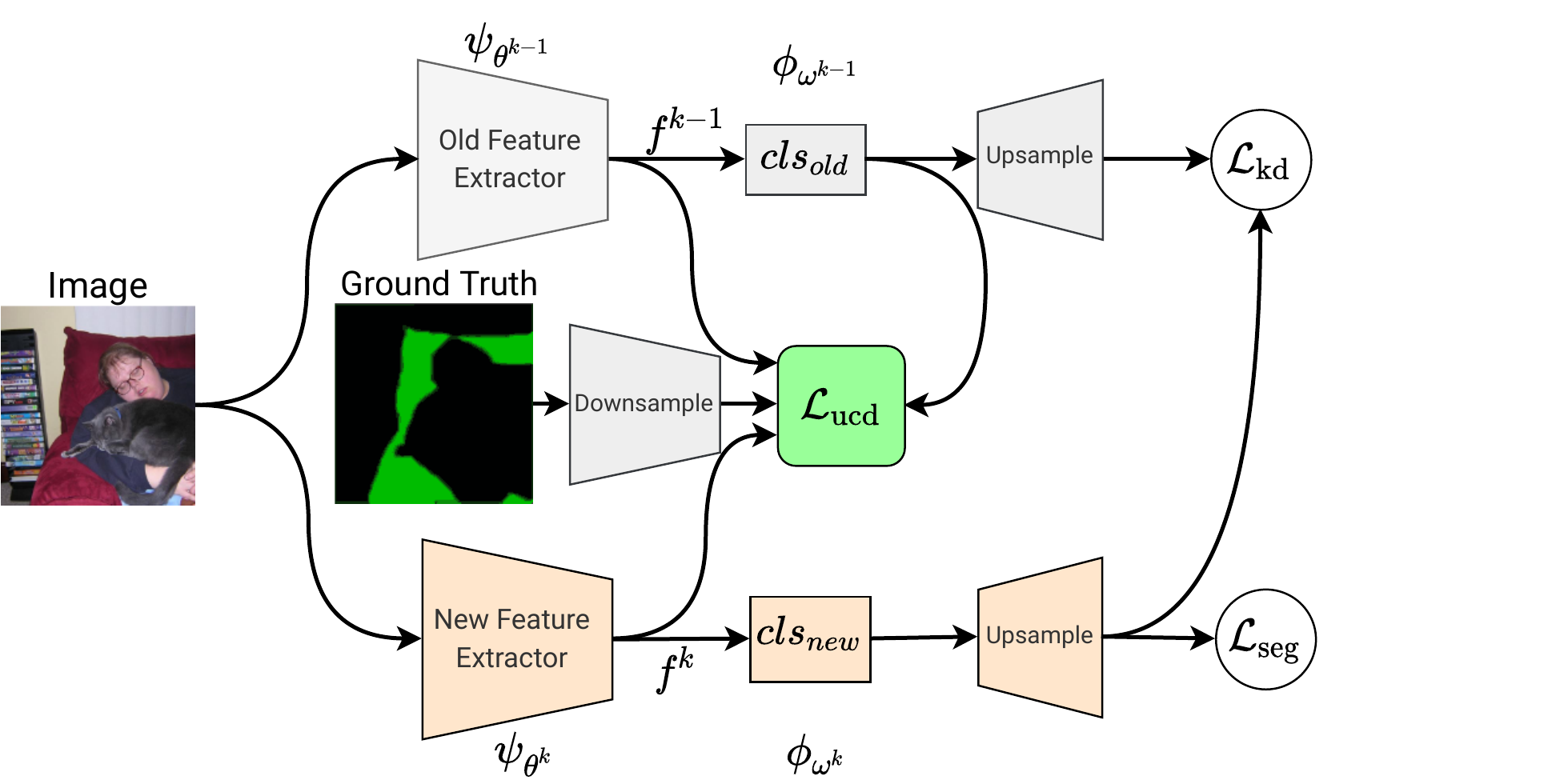}
    \caption{Overview of the proposed architecture at the incremental step $k$. The grey blocks denote the network trained on old tasks, while the light blue one indicates the network trained on old task.  The proposed network is made of a feature extractor and a classifier and it is trained on three losses. Our uncertainty-aware contrastive distillation framework corresponds to the green box. Downsample and Upsample denote the resolution decrease from $[{H},{W}]$ to $[\frac{H}{16},\frac{W}{16}]$ and the resolution increase $[\frac{H}{16},\frac{W}{16}]$ to $[{H},{W}]$, respectively.}
    \label{fig:overview}
\end{figure}

\subsection{Overall Idea}
Existing semantic segmentation frameworks are currently based on the use of the cross-entropy loss to align the probability distributions predicted by the new and old models in a pixel-wise fashion. However, this is suboptimal for two reasons: (i) the cross-entropy loss compares pixel-wise predictions independently and ignores relationships between pixels; (ii) due to the use of the softmax, the cross-entropy loss only depends on the relationships among logits and cannot directly supervise the learned feature representations.

A supervised contrastive framework \cite{khosla2020supervised} seems like a natural solution to address these problems. In fact, as recently shown in \cite{wang2021exploring}, by employing a contrastive loss, it is possible to explicitly model semantic relationships between pixels and learn an embedding space by pulling representations of pixel samples of the same class close and pushing those associated to different classes apart. 

In this paper, we propose to leverage from this idea and introduce a distillation loss that does not operate in a pixel-wise fashion but rather considers inter-pixel dependencies. In particular, we consider both relationships among representations of the new model, and between representations of the new and old models. We show that this enables better knowledge transfer and mitigates catastrophic forgetting.

\begin{figure}[t]
    \centering
    \includegraphics[width=0.95\linewidth]{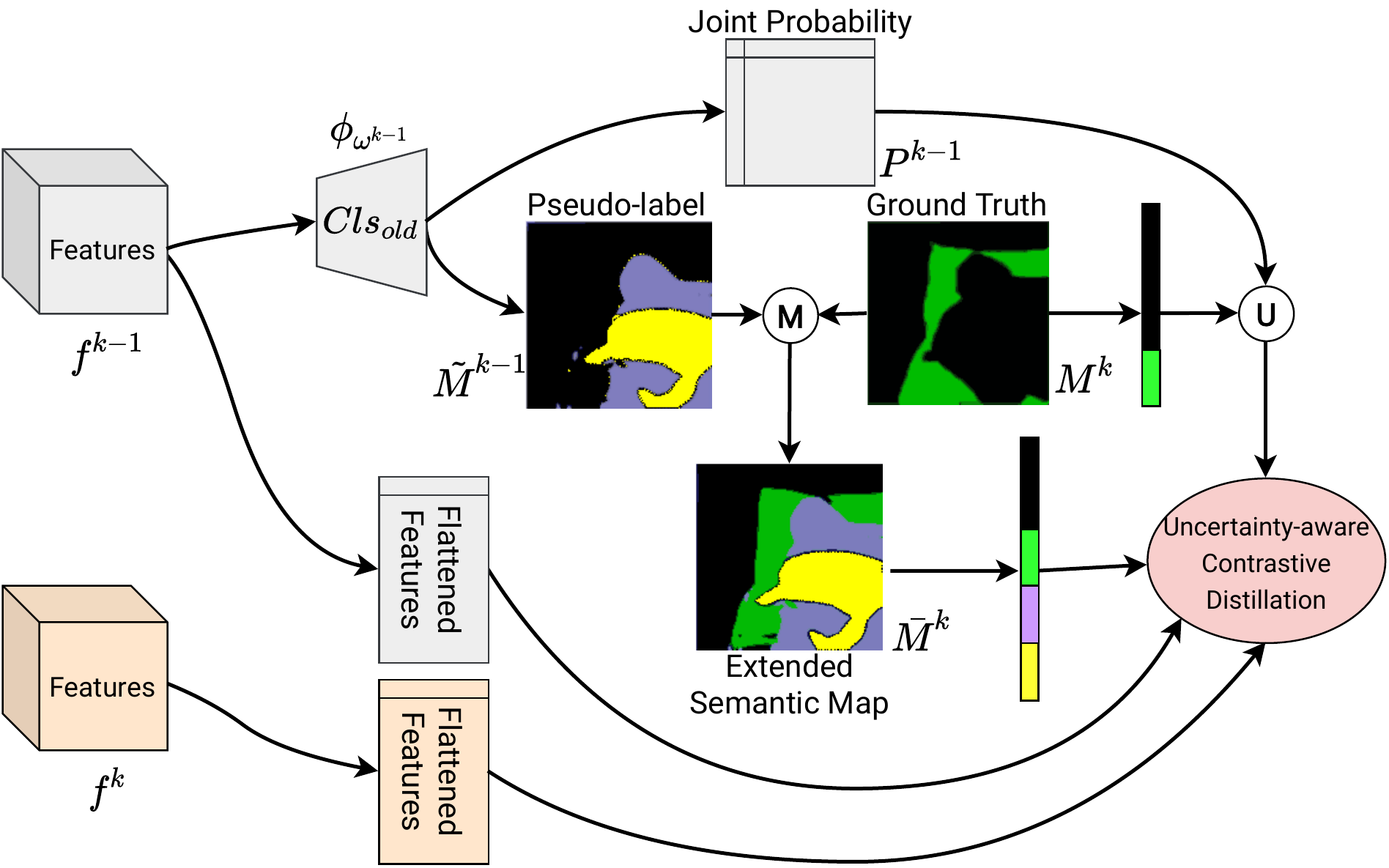}
    \caption{A detailed view of the k-th incremental learning step. The grey modules correspond to the network learned at the previous incremental learning step, which is frozen at step $k$ while the blue modules correspond to the network trained at step $k$. $\textcircled{\tiny{U}}$ is an uncertainty estimation operation. $\textcircled{\tiny{M}}$ is a merger operation.}
    \label{fig:contrastive_distillation}
\end{figure}


The overall pipeline used in the $k$-th incremental learning step is illustrated in Fig.~\ref{fig:overview}. The two streams represent the backbone for the current and previous learning steps respectively. Notably, at learning step $k$, the backbone associated with the old tasks $\psi_{\theta^{k-1}}$ and the classifier $\phi_{\omega^{k-1}}$ are frozen and used only during training. When a new batch of data sampled from the current task $k$ is received, the images are forwarded into both networks, producing two sets of features and probability distributions. The old classifier outputs probabilities for old classes only, while the new classifier also predicts the classes of the current task. In order to train the new classifier we need to provide supervision on all classes. Naturally, before introducing the contrastive distillation loss we explain how to extend the semantic map using the pseudo-labels for the old classes.

\subsection{Extended Semantic Map}
\label{sec:esm}
In this paper, we also make use of the pseudo-labels generated by the old model to counteract catastrophic forgetting. The pseudo-labels are obtained as follows:
\begin{equation}
    \tilde{M}^{k-1}_{n,p}=\arg\max(\phi_{\omega^{k-1}}(f^{k-1}_{n,p})).
\end{equation}
For each pixel in the image, $\tilde{M}^{k-1}$ contains the index of the most likely class according to the old model. Note that, since the old model is frozen during task $k$ it is not able to predict the classes in $\mathcal{Y}^k$, in fact it will probably confuse them with the background. However, since we have access to the ground truth for the current task, we can use it to correct the pseudo-label. In practice, we superimpose the ground truth (excluding the background) on top of the pseudo-labels, generating $\bar{M}^k$, the Extended Semantic Map (ESM). This merger operation, denoted as $\textcircled{\tiny{M}}$, can be summarised by the following equation:
\begin{equation}
    \bar{M}^k_{n,p} = \left\{ \begin{array}{ll}
    M^k_{n,p} & \text{ if } M^k_{n,p} \neq 0,  \\
    \tilde{M}^{k-1}_{n,p} & \text{otherwise}.\\
    \end{array}\right.
\end{equation}
With this simple mechanism, we are able to associate an ESM to each sample in the batch, and virtually to the whole dataset, namely: $\bar{\mathcal{T}}^k=\{I^k_n,\bar{M}^k_n\}_{n=1}^{N^k}$. Note that, since the pseudo-labels are estimated by a neural network, they might be noisy. The uncertainty associated with the ESM will be handled in section~\ref{uncertainty}, but first we introduce the contrastive distillation loss.

\begin{figure}[t!]
    \centering
    \includegraphics[width=0.489\textwidth]{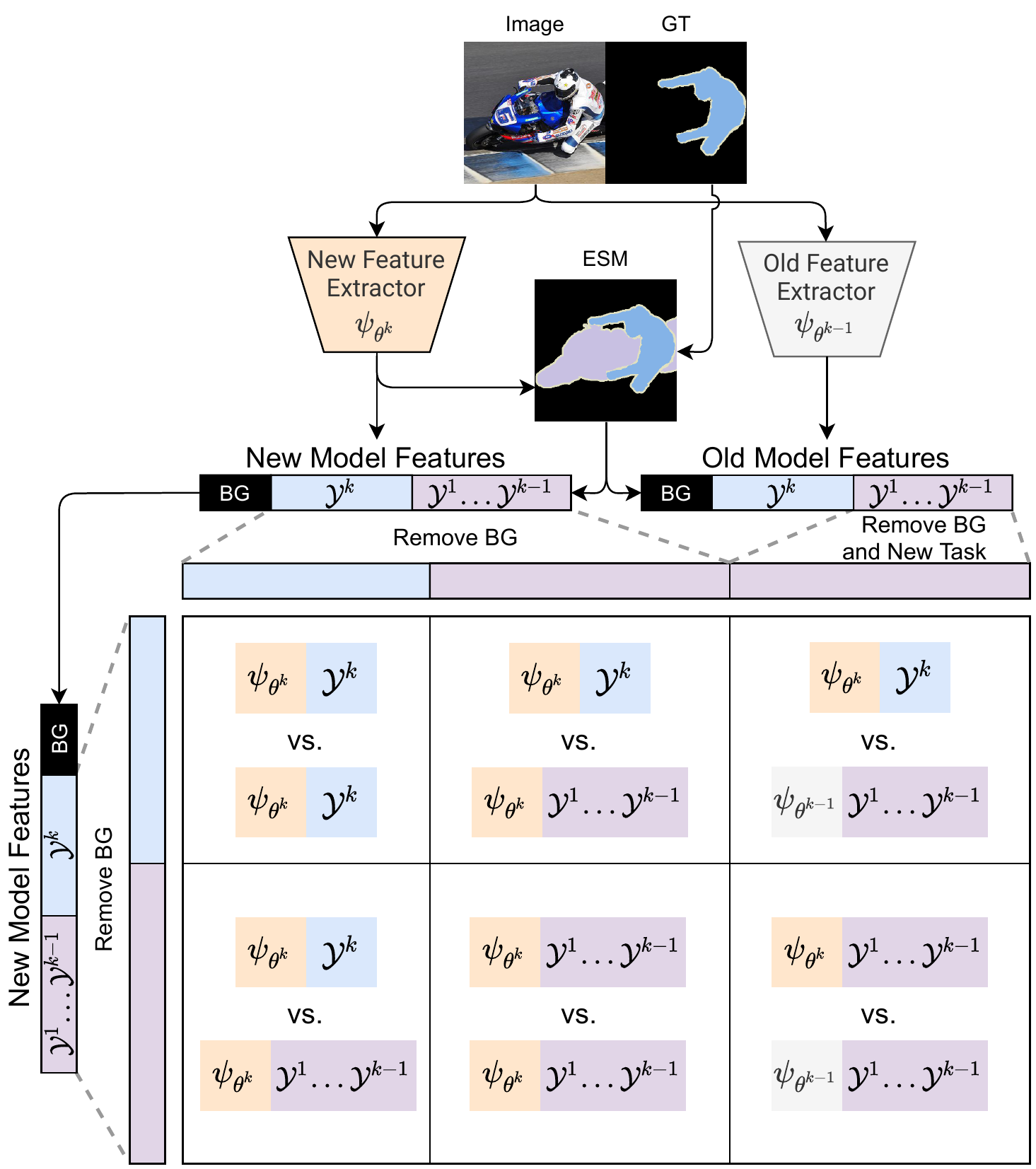}
    \caption{Visualization of the pairwise similarity matrix employed for efficient computation of the contrastive distillation loss. On the left side of the matrix we have the anchor features that are being compared with the contrast features on top. Blue and purple features and elements of the matrix belong to the new and old tasks respectively. {Grey and orange colors represent the old and new feature extractors.}}
    \label{fig:matrix}
\end{figure}

\subsection{Contrastive Distillation Loss}
\label{sec:contrastive_distillation_loss}
With this extended semantic map, we can now construct a contrastive feature distillation loss in the following way. For a given training image $I^k_n$ we can extract features with the previous and current backbones, and denote them by $f^k_n$ and $f^{k-1}_n$ respectively. {We must now interpret these two tensors as a list of $\frac{H}{16}\times \frac{W}{16}$ features of dimension $C$}. 
At this point, we need to mask out the pixels that do not contain interesting information.
{For $f^k_n$ we ignore the background pixels (\textit{i.e.} the pixels that do not belong to any of the new and old classes), since they can contain multiple objects, making their representations very heterogeneous.} We do this at batch-level by collecting the following set of indices:
\begin{equation}
\label{eq:remove_background}
    \mathcal{R}^{k} = \left\{\left(n, p\right) \mid \bar{M}^k_{n,p} \neq 0, \, p \in I^k_n, n \in \mathcal{B}\right\},
\end{equation}
where $p$ represents the coordinates of a pixel in an image and $\mathcal{B}$ is the current mini-batch. Similarly, for the features $f^{k-1}_n$ generated by the old model we need to apply an analogous masking mechanism. However, since the old model does not contain any knowledge about the new task, we need to mask out the pixels that belong to the current task, in addition to the background. As before, we gather the following indices:
\begin{equation}
    \mathcal{S}^{k} = \left\{\left(n, p\right) \mid \bar{M}^k_{n,p} \notin \mathcal{Y}^k \cup \left\{0\right\}, \, p \in I^k_n, n \in \mathcal{B}\right\}.
\end{equation}
Apart from the reasons listed above, filtering the indices of the representations of the old and new model into $\mathcal{R}^{k}$ and $\mathcal{S}^{k}$ is also useful to reduce the computational footprint of our contrastive distillation loss.
\begin{figure}[t!]
    \centering
    \includegraphics[width=0.489\textwidth]{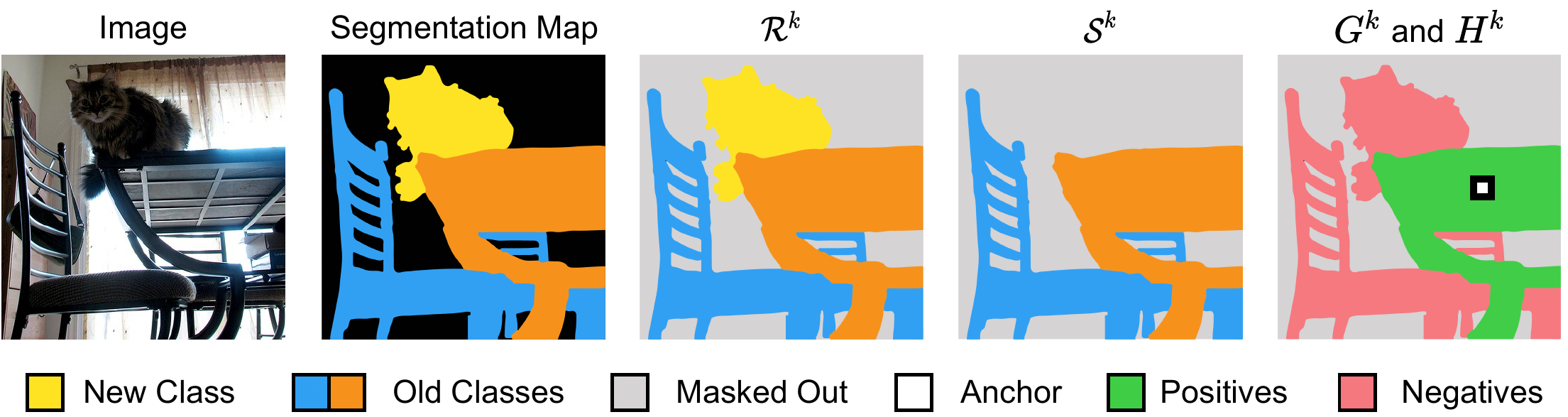}
    \caption{{An intuitive visualization of Eq.~3, 4, 5 and 6}}
    \label{fig:method_with_colors}
\end{figure}
In order to mitigate catastrophic forgetting, we want to contrast the new and old representations. Nonetheless, it is also critical that new features remain consistent with themselves, especially for pixels of new classes.
More precisely, we examine all possible new-new pairs of features, i.e.\ both features extracted from $\psi_{\theta^k}$, and all possible old-new pairs of features, i.e.\ one feature extracted from $\psi_{\theta^k}$ and one from $\psi_{\theta^{k-1}}$ (subject to their membership to $\mathcal{R}^{k}$ and $\mathcal{S}^{k}$). 
{To make them easier to grasp, we show an intuitive visualization of $\mathcal{R}^{k}$ and $\mathcal{S}^{k}$ in Fig.~\ref{fig:method_with_colors}.}
Moreover, for each anchor feature we select, we need to construct a set of positives and a set of negatives in order to perform the contrast. Given the index $a$ of an anchor, we select as positives all the pixels that belong to the same class:
\begin{equation}
\begin{aligned}
    G_{\mathcal{R}}^k(a) &= \left\{f^k_r \mid \bar{M}^k_{a} = \bar{M}^k_{r}, \, r \in \mathcal{R}^{k} \setminus \{a\}\right\}, \\
    G_{\mathcal{S}}^{k}(a) &= \left\{f^{k-1}_r \mid \bar{M}^k_{a} = \bar{M}^k_{r}, \, r \in \mathcal{S}^{k} \right\}, \\
    G^k(a) &= G_{\mathcal{R}}^k(a) \cup  G_{\mathcal{S}}^{k}(a),
\end{aligned}
\end{equation}
where $G_{\mathcal{R}}^k(a)$ and $G_{\mathcal{S}}^{k}(a)$ are the positives mined from the new and old features respectively, and $G^k(a)$ is the full set of positives for $f_a$. Similarly for the negatives:
\begin{equation}
\begin{aligned}
    H_{\mathcal{R}}^k(a) &= \left\{f^k_r \mid \bar{M}^k_{a} \neq \bar{M}^k_{r}, \, r \in \mathcal{R}^{k}\right\}, \\
    H_{\mathcal{S}}^{k}(a) &= \left\{f^{k-1}_r \mid \bar{M}^k_{a} \neq \bar{M}^k_{r}, \, r \in \mathcal{S}^{k} \right\}, \\
    H^k(a) &= H_{\mathcal{R}}^k(a) \cup  H_{\mathcal{S}}^{k}(a)
\end{aligned}
\end{equation}
{The positives $G^k(a)$ and negatives $H^k(a)$ are shown in Fig.~\ref{fig:method_with_colors} to highlight the relationship among anchor, positives and negatives.  }
Note that in this second case we are selecting all the pixels that do not share the same class with $a$ (negatives). We can now write the contrastive distillation loss as follows:
\begin{equation} \label{eq:cd}
 \mathcal{L}_{\textrm{cd}} = \frac{1}{|\mathcal{R}^k|}\sum_{a\in\mathcal{R}^{k}} \frac{-1}{|G^k(a)|}\sum_{f^+\in G^k(a)}
 \mathcal{L}_{\textrm{lc}}\left(f_a^k, f^+, a\right),
\end{equation}
where $\mathcal{L}_{\textrm{lc}}$ is the log-contrast loss involving an anchor feature $f_a^k$ generated by the new model, a positive feature $f_g$ and the set of negatives for the anchor, namely:
\begin{equation}
    \mathcal{L}_{\textrm{lc}} = \log\frac{\displaystyle\exp\left({\delta(f_{a}^k, f^+)}/{\tau}\right)}{\displaystyle\sum_{f^-\in H^k(a)}\exp\left({\delta(f_a^k, f^-)}/{\tau}\right)},
\end{equation}
where $\delta$ denotes the cosine similarity $\delta\left(u,v\right) = \frac{u}{||u||} \cdot \frac{v}{||v||}$ and $\tau$ is the temperature of the contrast. In practice to compute the contrastive distillation loss efficiently using modern deep learning libraries we compute a matrix multiplication over the whole batch to obtain the pairwise similarity matrix (see Fig.~\ref{fig:matrix}). Each element in the matrix contains the cosine similarity between two pixels. Then we sum over the rows using masks for positives and negatives to obtain the loss value.


Overall, $\mathcal{L}_{\textrm{cd}}$ pushes the features of all same-class pixels together, while pulling the pixels with different class apart. Very importantly, the proposed loss considers all features within the same batch as potential pairs, thus adding many more positive pairs, and regularising \textit{de facto} the learning. We recall that the feature extractor from the previous learning step is frozen. Therefore, features $f^{k-1}$ are also frozen and used to attract the new features $f^k$ towards a representation that encodes the classes of the previous learning step, hence the name \textit{contrastive distillation loss}.

\subsection{Uncertainty-aware Contrastive Distillation} \label{uncertainty}

As discussed in the previous section, the contrastive distillation loss requires to have pseudo-labels of the semantic classes of the previous incremental learning steps. {In order to achieve this, we exploit the semantic classifier learned at the previous step, that outputs a per-class probability tensor $P^{k-1}\in[0,1]^{\frac{H}{16}\times \frac{W}{16}\times T^{k-1}}$, $\sum_{l=1}^{T^{k-1}}P^{k-1}_{p,l}=1, \forall pcx$ (recall that $T^{k-1}$ is the total number of classes until step $k-1$).} This probability map is extended with the ground-truth at learning step $k$ in the following way:
\begin{equation}
    \bar{P}_p^k = \left\{ \begin{array}{ll}
    [P_p^{k-1},0_{L^k}] & \text{ if } M_p^k = 0,  \\
    \mathbb{I}_{M_p^k} & \text{otherwise},\\
    \end{array}\right.
\end{equation}
where, $0_{d}$ is the zero vector of dimension $d$ and $\mathbb{I}_{c}$ is a one-hot vector at coordinate $c$. In other words, the new probability tensor copies from the previous one when the ground-truth points to the background, and is a one-hot vector with the ground-truth class otherwise. 

Therefore, the extended probability tensor is $\bar{P}^{k}\in[0,1]^{H\times W\times T^{k}}$. By arg-maxing this tensor over the classes, we obtain the pseudo-labels in the extended semantic map $\bar{M}_n^K$. The direct application of these pseudo-labels in the contrastive distillation loss discussed in Eq.~(\ref{eq:cd}). Our intuition is that the arg-max operation takes a hard-decision that does not account for the uncertainty. We have therefore developed a principled strategy to account for the uncertainty in the classification of each pixel using the previous backbone.

The extended probability map can used be also to produce an estimate of the uncertainty of the pseudo-label. Indeed, one can compute the probability of two pixels $a$ (anchor) and $g$ (positive) belonging to the same class at incremental step $k$ using:
\begin{equation}
    \sigma_{a,g}^k = \bar{P}^k_a \cdot \bar{P}^k_g.
\end{equation}
This formulation naturally takes into account two important aspects that are particularly relevant for contrastive learning: (i) the \textit{confidence} of the network in predicting a certain class; (ii) the \textit{alignment} of the two probability distributions. if the network is not confident (high entropy) or the alignment is poor then $\sigma_{a,g}^k$ is low, reducing the strength of the attraction. This mechanism intrinsically accounts for the uncertainty, reducing the noise and therefore leading to better performance.

Finally, we can use this probability to modify $\mathcal{L}_{\textrm{cd}}$ in~(\ref{eq:cd}) into $\mathcal{L}_{\textrm{ucd}}$ -- the \textit{uncertainty-aware contrastive distillation} -- defined as: 
\begin{equation}    \label{eq:ucdl}
 \mathcal{L}_{\textrm{ucd}} = \frac{1}{|\mathcal{R}^k|}\sum_{a\in\mathcal{R}^{k}} \frac{-1}{|G(a)|}\sum_{f^+_g\in G(a)} \sigma_{a,g}^k \mathcal{L}_{\textrm{lc}},
\end{equation}
where $g$ is the pixel index associated to $f^+_g$.

 We also summary all step of in Fig~\ref{fig:contrastive_distillation} and explain more detail about contrastive distillation with uncertainty (red block in Fig~\ref{fig:contrastive_distillation}). 

\subsection{UCD as a generic framework}
The proposed uncertainty-aware contrastive distillation framework can be integrated seamlessly on top of virtually any other baseline. In the following we describe how to integrate UCD with two state of the art methods.

\noindent \textbf{MiB-UCD.}
\label{sec:our-method}
Recently, Cermelli \textit{et al.} \cite{cermelli2020modeling} proposed Modeling the Background for Incremental Learning (MiB), a distillation-based approach to learn the predictor $\phi_{\theta^k}$ at step $k$ given a training set $\mathcal{T}^k$ by distilling knowledge using the predictions of model $\phi_{\theta^{k-1}}$ on the previous step. In particular, they propose to  minimize the following loss function:
\begin{equation}
   \label{eq:obj-general}
    \mathcal{L}= \frac{1}{|\mathcal{T}^k| 
    }\sum_{(I_n^k,M_n^k)\in\mathcal{T}^k}
    \left(\mathcal{L}_{\textrm{seg}}(I_n^k,M_n^k) + \lambda \mathcal{L}_{\textrm{kd}}(I_n^k) \right)
\end{equation}
where $\lambda>0$ is a hyperparameter balancing the importance of two loss terms $\mathcal{L}_{\textrm{seg}}(x,y)$ and $\mathcal{L}_{KD}(x)$. These are defined in order to specifically take into account the role of the background class $\textsc{B}$, which contains different semantic categories over subsequent time steps. 
In particular, the loss $\mathcal{L}_{\textrm{seg}}$ is a revisited cross-entropy loss defined as follows:
\begin{equation}
   \label{eq:our-CE}
    \mathcal{L}_{\textrm{seg}}(I_n^k, M_{n,p}^k) = -\frac{1}{N}\sum_{p \in I}\log \tilde{\phi}_{\omega^k}(f_n^k)[p,M_{n,p}^k]\,, 
 \end{equation}
where $f_n^k = \psi_{\theta^k}(I_n^k)$ and:
{
\begin{equation}
    \label{eq:cases-ce}
    \tilde{\phi}_{\omega^k}(f_n^k)[p,c] = \begin{cases}
       \phi_{\omega^k}(f_n^k)[p,c]\;\;& \text{if}\ c\neq0\, ,\\
      \sum_{q\in\mathcal{{Y}}^{t-1}}\phi_{\omega^k}(f_n^k)[p,q]\;\;& \text{if}\ c=0\,.
    \end{cases}
\end{equation}}
This loss is meant to compensate the fact that 
the training set $\mathcal{T}^k$ might include also pixels associated to the categories previously observed in the old task. 

Furthermore, a distillation loss \cite{hinton2015distilling} is introduced in order to encourage $\phi_{\omega^k}$ to produce probabilities close to the ones produced by $\phi_{\omega^{k-1}}$. The distillation loss writes:
\begin{equation}
\small
    \label{eq:std-distill}
     \mathcal{L}_{\textrm{kd}}(I_n^k) = -\frac{1}{|I|}\sum_{i \in I}\sum_{c \in \mathcal{Y}^{k-1}}  \phi_{\omega^{k-1}}(f_n^{k-1})[i,c] \log \hat{\phi}_{\omega^{k}}(f_n^k)[i,c] \,,
\end{equation}
where:
\begin{equation}
    \label{eq:cases-ukd}
    \hat{\phi}_{\omega^{k}}(f_n^k)[p,c]= \begin{cases}
      {\phi}_{\omega^{k}}(x)[p,c]\;\;& \text{if}\ c\neq\mathtt{B}\\
      \sum_{q\in \mathcal{Y}^k}{\phi}_{\omega^{k}}(f_n^k)[p,q]\;\;& \text{if}\ c=\mathtt{B}\,.
    \end{cases}
\end{equation}
This loss is intended to account for the fact that
$\phi_{\theta^{k-1}}$ might predict as background pixels of classes that we are currently trying to learn. This aspect is crucial to perform a correct distillation of the old model into the new one.

Nonetheless, distilling the output probabilities might not be sufficient in order to defy catastrophic forgetting. For this reason, MiB seems to be a natural candidate for enhancement using UCD. The overall loss MiB-UCD is:
\begin{equation}
    \mathcal{L}=\mathcal{L}_{\textrm{seg}} + \lambda_{\textrm{kd}}\mathcal{L}_{\textrm{kd}} + \lambda_{\textrm{ucd}}\mathcal{L}_{\textrm{ucd}},
\end{equation}

\noindent\textbf{PLOP-UCD.} A different approach for incremental learning in semantic segmentation was explored in \cite{douillard2020plop}. While MiB~\cite{cermelli2020modeling} performs distillation on output probabilities, Pseudo-label and LOcal POD (PLOP)~\cite{douillard2020plop} proposes to use intermediate representations to transfer richer information. This is achieved using a technique called Pooled Output Distillation (POD)~\cite{douillard2020podnet} that was already shown to work well in classification settings. In addition, in order to preserve details that would be neglected by the pooling, PLOP uses a multi-scale distillation scheme. In practice, for every layer $l$ of the network, pooled features (Local POD embeddings) are collected at several scales and concatenated to obtain a feature vector $\mathcal{F}_l$. This process is carried out on both old and current models, and then the concatenated features are compared using an $_2$ loss:
\begin{equation}
\mathcal{L}_{\textrm{pod}}=\frac{1}{L} \sum_{l=1}^{L}\left\|\mathcal{F}_l^{k} -\mathcal{F}_{l}^{k-1} \right\|^{2},
\end{equation}
where $L$ is the total number of layers of the network.

Besides Local POD, PLOP also introduces a simpler way of solving background shift: a pseudo-labeling strategy for background pixels. Predictions of the old model for background pixels are used as clues regarding their real class, if they belong to any of the old classes. An $\textrm{argmax}$ operation is applied on every pixel, subject to a class-specific confidence threshold. In other words, in the case of non-background pixels they copy the ground truth label; otherwise, if the old model is confident enough, the most likely class predicted by the old model is selected. The pseudo-label $S$ is then used in the classification loss as follows:
\begin{equation}
\mathcal{L}_{\text {pseudo }} = -\frac{\nu}{|I|}\sum_{p \in I}\log \phi_{\omega^k}(f_n^k)[p,S_{n,p}]\,, 
\end{equation}
where $\nu$ is the ratio of accepted old classes pixels over the total number of such pixels.

Combining these two ideas, PLOP was shown to be able to outperform MiB in the overlapped setting, while its superiority on the harder disjoint setting is still unclear. Nonetheless, we believe PLOP can take advantage of our contrastive distillation loss. In fact, although it already performs distilation of intermediate features, PLOP lacks two critical aspects: (i) it does not take into account the relationships between pixel representations for different images (ii) it does not use the pseudo-label for feature distillation. Also, the way PLOP estimates the confidence of the pseudo-label is very primitive and cursed with class-specific hyper-parameter tuning. On the contrary, our UCD loss naturally accounts for the uncertainty in a more elegant way. The overall loss of PLOP-UCD is the following:
\begin{equation}
    \mathcal{L}=\mathcal{L}_{\textrm{pseudo}} + \lambda_{\textrm{pod}}\mathcal{L}_{\textrm{pod}} + \lambda_{\textrm{ucd}}\mathcal{L}_{\textrm{ucd}},
\end{equation}

%% file: experiments_clean.tex
\begin{table*}[!h]
\caption{Mean IoU on the Pascal-VOC 2012 dataset for different incremental class learning scenarios. $*$ means results come from re-implementation. $\dag$ means a updated version.
$\sharp$ means the conditional GAN model pretrained on ImageNet~\cite{deng2009imagenet} is used.
UDC means uncertainty-aware contrastive distillation. Best among table in \textbf{bold}, best among part in \underline{underlined}.}
\label{tab:overall voc}
\resizebox{\textwidth}{!}{%
\begin{tabular}{lcccccccccccccccccc}
\toprule
\multirow{3.5}{*}{\textbf{Method}} & \multicolumn{6}{c|}{\textbf{19-1}} & \multicolumn{6}{c|}{\textbf{15-5}} & \multicolumn{6}{c}{\textbf{15-1}} \\\cmidrule{2-19}
& \multicolumn{3}{c|}{\textbf{Disjoint}} & \multicolumn{3}{c|}{\textbf{Overlapped}} & \multicolumn{3}{c|}{\textbf{Disjoint}} & \multicolumn{3}{c|}{\textbf{Overlapped}} & \multicolumn{3}{c|}{\textbf{Disjoint}} & \multicolumn{3}{c}{\textbf{Overlapped}} \\\cmidrule{2-19}
& \textbf{1-19} & \multicolumn{1}{c|}{\textbf{20}} & \multicolumn{1}{c|}{\textbf{all}} & \textbf{1-19} & \multicolumn{1}{c|}{\textbf{20}} & \multicolumn{1}{c|}{\textbf{all}} & \textbf{1-15} & \multicolumn{1}{c|}{\textbf{16-20}} & \multicolumn{1}{c|}{\textbf{all}} & \textbf{1-15} & \multicolumn{1}{c|}{\textbf{16-20}} & \multicolumn{1}{c|}{\textbf{all}} & \textbf{1-15} & \multicolumn{1}{c|}{\textbf{16-20}} & \multicolumn{1}{c|}{\textbf{all}} & \textbf{1-15} & \multicolumn{1}{c|}{\textbf{16-20}} & \textbf{all} \\
\midrule
\multicolumn{1}{l|}{FT} & 5.8 & \multicolumn{1}{c|}{12.3} & \multicolumn{1}{c|}{6.2} & 6.8 & \multicolumn{1}{c|}{12.9} & \multicolumn{1}{c|}{7.1} & 1.1 & \multicolumn{1}{c|}{33.6} & \multicolumn{1}{c|}{9.2} & 2.1 & \multicolumn{1}{c|}{33.1} & \multicolumn{1}{c|}{9.8} & 0.2 & \multicolumn{1}{c|}{1.8} & \multicolumn{1}{c|}{0.6} & 0.2 & \multicolumn{1}{c|}{1.8} & 0.6 \\
\multicolumn{1}{l|}{PI\cite{zenke2017continual}} & 5.4 & \multicolumn{1}{c|}{14.1} & \multicolumn{1}{c|}{5.9} & 7.5 & \multicolumn{1}{c|}{14.0} & \multicolumn{1}{c|}{7.8} & 1.3 & \multicolumn{1}{c|}{34.1} & \multicolumn{1}{c|}{9.5} & 1.6 & \multicolumn{1}{c|}{33.3} & \multicolumn{1}{c|}{9.5} & 0.0 & \multicolumn{1}{c|}{1.8} & \multicolumn{1}{c|}{0.4} & 0.0 & \multicolumn{1}{c|}{1.8} & 0.4 \\
\multicolumn{1}{l|}{EWC\cite{kirkpatrick2017overcoming}} & 23.2 & \multicolumn{1}{c|}{16.0} & \multicolumn{1}{c|}{22.9} & 26.9 & \multicolumn{1}{c|}{14.0} & \multicolumn{1}{c|}{26.3} & 26.7 & \multicolumn{1}{c|}{37.7} & \multicolumn{1}{c|}{29.4} & 24.3 & \multicolumn{1}{c|}{35.5} & \multicolumn{1}{c|}{27.1} & 0.3 & \multicolumn{1}{c|}{4.3} & \multicolumn{1}{c|}{1.3} & 0.3 & \multicolumn{1}{c|}{4.3} & 1.3 \\
\multicolumn{1}{l|}{RW\cite{chaudhry2018riemannian}} & 19.4 & \multicolumn{1}{c|}{15.7} & \multicolumn{1}{c|}{19.2} & 23.3 & \multicolumn{1}{c|}{14.2} & \multicolumn{1}{c|}{22.9} & 17.9 & \multicolumn{1}{c|}{36.9} & \multicolumn{1}{c|}{22.7} & 16.6 & \multicolumn{1}{c|}{34.9} & \multicolumn{1}{c|}{21.2} & 0.2 & \multicolumn{1}{c|}{5.4} & \multicolumn{1}{c|}{1.5} & 0.0 & \multicolumn{1}{c|}{5.2} & 1.3 \\
\multicolumn{1}{l|}{LwF\cite{li2017learning}} & 53.0 & \multicolumn{1}{c|}{9.1} & \multicolumn{1}{c|}{50.8} & 51.2 & \multicolumn{1}{c|}{8.5} & \multicolumn{1}{c|}{49.1} & 58.4 & \multicolumn{1}{c|}{37.4} & \multicolumn{1}{c|}{53.1} & 58.9 & \multicolumn{1}{c|}{36.6} & \multicolumn{1}{c|}{53.3} & 0.8 & \multicolumn{1}{c|}{3.6} & \multicolumn{1}{c|}{1.5} & 1.0 & \multicolumn{1}{c|}{3.9} & 1.8 \\
\multicolumn{1}{l|}{LwF-MC\cite{rebuffi2017icarl}} & 63.0 & \multicolumn{1}{c|}{13.2} & \multicolumn{1}{c|}{60.5} & 64.4 & \multicolumn{1}{c|}{13.3} & \multicolumn{1}{c|}{61.9} & 67.2 & \multicolumn{1}{c|}{41.2} & \multicolumn{1}{c|}{60.7} & 58.1 & \multicolumn{1}{c|}{35.0} & \multicolumn{1}{c|}{52.3} & 4.5 & \multicolumn{1}{c|}{7.0} & \multicolumn{1}{c|}{5.2} & 6.4 & \multicolumn{1}{c|}{8.4} & 6.9 \\
\multicolumn{1}{l|}{ILT\cite{michieli2019incremental}} & 69.1 & \multicolumn{1}{c|}{16.4} & \multicolumn{1}{c|}{66.4} & 67.1 & \multicolumn{1}{c|}{12.3} & \multicolumn{1}{c|}{64.4} & 63.2 & \multicolumn{1}{c|}{39.5} & \multicolumn{1}{c|}{57.3} & 66.3 & \multicolumn{1}{c|}{40.6} & \multicolumn{1}{c|}{59.9} & 3.7 & \multicolumn{1}{c|}{5.7} & \multicolumn{1}{c|}{4.2} & 4.9 & \multicolumn{1}{c|}{7.8} & 5.7 \\
\multicolumn{1}{l|}{ILT$^\dag$\cite{michieli2021knowledge}} & 69.4 & \multicolumn{1}{c|}{16.5} & \multicolumn{1}{c|}{66.7} & 67.4 & \multicolumn{1}{c|}{12.4} & \multicolumn{1}{c|}{64.7} & 63.3 & \multicolumn{1}{c|}{39.6} & \multicolumn{1}{c|}{57.4} & 66.4 & \multicolumn{1}{c|}{40.8} & \multicolumn{1}{c|}{60.0} & 4.4 & \multicolumn{1}{c|}{6.4} & \multicolumn{1}{c|}{4.9} & 5.5 & \multicolumn{1}{c|}{8.0} & 6.1 \\
\multicolumn{1}{l|}{SDR\cite{michieli2021continual}} & 69.9 & \multicolumn{1}{c|}{{37.3}} & \multicolumn{1}{c|}{68.4} & 69.1 & \multicolumn{1}{c|}{32.6} & \multicolumn{1}{c|}{67.4} & {\underline{73.5}} & \multicolumn{1}{c|}{{47.3}} & \multicolumn{1}{c|}{{\underline{67.2}}} & 75.4 & \multicolumn{1}{c|}{52.6} & \multicolumn{1}{c|}{69.9} & {{59.2}} & \multicolumn{1}{c|}{12.9} & \multicolumn{1}{c|}{{{48.1}}} & 44.7 & \multicolumn{1}{c|}{{21.8}} & 39.2 \\
\multicolumn{1}{l|}{{RECALL$^{\sharp}$\cite{maracani2021recall}}} & 65.2 & \multicolumn{1}{c|}{\textbf{\underline{50.1}}} & \multicolumn{1}{c|}{65.8} & 67.9 & \multicolumn{1}{c|}{\textbf{\underline{53.5}} } & \multicolumn{1}{c|}{68.4} & {{66.3}} & \multicolumn{1}{c|}{\textbf{\underline{49.8}}} & \multicolumn{1}{c|}{{{63.5}}} & 66.6 & \multicolumn{1}{c|}{50.9} & \multicolumn{1}{c|}{64.0} & \textbf{\underline{66.0}} & \multicolumn{1}{c|}{\textbf{\underline{44.9}} } & \multicolumn{1}{c|}{\textbf{\underline{62.1}}} & \underline{65.7} & \multicolumn{1}{c|}{\textbf{\underline{47.8}}} & \textbf{\underline{62.7}} \\
\multicolumn{1}{l|}{{UCD}} & \underline{73.4} & \multicolumn{1}{c|}{{33.7}} & \multicolumn{1}{c|}{\underline{71.5}} & \underline{71.4} & \multicolumn{1}{c|}{47.3} & \multicolumn{1}{c|}{\underline{70.0}} & {{71.9}} & \multicolumn{1}{c|}{{49.5}} & \multicolumn{1}{c|}{{{66.2}}} & \underline{77.5} & \multicolumn{1}{c|}{\textbf{\underline{53.1}} } & \multicolumn{1}{c|}{\underline{71.3} } & {{53.1}} & \multicolumn{1}{c|}{13.0} & \multicolumn{1}{c|}{{{42.9}}} & 49.0 & \multicolumn{1}{c|}{{19.5}} & 41.9 \\
\midrule
\multicolumn{1}{l|}{MiB\cite{cermelli2020modeling}} & 69.6 & \multicolumn{1}{c|}{25.6} & \multicolumn{1}{c|}{67.4} & 70.2 & \multicolumn{1}{c|}{22.1} & \multicolumn{1}{c|}{67.8} & 71.8 & \multicolumn{1}{c|}{43.3} & \multicolumn{1}{c|}{64.7} & 75.5 & \multicolumn{1}{c|}{49.4} & \multicolumn{1}{c|}{69.0} & 46.2 & \multicolumn{1}{c|}{12.9} & \multicolumn{1}{c|}{37.9} & 35.1 & \multicolumn{1}{c|}{13.5} & 29.7 \\
\multicolumn{1}{l|}{MiB+SDR\cite{michieli2021continual}} & 70.8 & \multicolumn{1}{c|}{\underline{31.4}} & \multicolumn{1}{c|}{68.9} & 71.3 & \multicolumn{1}{c|}{23.4} & \multicolumn{1}{c|}{69.0} & \textbf{\underline{74.6}} & \multicolumn{1}{c|}{{44.1}} & \multicolumn{1}{c|}{\textbf{\underline{67.3}}} & 76.3 & \multicolumn{1}{c|}{50.2} & \multicolumn{1}{c|}{70.1} & {\underline{59.4}} & \multicolumn{1}{c|}{14.3} & \multicolumn{1}{c|}{{\underline{48.7}}} & 47.3 & \multicolumn{1}{c|}{\underline{14.7}} & 39.5 \\
\multicolumn{1}{l|}{MiB+UCD} & \underline{74.3} & \multicolumn{1}{c|}{28.4} & \multicolumn{1}{c|}{\underline{72.0}} & \underline{73.7} & \multicolumn{1}{c|}{\underline{34.0}} & \multicolumn{1}{c|}{\underline{71.7}} & 73.0 & \multicolumn{1}{c|}{\textbf{\underline{46.2}}} & \multicolumn{1}{c|}{66.3} & \textbf{\underline{78.5}} & \multicolumn{1}{c|}{{\underline{50.7}}} & \multicolumn{1}{c|}{\textbf{\underline{71.5}}} & 53.3 & \multicolumn{1}{c|}{{\underline{14.4}}} & \multicolumn{1}{c|}{{43.5}} & \underline{51.9} & \multicolumn{1}{c|}{13.1} & \underline{42.2} \\
\midrule
\multicolumn{1}{l|}{PLOP$^*$~\cite{douillard2020plop}} & 75.1 & \multicolumn{1}{c|}{\underline{38.2}} & \multicolumn{1}{c|}{73.2} & {75.0} & \multicolumn{1}{c|}{{39.1}} & \multicolumn{1}{c|}{{73.2}} & 66.5 & \multicolumn{1}{c|}{\underline{39.6}} & \multicolumn{1}{c|}{59.8} & {74.7} & \multicolumn{1}{c|}{{49.8}} & \multicolumn{1}{c|}{{68.5}} & 49.0 & \multicolumn{1}{c|}{\underline{13.8}} & \multicolumn{1}{c|}{40.2} & {65.2} & \multicolumn{1}{c|}{{\underline{22.4}}} & {54.5} \\
\multicolumn{1}{l|}{PLOP+ UCD} & \textbf{\underline{75.7}} & \multicolumn{1}{c|}{{31.8}} & \multicolumn{1}{c|}{\textbf{\underline{73.5}}} & \textbf{\underline{75.9}} & \multicolumn{1}{c|}{{\underline{39.5}}} & \multicolumn{1}{c|}{\textbf{\underline{74.0}}} & \underline{67.0} & \multicolumn{1}{c|}{39.3} & \multicolumn{1}{c|}{\underline{60.1}} & \underline{75.0} & \multicolumn{1}{c|}{\underline{51.8}} & \multicolumn{1}{c|}{\underline{69.2}} & \underline{50.8} & \multicolumn{1}{c|}{13.3} & \multicolumn{1}{c|}{\underline{41.4}} & \textbf{\underline{66.3}} & \multicolumn{1}{c|}{21.6} & {\underline{55.1}} \\
\midrule
\multicolumn{1}{l|}{Joint} & 77.4 & \multicolumn{1}{c|}{78.0} & \multicolumn{1}{c|}{77.4} & 77.4 & \multicolumn{1}{c|}{78.0} & \multicolumn{1}{c|}{77.4} & 79.1 & \multicolumn{1}{c|}{72.6} & \multicolumn{1}{c|}{77.4} & 79.1 & \multicolumn{1}{c|}{72.6} & \multicolumn{1}{c|}{77.4} & 79.1 & \multicolumn{1}{c|}{72.6} & \multicolumn{1}{c|}{77.4} & 79.1 & \multicolumn{1}{c|}{72.6} & 77.4\\
\bottomrule
\end{tabular}%
}
\end{table*}

\begin{table*}[]
\centering
\caption{{Mean IoU on the Pascal-VOC 2012 dataset for more incremental class learning scenarios. $*$ means results come from re-implementation.
$\sharp$ means the conditional GAN model pretrained on ImageNet~\cite{deng2009imagenet} is used.
UDC means uncertainty-aware contrastive distillation.
Best among table in \textbf{bold}, best among part in \underline{underlined}.}}
\label{tab:more voc}

\begin{tabular}{lcc|c|cc|c|cc|c|cc|c}
\toprule
\multirow{3}[3]{*}{\textbf{Method}} & \multicolumn{6}{c|}{\textbf{10-10}}& \multicolumn{6}{c}{\textbf{10-1}}  \\\cmidrule{2-13}
& \multicolumn{3}{c|}{\textbf{Disjoint}} & \multicolumn{3}{c|}{\textbf{Overlapped}} & \multicolumn{3}{c|}{\textbf{Disjoint}} & \multicolumn{3}{c}{\textbf{Overlapped}} \\\cmidrule{2-13}
 & \textbf{1-10}    & \textbf{11-20}    & \textbf{all}     & \textbf{1-10}     & \textbf{11-20}     & \textbf{all}     & \textbf{1-10}    & \textbf{11-20}    & \textbf{all}     & \textbf{1-10}     & \textbf{11-20}     & \textbf{all}     \\
\midrule
FT & 7.7 & 60.8 & 33.0 & 7.8 & 58.9 & 32.1 & 6.3 & 2.0 & 4.3 & 6.3 & 2.8 & 4.7 \\
LwF\cite{li2017learning} & 63.1 & 61.1 & 62.2 & \underline{70.7} & \underline{63.4} & \underline{67.2} & 6.7 & 6.5 & 6.6 & 16.6 & 14.9 & 15.8 \\
LwF-MC\cite{rebuffi2017icarl} & 52.4 & 42.5 & 47.7 & 53.9 & 43.0 & 48.7 & 6.9 & 1.7 & 4.4 & 11.2 & 2.5 & 7.1 \\
ILT\cite{michieli2019incremental} & \textbf{\underline{67.7}} & \textbf{\underline{61.3}} & \textbf{\underline{64.7}} & 70.3 & 61.9 & 66.3 & 14.1 & 0.6 & 7.5 & 16.5 & 1.0 & 9.1 \\
{RECALL$^{\sharp}$\cite{maracani2021recall} }& 62.6 & 56.1 & 60.8 & 65.0 & 58.4 & 63.1 & \textbf{\underline{58.3}} & \textbf{\underline{46.0}} & \textbf{\underline{53.9}} & \textbf{\underline{59.5}} &\textbf{ \underline{46.7}} & \textbf{\underline{54.8}} \\
\midrule
MiB\cite{cermelli2020modeling} & 66.9 & 57.5 & 62.4 & 70.4 & 63.7 & 67.2 & 14.9 & 9.5 & 12.3 & 15.1 & 14.8 & 15.0 \\
MiB+SDR\cite{michieli2021continual} & \underline{67.5} & 57.9 & \underline{62.9} & 70.5 & 63.9 & 67.4 & 25.5 & 15.7 & 20.8 & 26.3 & 19.7 & 23.2 \\
MiB+UCD & 65.0 & \underline{58.7}  & 61.9  & \textbf{\underline{71.8}}  & \textbf{\underline{65.2}} & \textbf{\underline{68.5}} & {\underline{33.1}} & {\underline{26.1}} & {\underline{30.6}} & {\underline{33.7}} & {\underline{26.5}} & {\underline{31.1}}\\
\midrule
PLOP$^*$~\cite{douillard2020plop} & 61.8 & \multicolumn{1}{c|}{{53.1}} & \multicolumn{1}{c|}{57.5} & {65.0} & \multicolumn{1}{c|}{{58.8}} & \multicolumn{1}{c|}{{61.9}} & 39.5 & \multicolumn{1}{c|}{{13.9}} & \multicolumn{1}{c|}{26.7}  & {40.0} & \multicolumn{1}{c|}{{{21.7}}} & {30.8} \\
{PLOP+ UCD} & {\underline{63.0}} & \multicolumn{1}{c|}{\underline{55.8}} & \multicolumn{1}{c|}{{\underline{59.4}}} & {\underline{71.5}} & \multicolumn{1}{c|}{{\underline{58.9}}} & \multicolumn{1}{c|}{{\underline{65.2}}} & {\underline{42.6}} & \multicolumn{1}{c|}{{\underline{15.0}}} & {\underline{28.8}} & \multicolumn{1}{c}{{\underline{42.3}}} & \multicolumn{1}{c|}{{\underline{28.3}}} & {\underline{35.3}} \\
\midrule
Joint & 78.6 & 76.0 & 77.4 & 78.6 & 76.0 & 77.4 & 78.6 & 76.0 & 77.4 & 78.6 & 76.0 & 77.4 \\  
\bottomrule
\end{tabular}
\end{table*}

\begin{figure*}[t]
\centering
\subfigure[Image]{
    \begin{minipage}{0.14\linewidth}
        \centering
        \includegraphics[width=0.993\textwidth,height=0.9in]{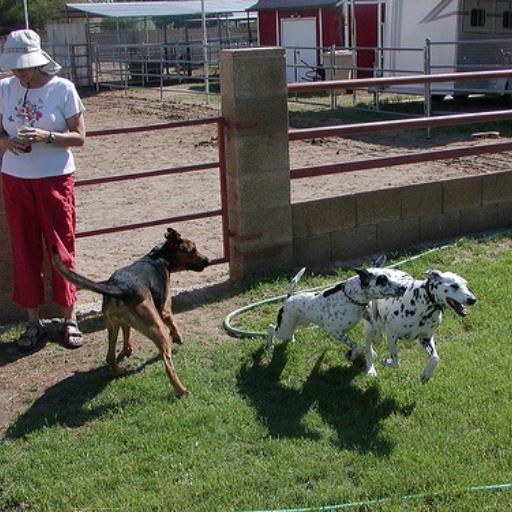}\\ 
        \includegraphics[width=0.993\textwidth,height=0.9in]{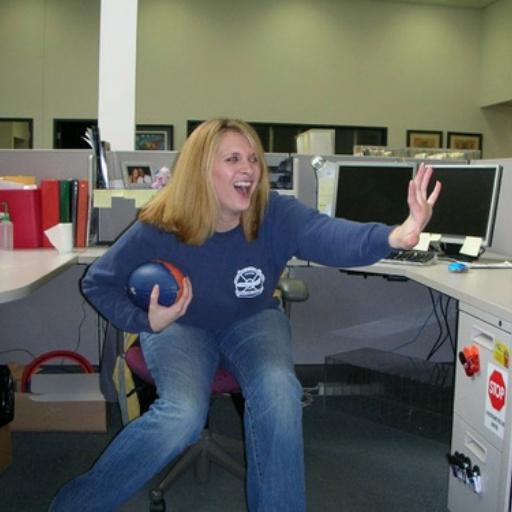}\\ 
        \includegraphics[width=0.993\textwidth,height=0.9in]{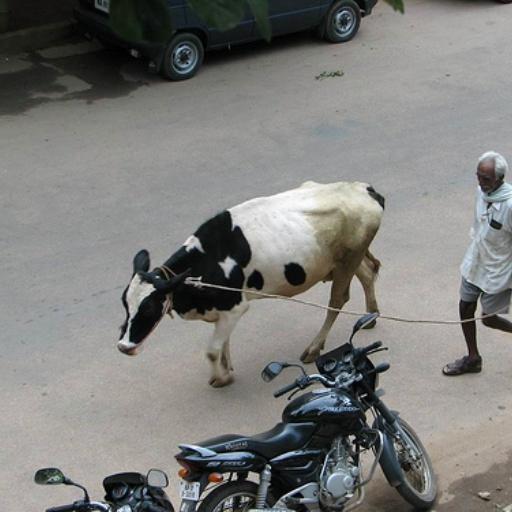}\\ 
        \includegraphics[width=0.993\textwidth,height=0.9in]{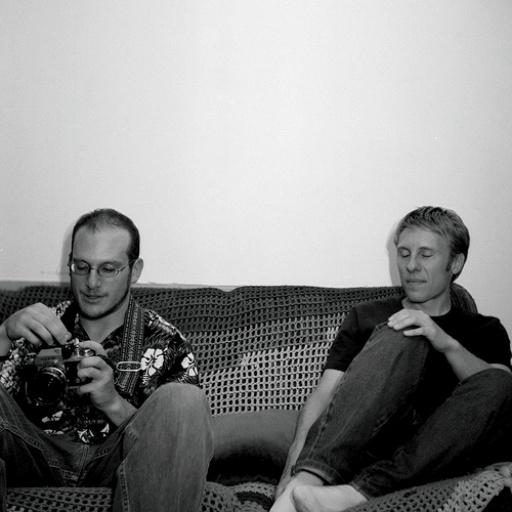}\\
        \vspace{2px}
    \end{minipage}%
}%
\subfigure[ILT~\cite{michieli2019incremental}]{
    \begin{minipage}{0.14\linewidth}
        \centering
        \includegraphics[width=0.993\textwidth,height=0.9in]{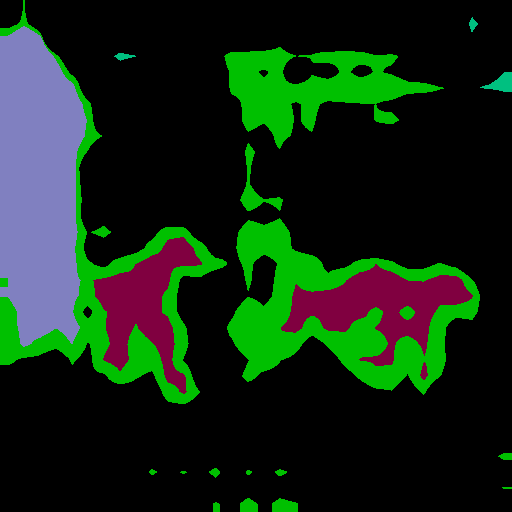}\\ 
        \includegraphics[width=0.993\textwidth,height=0.9in]{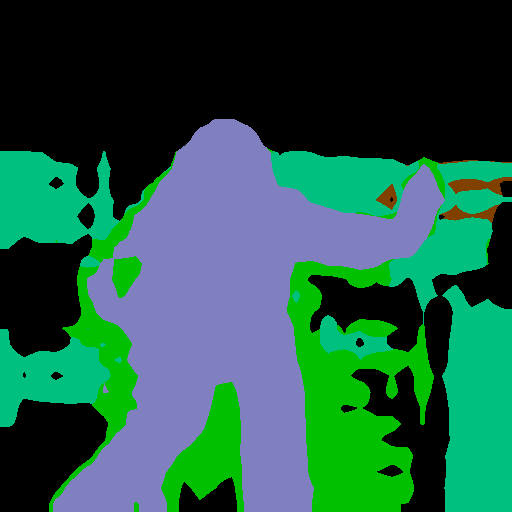}\\ 
        \includegraphics[width=0.993\textwidth,height=0.9in]{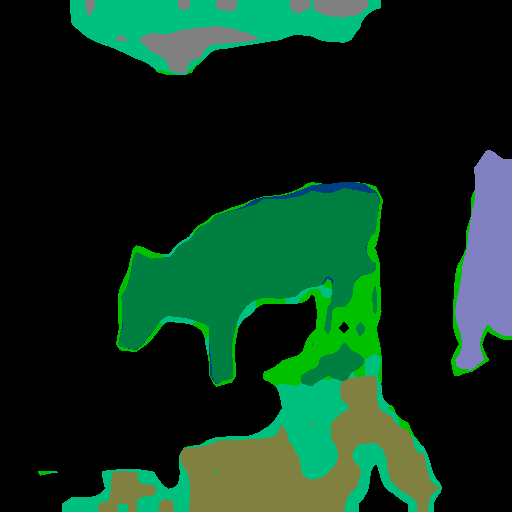}\\ 
        \includegraphics[width=0.993\textwidth,height=0.9in]{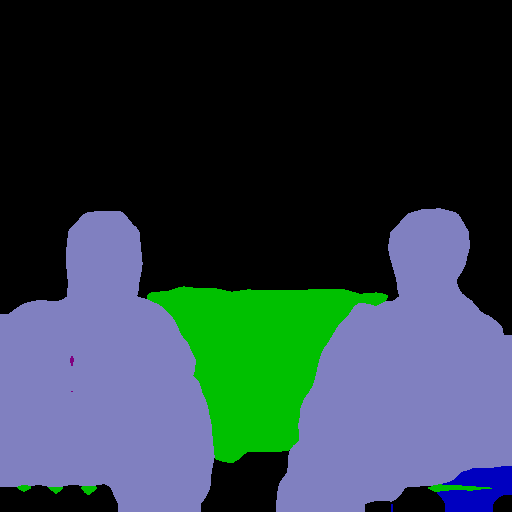}\\ 
        \vspace{2px}
    \end{minipage}%
}%
\subfigure[PLOP~\cite{douillard2020plop}]{
    \begin{minipage}{0.14\linewidth}
        \centering
        \includegraphics[width=0.993\textwidth,height=0.9in]{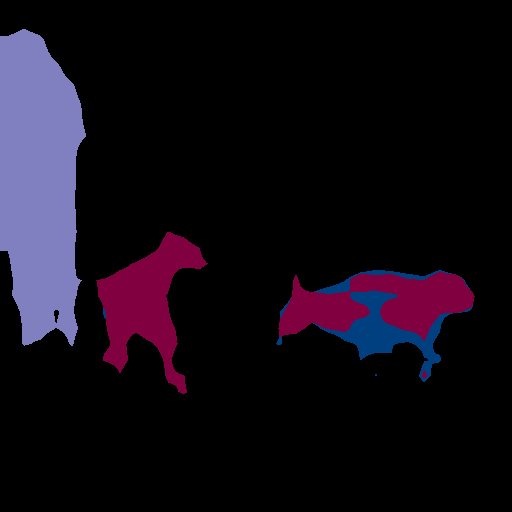}\\ 
        \includegraphics[width=0.993\textwidth,height=0.9in]{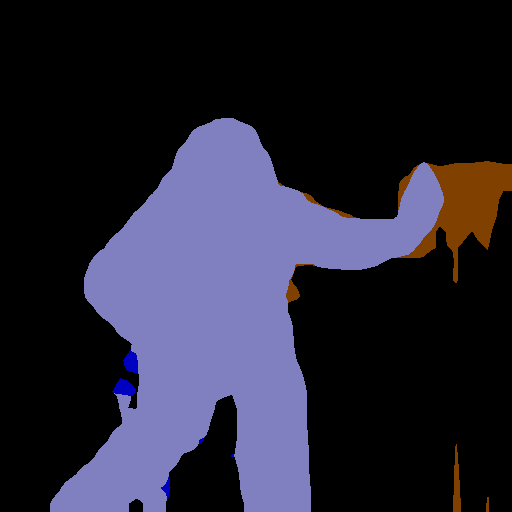}\\ 
        \includegraphics[width=0.993\textwidth,height=0.9in]{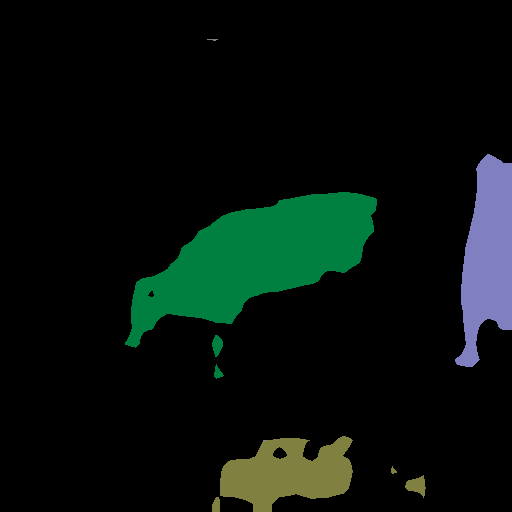}\\ 
        \includegraphics[width=0.993\textwidth,height=0.9in]{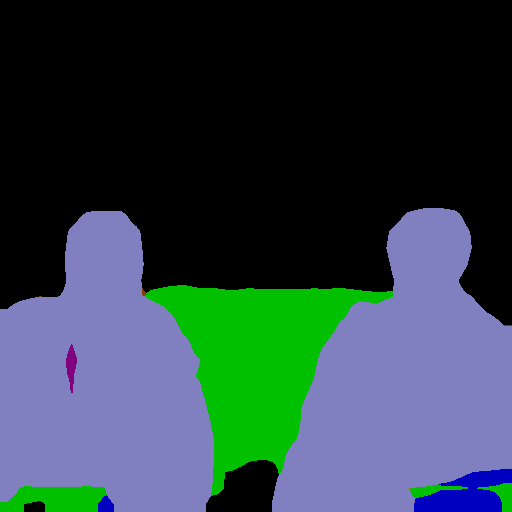}\\ 
        \vspace{2px}
    \end{minipage}%
}%
\subfigure[PLOP+UCD]{
    \begin{minipage}{0.14\linewidth}
        \centering
        \includegraphics[width=0.993\textwidth,height=0.9in]{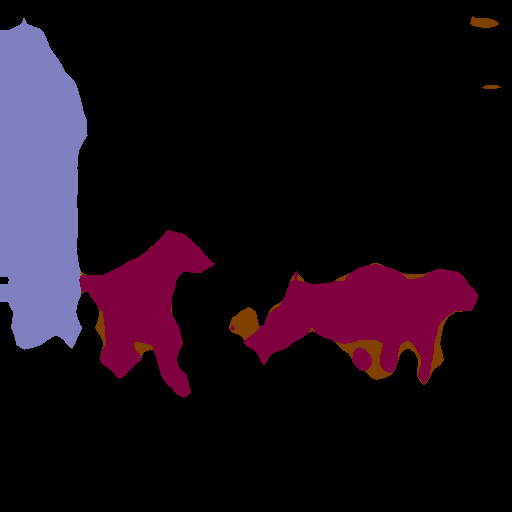}\\ 
        \includegraphics[width=0.993\textwidth,height=0.9in]{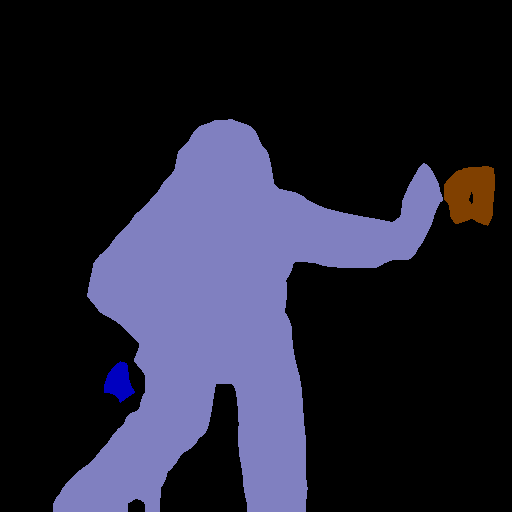}\\ 
        \includegraphics[width=0.993\textwidth,height=0.9in]{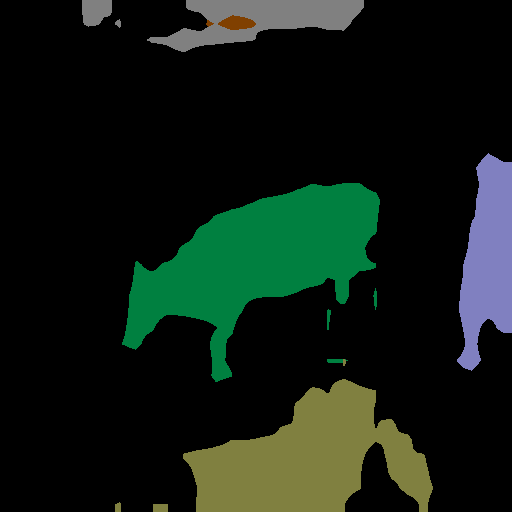}\\ 
        \includegraphics[width=0.993\textwidth,height=0.9in]{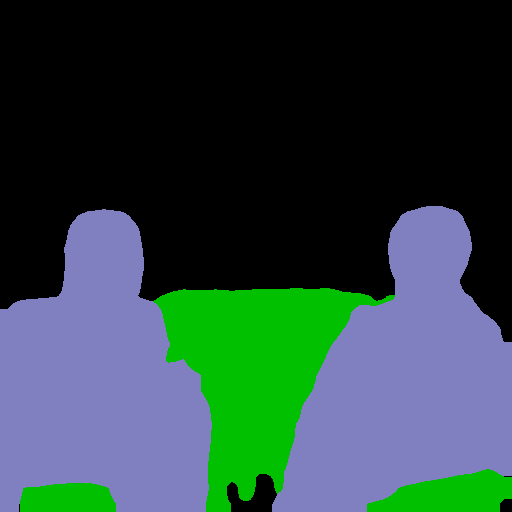}\\
        \vspace{2px}
    \end{minipage}%
}%
\subfigure[MiB~\cite{cermelli2020modeling}]{
    \begin{minipage}{0.14\linewidth}
        \centering
        \includegraphics[width=0.993\textwidth,height=0.9in]{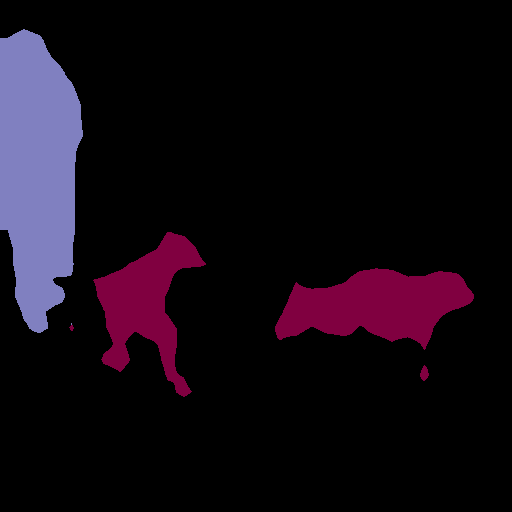}\\ 
        \includegraphics[width=0.993\textwidth,height=0.9in]{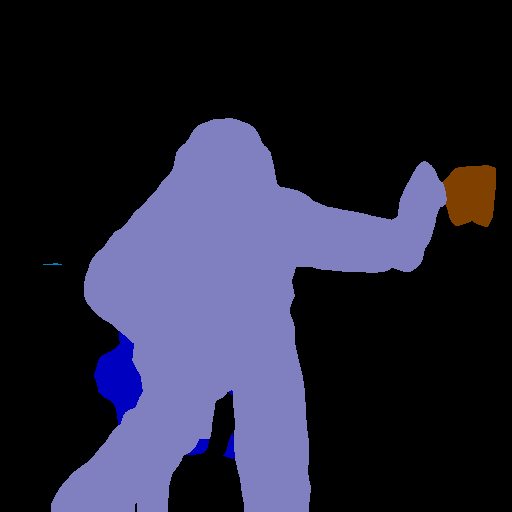}\\ 
        \includegraphics[width=0.993\textwidth,height=0.9in]{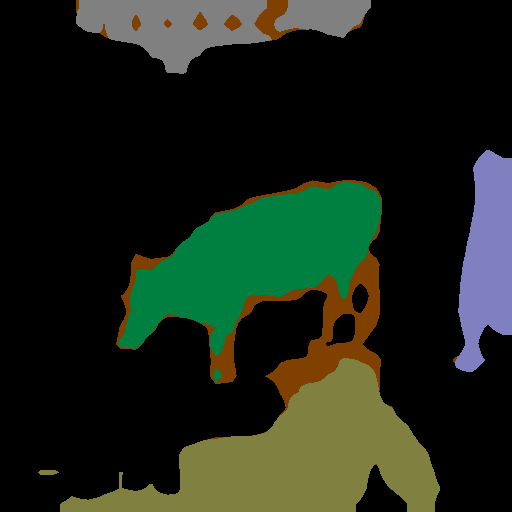}\\ 
        \includegraphics[width=0.993\textwidth,height=0.9in]{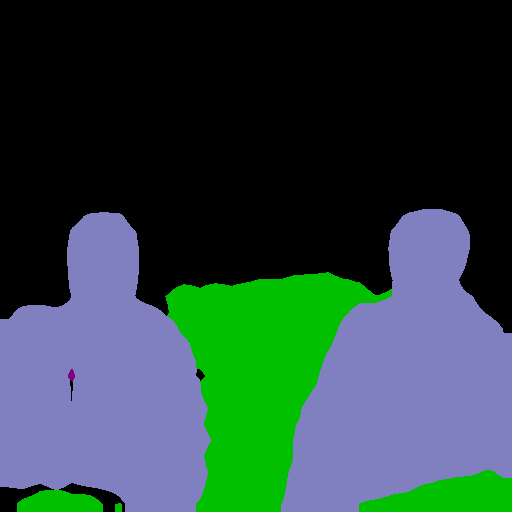}\\ 
        \vspace{2px}
    \end{minipage}%
}%
\subfigure[MiB+UCD]{
    \begin{minipage}{0.14\linewidth}
        \centering
        \includegraphics[width=0.993\textwidth,height=0.9in]{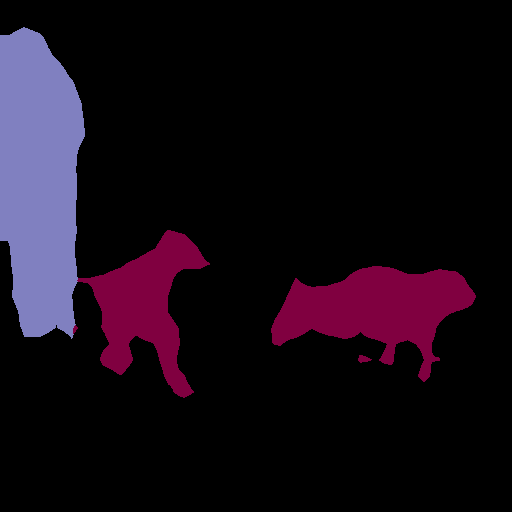}\\ 
        \includegraphics[width=0.993\textwidth,height=0.9in]{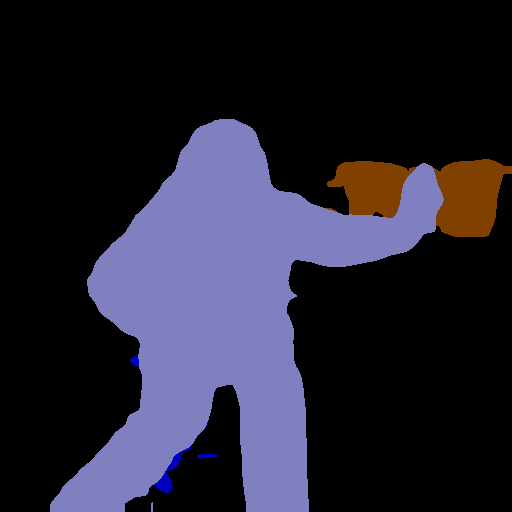}\\ 
        \includegraphics[width=0.993\textwidth,height=0.9in]{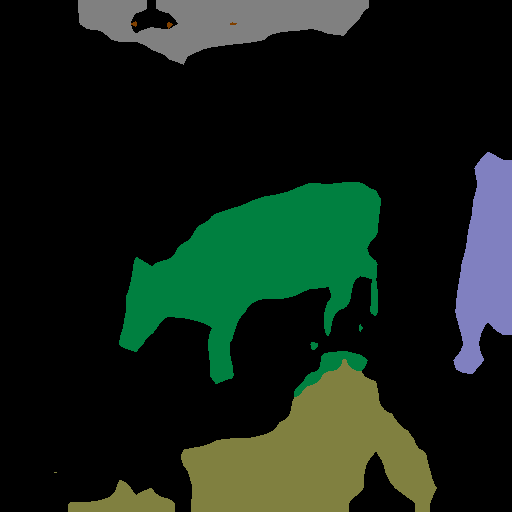}\\ 
        \includegraphics[width=0.993\textwidth,height=0.9in]{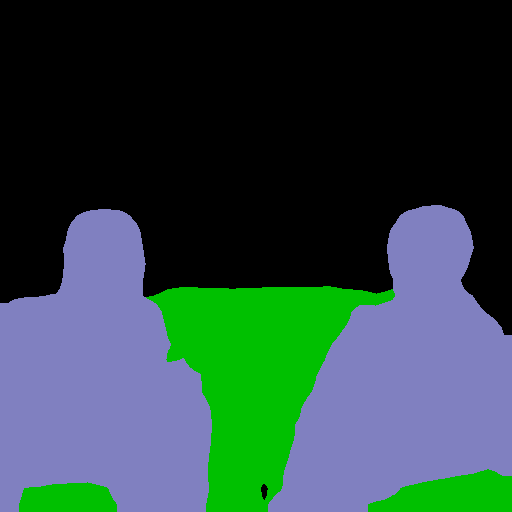}\\ 
        \vspace{2px}
    \end{minipage}%
}%
\subfigure[GT]{
    \begin{minipage}{0.14\linewidth}
        \centering
        \includegraphics[width=0.993\textwidth,height=0.9in]{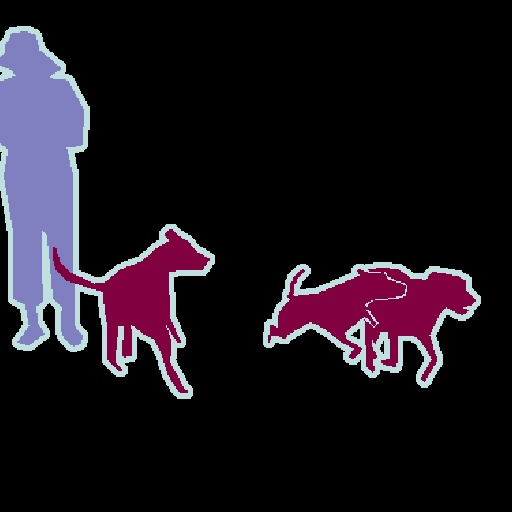}\\ 
        \includegraphics[width=0.993\textwidth,height=0.9in]{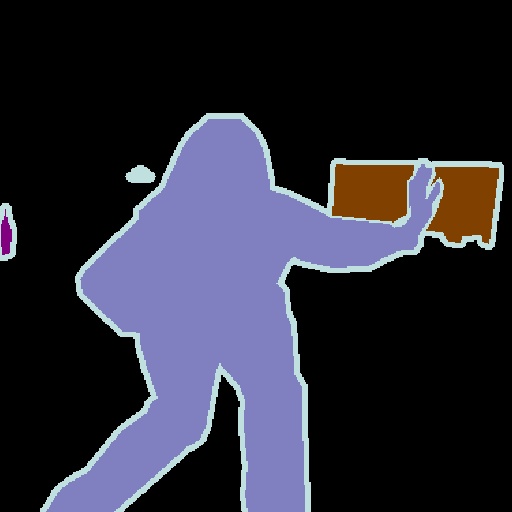}\\ 
        \includegraphics[width=0.993\textwidth,height=0.9in]{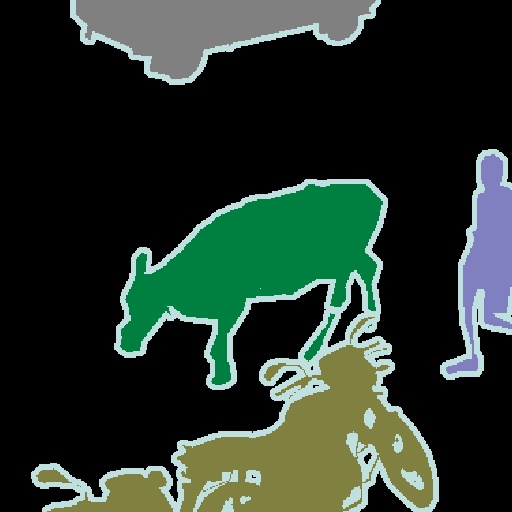}\\ 
        \includegraphics[width=0.993\textwidth,height=0.9in]{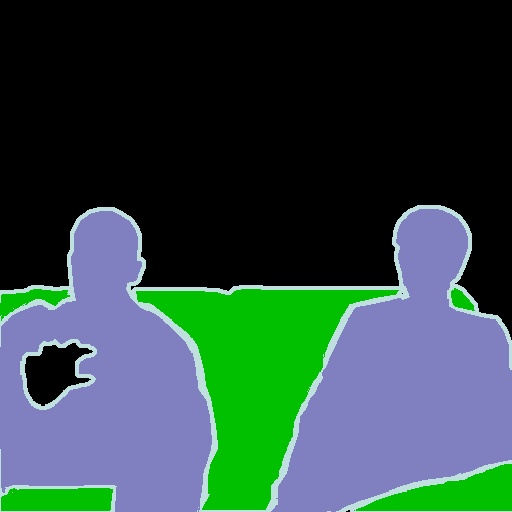}\\ 
        \vspace{2px}
    \end{minipage}%
}
\centering
\vspace{-5px}
\caption{Qualitative results on the VOC 2012 dataset (15-5 disjoint setting).}
\label{fig:vis_voc}
\end{figure*}

\begin{figure*}[htp]
    \centering
    \includegraphics[width=0.9\linewidth]{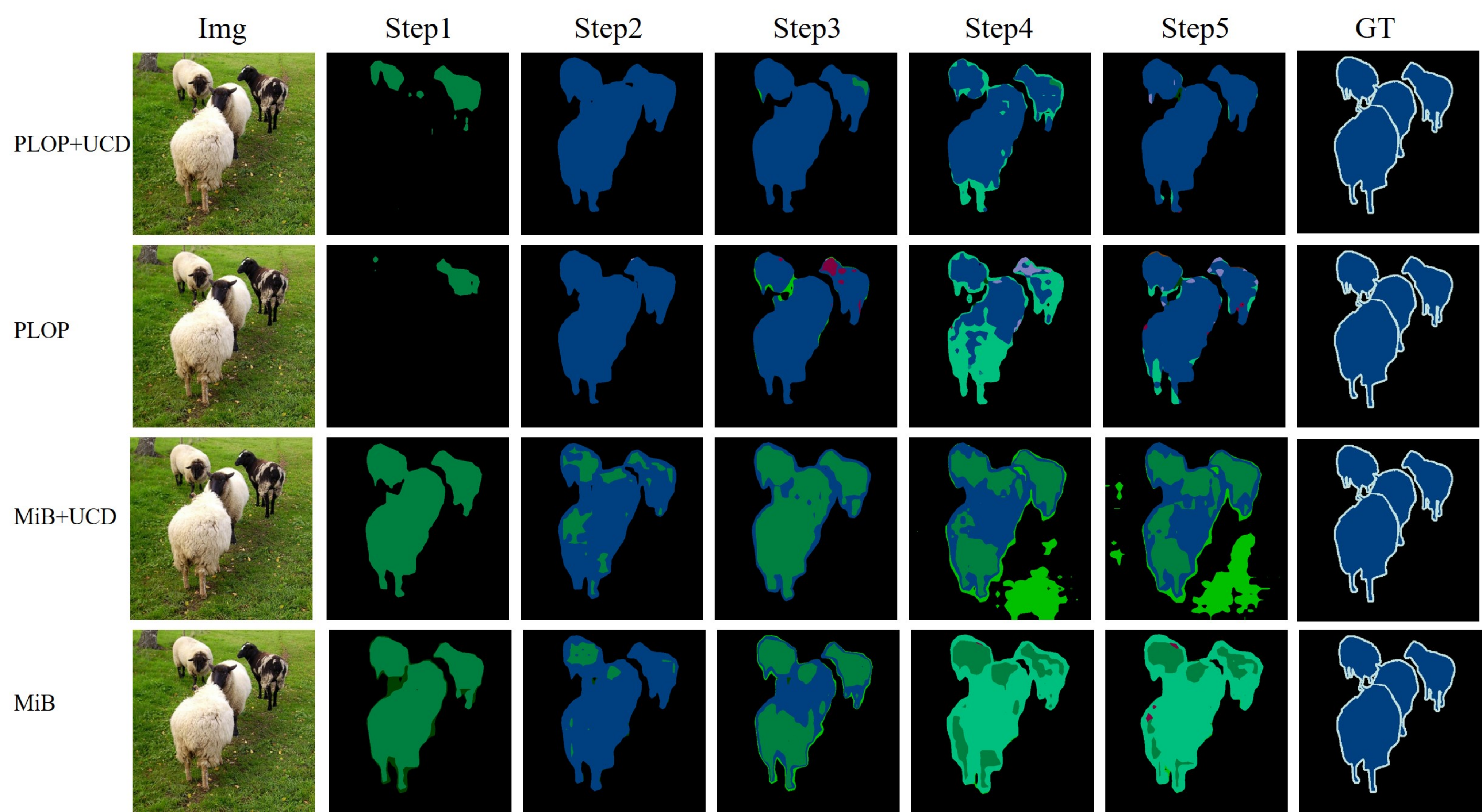}
    \caption{Qualitative results on the VOC 2012 dataset (15-1 overlapped setting) }
    \label{fig:vis 15-1}
\end{figure*}

\section{Experiments}

In this section we first introduce the datasets used for testing \method~against state of the art approaches. Our experimental results are then shown and illustrated. Additional ablation experiments are also provided, aiming to show the impact of every module included in the proposed architecture. 


\subsection{Datasets}
\par\noindent\textbf{PASCAL-VOC 2012.} 
The Pascal VOC2012 dataset~\cite{everingham2010pascal} contains 11,530 training/validation images of a variable size and with semantic labels correspondint to 20 classes plus background. Following previous works~\cite{michieli2019incremental,shmelkov2017incremental,cermelli2020modeling,tasar2019incremental}, we propose two different experimental settings: \textit{disjoint} and \textit{overlapped}. 
{In the first setting~\cite{michieli2019incremental}, each learning step contains a unique set of images, whose pixels belong to classes seen either in the current or in the previous learning steps. In other words, each image is seen in only one learning step.} On the contrary, in the \textit{overlapped} setting~\cite{shmelkov2017incremental}, a few classes appear on both tasks. Additionally, results are proposed in multiple experiments, using different class-incremental procedures: the first adds only one class during the new task (19-1), the second adds 5 classes at once (15-5) and the third adds 5 classes one at the time averaging the final results (15-1). Results are reported in terms of mIoU.

\par\noindent\textbf{ADE20K.} 
ADE20K~\cite{zhou2017scene} is a large-scale dataset including 150 classes. Unlike Pascal-VOC 2012 it comprises also non-object classes such as sky, ground, water and wall to name a few, resulting in a more fluid label set. The results are presented using \textit{disjoint} and \textit{overlapped} sets following the experimental setup proposed in~\cite{cermelli2020modeling,douillard2020plop}: addition of 50 classes in one single step (100-50), multi-step addition of 50 classes (100-10) and three steps of 50 classes each (50-50). Coherently with the VOC dataset, all classes that are not considered in the ground truth of the current task fall into the background macro-class. Results are presented using the mIoU metrics.

\par\noindent\textbf{Cityscapes.} The Cityscapes dataset~\cite{cordts2016cityscapes} has been built for Autonomous Driving applications. The release includes sets of images with different labeling accuracy, and since this is the first time the dataset has been used for Incremental Semantic Segmentation, we adopt the 3,475 finely annotated images, divided into 2,975/500 images for training and validation respectively. Following the setup of other datasets proposed in related works~\cite{cermelli2020modeling}, results are presented using the \textit{ovelapped} setting where tasks are arranged following two strategies: six classes added at once in the new task (13-6) and adding one of the 6 classes at a time and average the final result (13-1). As in previous cases the final results are reported in mIoU.

\subsection{Network Details}
{The proposed method is implemented in~PyTorch. The experiments are conducted on two Nvidia RTX 6000 GPUs during training while one for testing.}
The feature extractor used in this work includes a backbone network and a decoder. In detail, the ResNet-101~\cite{he2016deep} is chosen as backbone while features are decoded using the Deeplab-v3 architecture~\cite{chen2017rethinking}.
The weights are initialized on ImageNet~\cite{rota2018place} and trained using momentum and weight decay proposed in~\cite{chen2017rethinking}.  The initial learning rate is set to $10^{-2}$ for the first learning step, then it is diminished for the following steps to $10^{-3}$ for PASCAL-VOC 2012 and Cityscapes, while it is kept unchanged ($10^{-2}$) for ADE20K.
{We train the model with a batch-size of 24 for 30 epochs for Pascal-VOC 2012 and 60 epochs for ADE20K and Cityscapes in every learning step.}
The input image is cropped to meet the input dimension of $512 \times 512$ and transformed using the augmentation protocol proposed in~\cite{chen2017rethinking}. The temperature value of the contrastive loss has been set to $0.07$ (additional details are shown in Section~\ref{sec:ablation}), {while the weighting parameters $\lambda_{ucd}$, $\lambda_{pod}$and $\lambda_{kd}$ are set to $0.01$, $0.01$ and $10$ respectively.}


\subsection{Experimental Results}
Results are presented in this section, first in terms overall accuracy against the current state of the art (i.e. Tables~\ref{tab:overall voc}-\ref{tab:overall ade}-\ref{tab:overall city}, secondly with an extensive ablation study to weigh the importance of every single component proposed in this work. As further reference, two additional experiments are reported to serve as a lower and upper reference: in the first the new task is Fine Tuned (\textit{FT}) starting from the old model magnifying the catastrophic forgeting phenomena typical of neural networks in this setting, while the second is a joint training (\textit{Joint}) of both tasks, therefore breaking the incremental paradigm hypothesis.

\begin{table*}[h]
\caption{Mean IoU on the ADE20K dataset for different incremental class learning disjoint setting.$*$ means results come from re-implementation. UDC means uncertainty-aware contrastive distillation. Best among table in \textbf{bold}, best among part in \underline{underlined}.
}
\label{tab:overall ade}
\resizebox{\textwidth}{!}{%
\begin{tabular}{lcccccccccccccc}
\toprule
\multirow{2.5}{*}{Method} & \multicolumn{3}{c|}{100-50} & \multicolumn{7}{c|}{100-10} & \multicolumn{4}{c}{50-50} \\\cmidrule{2-15}
& 1-100 & \multicolumn{1}{c|}{101-150} & \multicolumn{1}{c|}{all} & 1-100 & 100-110 & 110-120 & 120-130 & 130-140 & \multicolumn{1}{c|}{140-150} & \multicolumn{1}{c|}{all} & 1-50 & 51-100 & \multicolumn{1}{c|}{101-150} & all \\
\midrule
\multicolumn{1}{l|}{FT} & 0.0 & \multicolumn{1}{c|}{24.9} & \multicolumn{1}{c|}{8.3} & 0.0 & 0.0 & 0.0 & 0.0 & 0.0 & \multicolumn{1}{c|}{16.6} & \multicolumn{1}{c|}{1.1} & 0.0 & 0.0 & \multicolumn{1}{c|}{22.0} & 7.3 \\
\multicolumn{1}{l|}{LwF \cite{li2017learning}} & 21.1 & \multicolumn{1}{c|}{25.6} & \multicolumn{1}{c|}{22.6} & 0.1 & 0.0 & 0.4 & 2.6 & 4.6 & \multicolumn{1}{c|}{\underline{16.9}} & \multicolumn{1}{c|}{1.7} & 5.7 & 12.9 & \multicolumn{1}{c|}{\underline{22.8}} & 13.9 \\
\multicolumn{1}{l|}{LwF-MC \cite{rebuffi2017icarl}} & 34.2 & \multicolumn{1}{c|}{10.5} & \multicolumn{1}{c|}{26.3} & 18.7 & 2.5 & 8.7 & 4.1 & 6.5 & \multicolumn{1}{c|}{5.1} & \multicolumn{1}{c|}{14.3} & 27.8 & 7.0 & \multicolumn{1}{c|}{10.4} & 15.1 \\
\multicolumn{1}{l|}{ILT \cite{michieli2019incremental}} & 22.9 & \multicolumn{1}{c|}{18.9} & \multicolumn{1}{c|}{21.6} & 0.3 & 0.0 & 1.0 & 2.1 & 4.6 & \multicolumn{1}{c|}{10.7} & \multicolumn{1}{c|}{1.4} & 8.4 & 9.7 & \multicolumn{1}{c|}{14.3} & 10.8 \\
\multicolumn{1}{l|}{{Inc.Seg \cite{yan2021framework}}} & 36.6 & \multicolumn{1}{c|}{0.4} & \multicolumn{1}{c|}{24.6} & \underline{32.4} & 0.0 & 0.2 & 0.0 & 0.0 & \multicolumn{1}{c|}{0.0} & \multicolumn{1}{c|}{21.7} & 40.2 & 1.3 & \multicolumn{1}{c|}{0.3} & 14.1 \\
\multicolumn{1}{l|}{SDR \cite{michieli2021continual}} & 37.4 & \multicolumn{1}{c|}{24.8} & \multicolumn{1}{c|}{33.2} & 28.9 & - & - & - & - & \multicolumn{1}{c|}{-} & \multicolumn{1}{c|}{21.7} & \underline{40.9} & - & \multicolumn{1}{c|}{-} & \underline{29.5} \\
\multicolumn{1}{l|}{{UCD}} & \underline{40.4} & \multicolumn{1}{c|}{\underline{27.3}} & \multicolumn{1}{c|}{\underline{36.0}} & 28.6 & \underline{13.0} & \underline{13.1} & \underline{9.2} & \underline{10.7} & \multicolumn{1}{c|}{16.1} & \multicolumn{1}{c|}{\underline{23.2}} & 39.3 & \underline{25.3} & \multicolumn{1}{c|}{19.1} & 27.9 \\
\midrule
\multicolumn{1}{l|}{MiB \cite{cermelli2020modeling}} & 37.9 & \multicolumn{1}{c|}{27.9} & \multicolumn{1}{c|}{34.6} & 31.8 & 10.4 & 14.8 & \textbf{\underline{12.8}} & \textbf{\underline{13.6}} & \multicolumn{1}{c|}{18.7} & \multicolumn{1}{c|}{25.9} & 35.5 & 22.2 & \multicolumn{1}{c|}{\textbf{\underline{23.6}}} & 27.0 \\
\multicolumn{1}{l|}{MiB + SDR \cite{michieli2021continual}} & 37.5 & \multicolumn{1}{c|}{25.5} & \multicolumn{1}{c|}{33.5} & 28.9 & - & - & - & - & \multicolumn{1}{c|}{-} & \multicolumn{1}{c|}{23.2} & \textbf{\underline{42.9}} & - & \multicolumn{1}{c|}{\textbf{-}} & 
\textbf{\underline{31.3}} \\
\multicolumn{1}{l|}{MiB + UCD} & \textbf{\underline{40.5}} & \multicolumn{1}{c|}{\textbf{\underline{28.1}}} & \multicolumn{1}{c|}{\textbf{\underline{36.4}} } & {\underline{33.4}} & \textbf{\underline{15.2}} & \textbf{\underline{15.3}} & 10.8 & 12.5 & \multicolumn{1}{c|}{\textbf{\underline{18.8}}} & \multicolumn{1}{c|}{\textbf{\underline{27.1}}} & {40.2} & \textbf{\underline{25.9}} & \multicolumn{1}{c|}{19.5} & {{28.5}} \\
\midrule
\multicolumn{1}{l|}{{PLOP$^*$}~\cite{douillard2020plop}} & 29.8 & \multicolumn{1}{c|}{4.2} & \multicolumn{1}{c|}{22.2 } & {32.1} &  1.9 & 10.0 & 0.8 & 1.2 & \multicolumn{1}{c|}{0.1} & \multicolumn{1}{c|}{22.3} & 19.2 & 0.4 & \multicolumn{1}{c|}{0.4} & 6.6 \\
\multicolumn{1}{l|}{{PLOP + UCD}} & \underline{33.2} & \multicolumn{1}{c|}{\underline{4.7}} & \multicolumn{1}{c|}{\underline{23.7}} & \textbf{\underline{35.7}} &  \underline{2.1} & \underline{11.1} & \underline{0.9} & \underline{1.4} & \multicolumn{1}{c|}{\underline{0.1}} & \multicolumn{1}{c|}{\underline{24.8}} & \underline{21.3} & \underline{0.4} & \multicolumn{1}{c|}{\underline{0.4}} & \underline{7.4} \\
\midrule
\multicolumn{1}{l|}{Joint} & 44.3 & \multicolumn{1}{c|}{28.2} & \multicolumn{1}{c|}{38.9} & 44.3 & 26.1 & 42.8 & 26.7 & 28.1 & \multicolumn{1}{c|}{17.3} & \multicolumn{1}{c|}{38.9} & 51.1 & 38.3 & \multicolumn{1}{c|}{28.2} & 38.9\\
\bottomrule
\end{tabular}%
}
\end{table*}

\begin{table*}[h]
\caption{Mean IoU on the ADE20K dataset for different incremental class learning overlapped setting. $*$ means results come from PLOP. UDC means uncertainty-aware contrastive distillation. Best among table in \textbf{bold}.
}
\label{tab:overall ade more}
\centering
\begin{tabular}{lccccccccc}
\toprule
\multirow{2}{*}{method}       & \multicolumn{3}{c}{100-50} & \multicolumn{3}{c}{100-10} & \multicolumn{3}{c}{50-50} \\  \cmidrule(lr){2-4} \cmidrule(lr){5-7} \cmidrule(lr){8-10}
 & 1-100  & 101-150  & mIoU   & 1-100  & 101-150  & mIoU   & 1-50   & 51-150  & mIoU   \\
\midrule                              
\multicolumn{1}{l|}{ILT$^*$\cite{michieli2019incremental}}      & 18.29  & 14.40    & 17.00  & 0.11   & 3.06     & 1.09   & 3.53   & 12.85   & 9.70   \\
\multicolumn{1}{l|}{MiB$^*$\cite{cermelli2020modeling}}      & 40.52  & 17.17    & 32.79  & 38.21  & 11.12    & 29.24  & 45.57  & 21.01   & 29.31  \\
\multicolumn{1}{l|}{PLOP~\cite{douillard2020plop}}     & 41.87  & 14.89    & 32.94  & 40.48  & 13.61    & 31.59  & \textbf{48.83}  & 20.99   & 30.40  \\
\multicolumn{1}{l|}{PLOP+UCD} & \textbf{42.12}   & \textbf{15.84}    & \textbf{33.31}  & \textbf{40.80}  & \textbf{15.23 }   & \textbf{32.29}  & 47.12  & \textbf{24.12}   & \textbf{31.79} \\
\bottomrule
\end{tabular}
\end{table*}

\begin{table}[]
\caption{Mean IoU on the Cityscapes dataset for different incremental class learning scenarios. UDC means uncertainty-aware contrastive distillation. Best among table in \textbf{bold}, best among part in \underline{underlined}.}
\label{tab:overall city}

\centering
\begin{tabular}{lcccccc}
\toprule
\multirow{2}{*}{method} & \multicolumn{3}{c}{13-6} & \multicolumn{3}{c}{13-1} \\ \cmidrule(lr){2-4} \cmidrule(lr){5-7} 
& 1-13   & 14-19   & mIoU  & 1-13   & 14-19   & mIoU   \\
\midrule               
\multicolumn{1}{l|}{FT} & 0.0 & \textbf{33.4} & 10.6 & 0.0 & 7.4 & 2.3\\
\multicolumn{1}{l|}{ILT~\cite{michieli2019incremental}} & 52.3 & 18.3 & 41.5 & 4.2 & 9.1 & 5.8\\
\midrule
\multicolumn{1}{l|}{PLOP~\cite{douillard2020plop}} & 53.2 & \underline{10.1} & 39.6 & 52.4 & 15.1 & 40.6\\
\multicolumn{1}{l|}{PLOP+UCD} & \textbf{\underline{53.4}} & 10.0 & \underline{39.7} & \textbf{\underline{52.7}} & \underline{15.3} & \underline{40.9}\\
\midrule
\multicolumn{1}{l|}{MiB~\cite{cermelli2020modeling}} & 52.8 & 17.9 & 41.8 & 51.6 & 22.9 & 42.5\\
\multicolumn{1}{l|}{MiB+UCD} & \underline{53.0} & \underline{{18.6}} & \textbf{\underline{42.1}} & \underline{52.2} & \textbf{\underline{23.4}} & \textbf{\underline{43.1}}\\
\midrule
\multicolumn{1}{l|}{Joint} & 53.5 & 54.2 & 53.7 & 53.5 & 54.2 & 53.7\\
\bottomrule
\end{tabular}
\end{table}

\subsubsection{Comparison with state of the art methods}
\par\noindent\textbf{Comparison on VOC 2012.}
Table~\ref{tab:overall voc} shows the performance of \method~with respect to competing method for class-incremental learning including PI~\cite{zenke2017continual}, EWC~\cite{kirkpatrick2017overcoming}, RW~\cite{chaudhry2018riemannian}, LwF~\cite{li2017learning}, LwF-MC~\cite{rebuffi2017icarl} and works specially designed for image segmentation such as ILT~\cite{michieli2019incremental}, MiB~\cite{cermelli2020modeling}, PLOP~\cite{douillard2020plop} and  SDR~\cite{michieli2021continual}. As mentioned earlier, fine tuning and joint training are also reported as a lower and upper bound reference.

\begin{figure*}[t]
\centering
\subfigure[Image]{
    \begin{minipage}{0.14\linewidth}
        \centering
        \includegraphics[width=0.993\textwidth,height=0.9in]{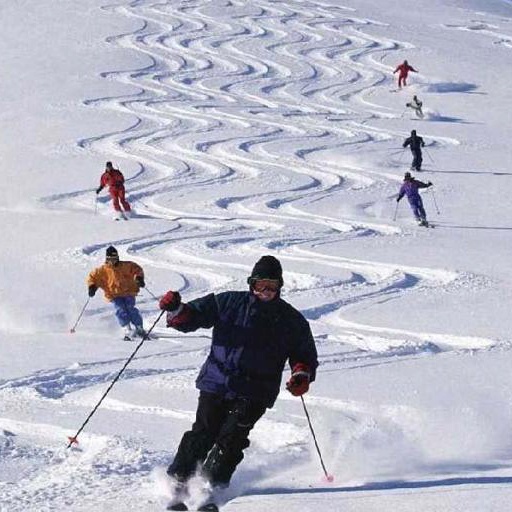}\\        
        \includegraphics[width=0.993\textwidth,height=0.9in]{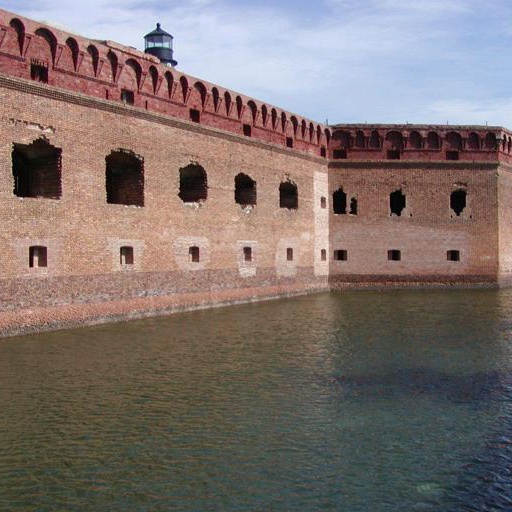}\\  
        \includegraphics[width=0.993\textwidth,height=0.9in]{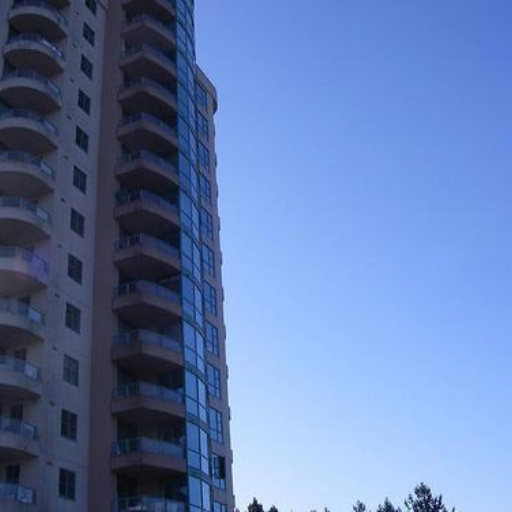}\\
        \includegraphics[width=0.993\textwidth,height=0.9in]{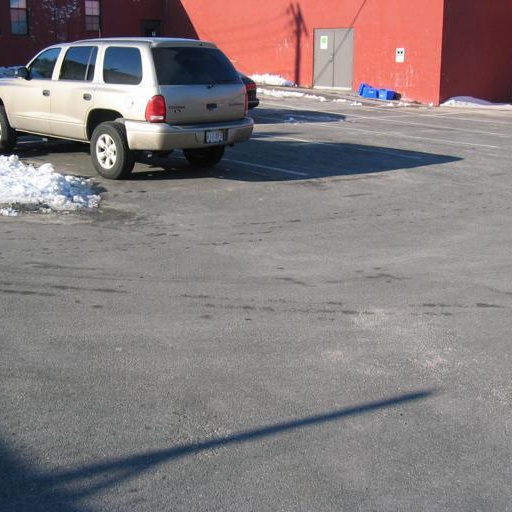}\\
        \includegraphics[width=0.993\textwidth,height=0.9in]{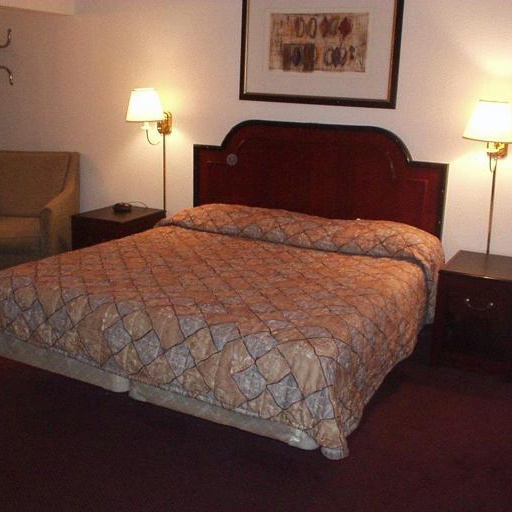}\\
        \vspace{2px}
    \end{minipage}%
}%
\subfigure[ILT~\cite{michieli2019incremental}]{
    \begin{minipage}{0.14\linewidth}
        \centering
        \includegraphics[width=0.993\textwidth,height=0.9in]{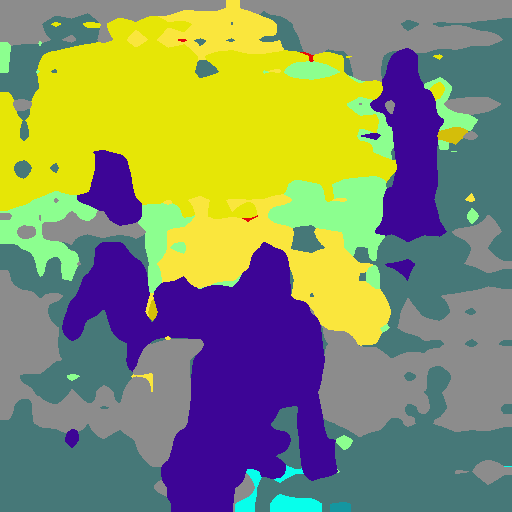}\\        
        \includegraphics[width=0.993\textwidth,height=0.9in]{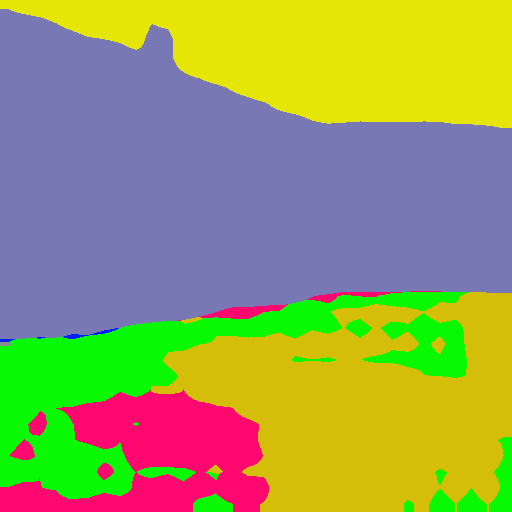}\\  
        \includegraphics[width=0.993\textwidth,height=0.9in]{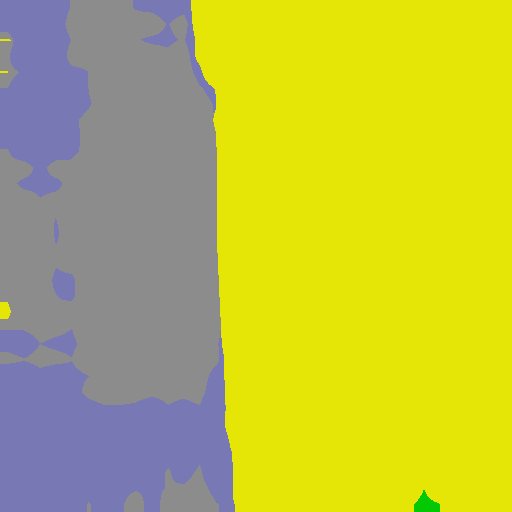}\\
        \includegraphics[width=0.993\textwidth,height=0.9in]{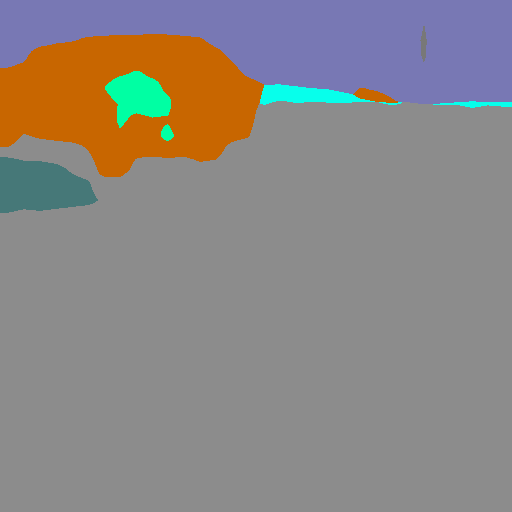}\\
        \includegraphics[width=0.993\textwidth,height=0.9in]{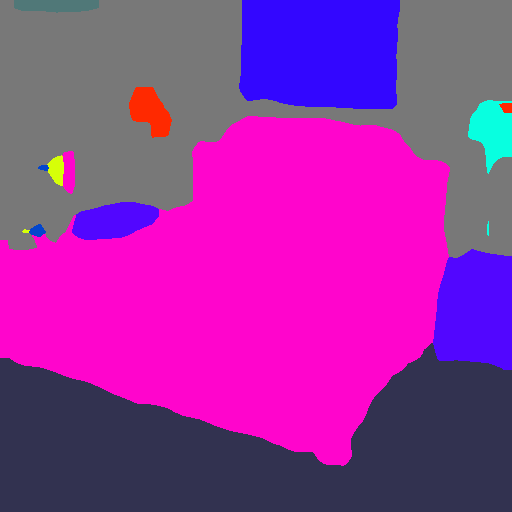}\\
        \vspace{2px}
    \end{minipage}%
}%
\subfigure[PLOP~\cite{douillard2020plop}]{
    \begin{minipage}{0.14\linewidth}
        \centering
        \includegraphics[width=0.993\textwidth,height=0.9in]{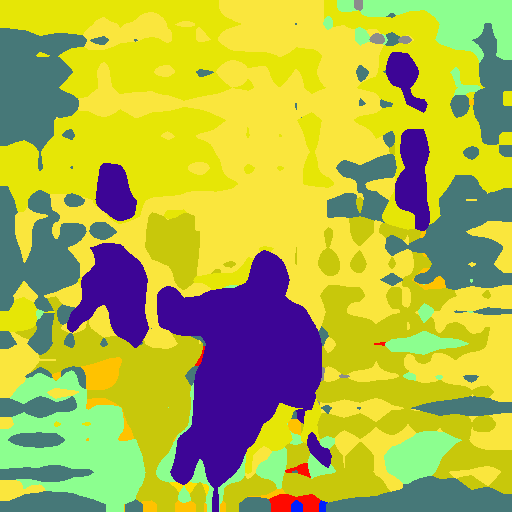}\\        
        \includegraphics[width=0.993\textwidth,height=0.9in]{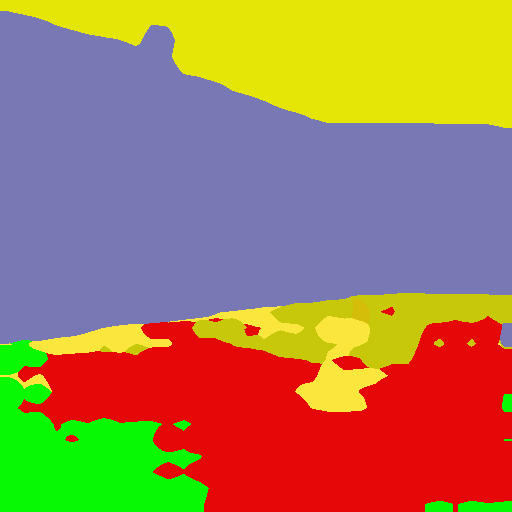}\\  
        \includegraphics[width=0.993\textwidth,height=0.9in]{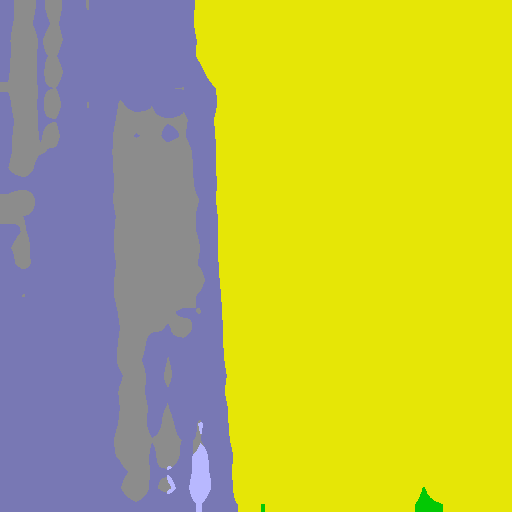}\\
        \includegraphics[width=0.993\textwidth,height=0.9in]{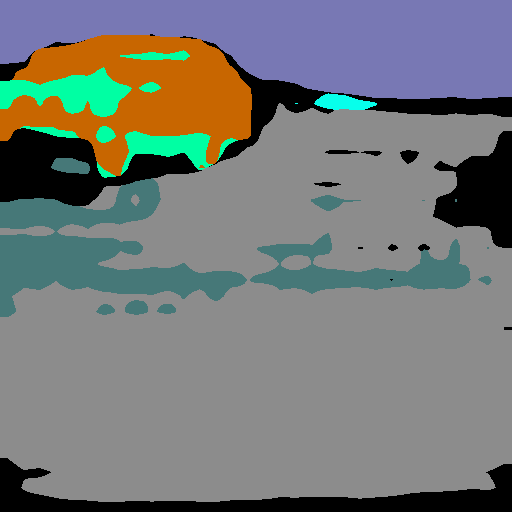}\\
        \includegraphics[width=0.993\textwidth,height=0.9in]{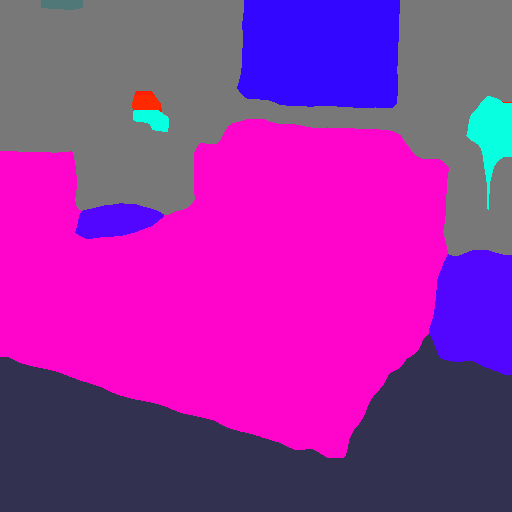}\\
        \vspace{2px}
    \end{minipage}%
}%
\subfigure[PLOP+UCD]{
    \begin{minipage}{0.14\linewidth}
        \centering
        \includegraphics[width=0.993\textwidth,height=0.9in]{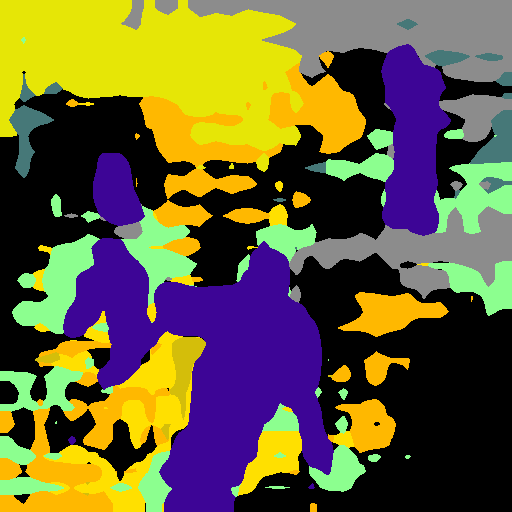}\\        
        \includegraphics[width=0.993\textwidth,height=0.9in]{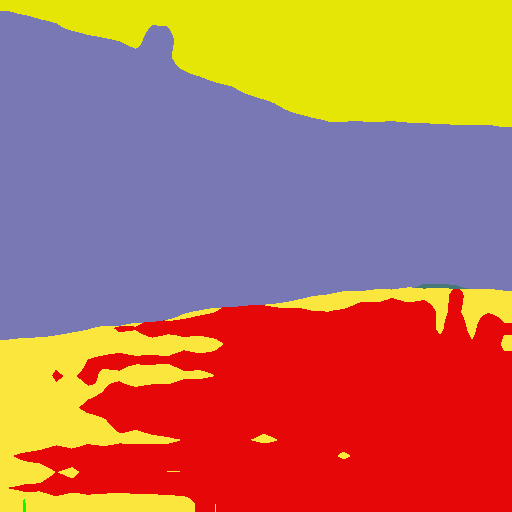}\\  
        \includegraphics[width=0.993\textwidth,height=0.9in]{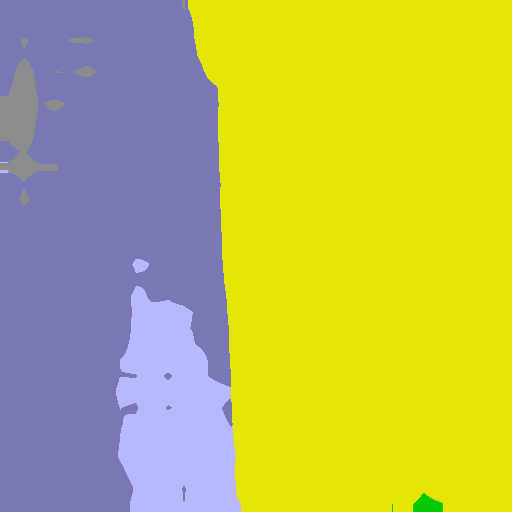}\\
        \includegraphics[width=0.993\textwidth,height=0.9in]{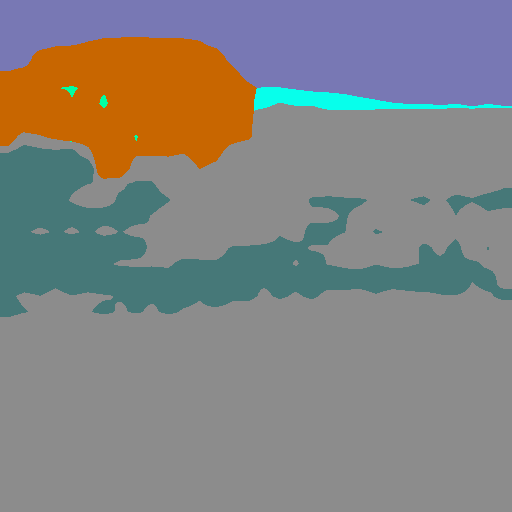}\\
        \includegraphics[width=0.993\textwidth,height=0.9in]{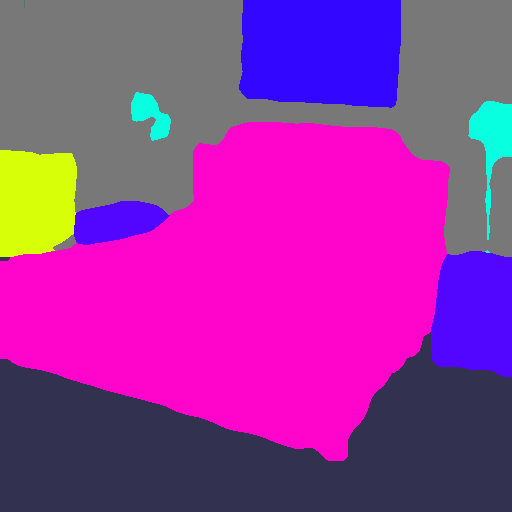}\\
        \vspace{2px}
    \end{minipage}%
}%
\subfigure[MiB~\cite{cermelli2020modeling}]{
    \begin{minipage}{0.14\linewidth}
        \centering
        \includegraphics[width=0.993\textwidth,height=0.9in]{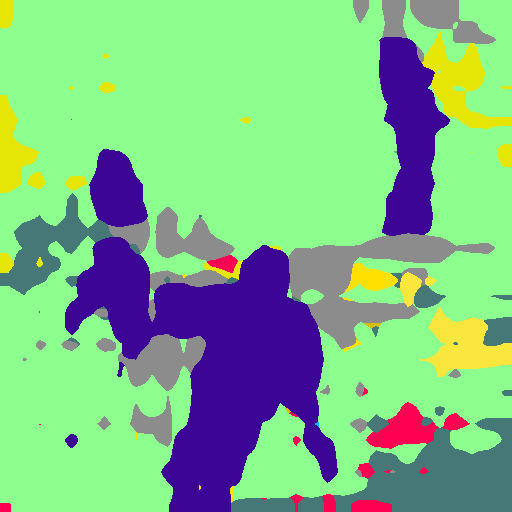}\\        
        \includegraphics[width=0.993\textwidth,height=0.9in]{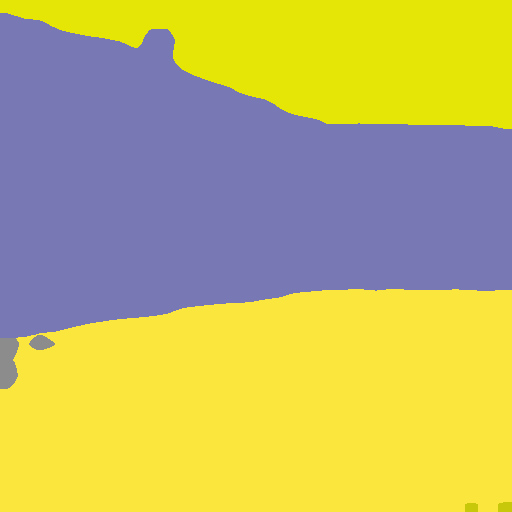}\\  
        \includegraphics[width=0.993\textwidth,height=0.9in]{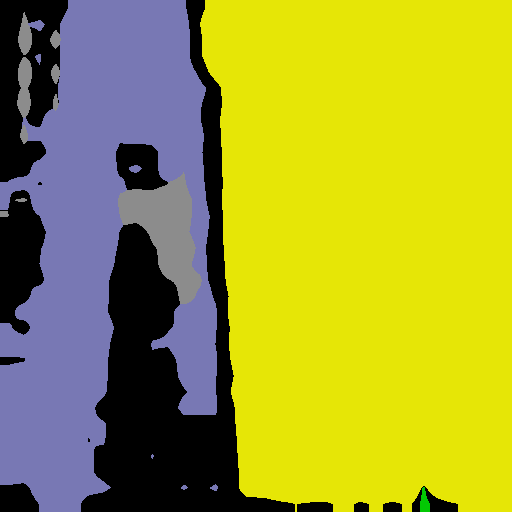}\\
        \includegraphics[width=0.993\textwidth,height=0.9in]{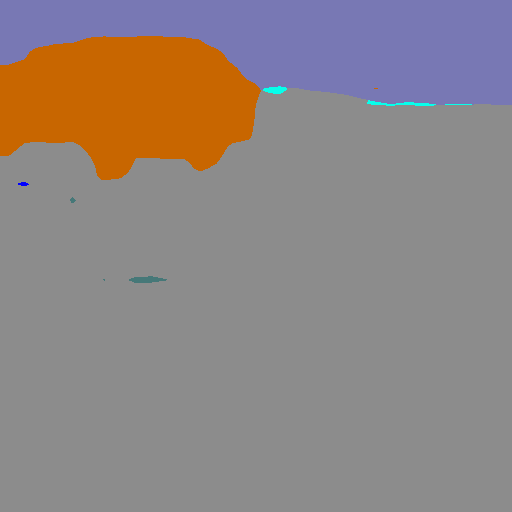}\\
        \includegraphics[width=0.993\textwidth,height=0.9in]{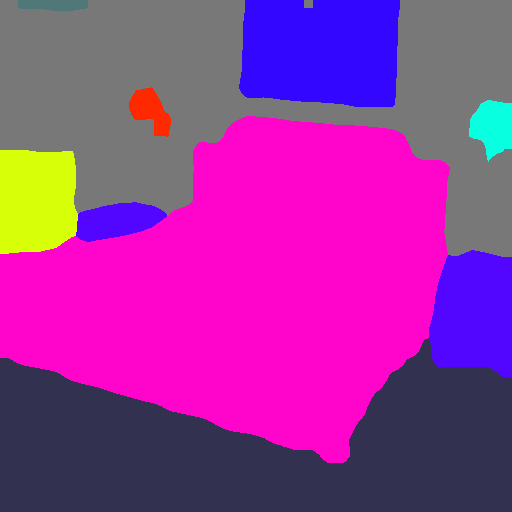}\\
        \vspace{2px}
    \end{minipage}%
}%
\subfigure[MiB+UCD]{
    \begin{minipage}{0.14\linewidth}
        \centering
        \includegraphics[width=0.993\textwidth,height=0.9in]{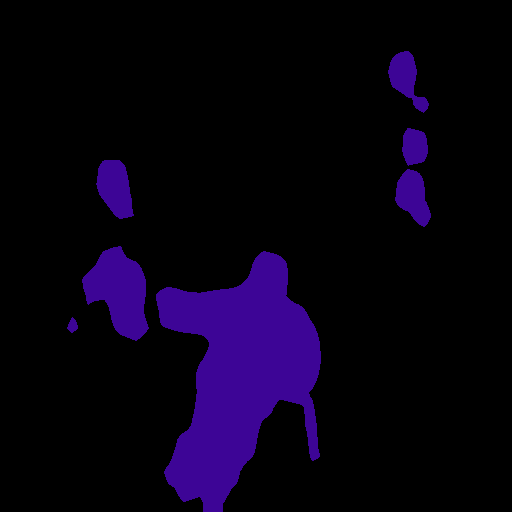}\\        
        \includegraphics[width=0.993\textwidth,height=0.9in]{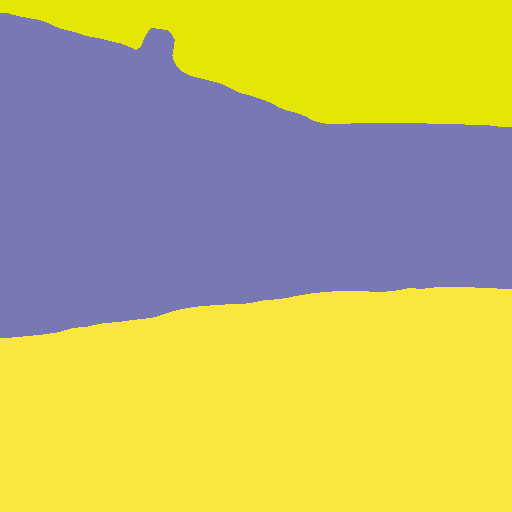}\\  
        \includegraphics[width=0.993\textwidth,height=0.9in]{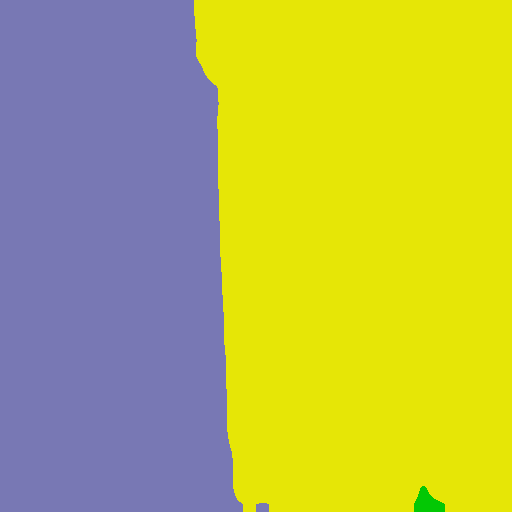}\\
        \includegraphics[width=0.993\textwidth,height=0.9in]{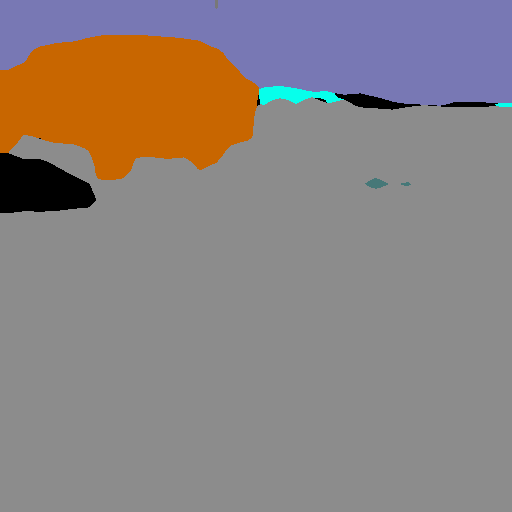}\\
        \includegraphics[width=0.993\textwidth,height=0.9in]{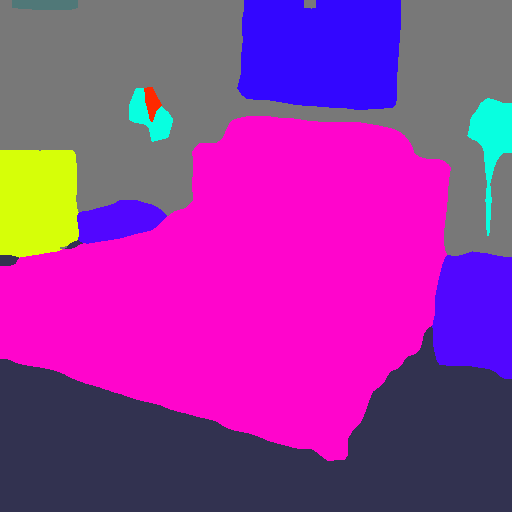}\\
        \vspace{2px}
    \end{minipage}%
}%
\subfigure[GT]{
    \begin{minipage}{0.14\linewidth}
        \centering
        \includegraphics[width=0.993\textwidth,height=0.9in]{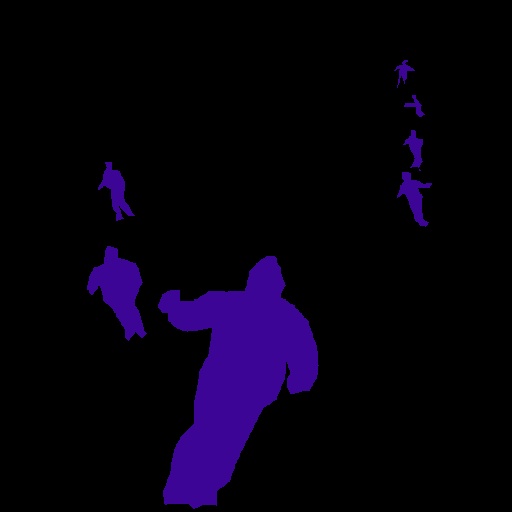}\\        
        \includegraphics[width=0.993\textwidth,height=0.9in]{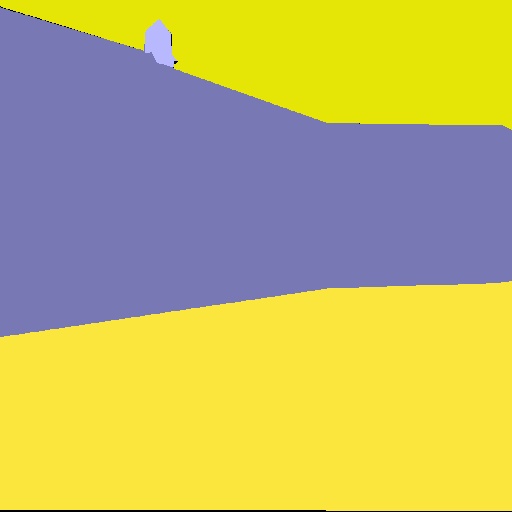}\\  
        \includegraphics[width=0.993\textwidth,height=0.9in]{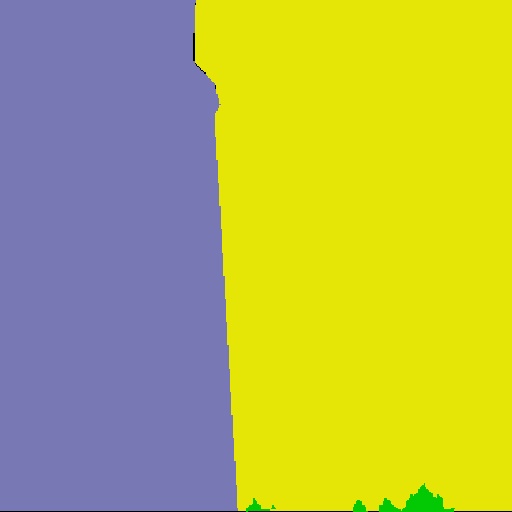}\\
        \includegraphics[width=0.993\textwidth,height=0.9in]{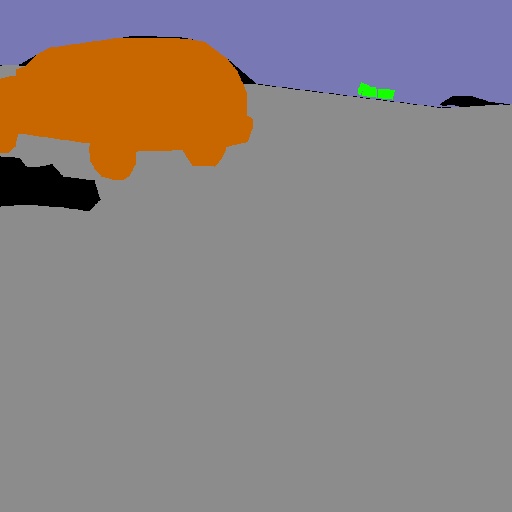}\\
        \includegraphics[width=0.993\textwidth,height=0.9in]{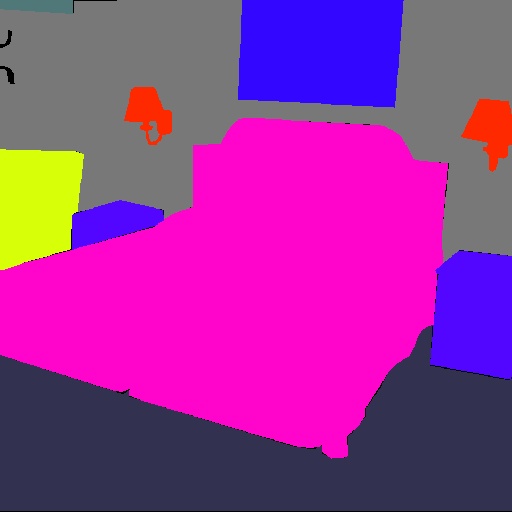}\\
        \vspace{2px}
    \end{minipage}%
}
\centering
\caption{Qualitative results on the ADE 20K dataset (100-50 disjoint setting).}
\label{fig:vis_ade}

\end{figure*}

Specifically, PLOP has been modified in order to not consider the background class in the accuracy computation for sake of fairness in the results comparison with the competing works. The training model proposed in this work (i.e. UCD) can be applied on top of other methods therefore Table~\ref{tab:overall voc} shows results of the best performing incremental learning algorithms for semantic segmentation evaluated using the UCD procedure. Results show that UCD generally improves the performances of other standalone models (i.e. MiB and PLOP), in both \textit{overlapped} and \textit{disjoint} settings. 
{However, UCD has a weaker performance than MiB+SDR~\cite{michieli2021continual},mainly in disjoint setup of VOC 2012. The first reason is that SDR uses attraction and repulsion losses with prototypes (calculated with a running average), which is fundamentally different from the contrastive loss. Averaging features into prototypes changes balancing between the classes in the loss. Probably, this makes results of old classes perform better because old classes already account for the majority of the total classes. Another reason is that in some cases it is difficult for the network to generate good extended semantic maps in the disjoint setting. This probably affects UCD more than SDR, because SDR counteracts this noise by using running averages and has additional regularization mechanisms.}
Another can notice that neither of the two methods mentioned above is outperforming the other in all settings, while adding UCD is generally beneficial. 
{Moreover, we also compare our with the SOTA on more incremental class learning scenarios in Table~\ref{tab:more voc}. For more challenging settings (10-10 and 10-1), UCD surpasses MiB+SDR and PLOP. This reinforces the idea that UCD tends to perform better when the number of old classes and new classes are more balanced.}
Qualitative results are shown in Figure~\ref{fig:vis_voc} and Figure~\ref{fig:vis 15-1} for 15-5 and 15-1 scenarios respectively. Note that classes such as human, bike, and cow are included in the old task but are still well represented in the resulting segmentation map when UCD is involved. {However, the improvements brought by UCD is larger for MiB than PLOP, even in thecases where MiB was better than PLOP. One reason is, in most of the settings, PLOP already achieves very good results, outperforming MiB by a large margins. This makes it hard for UCD to produce an extra improvement over PLOP. In addition, it seems PLOP is not well suited for the disjoint setting and therefore makes UCD less effective as well. Finally, since PLOP already has a feature distillation mechanism, it might be that the two feature distillation losses take care of similar aspects of knowledge transfer.  }

\par\noindent\textbf{Comparison on ADE20K.}
Quantitative results on ADE20K are reported in Table~\ref{tab:overall ade} and Table~\ref{tab:overall ade more}. Following the previous works \cite{cermelli2020modeling,douillard2020plop,iscen2020memory} the comparison is performed under three different settings differentiating mainly in how the new task data is handled: 100 classes in the old task and 50 classes at once for the new one (100-50), same number of old classes and 50 classes in batches of 10 for 5 times (100-10), an old task of 50 classes and then two batches of 50 classes each as new task (50-50).  
In this dataset UCD still demonstrates to improve results of base methods on the majority of the scenarios. In the specific the classes between 120-140 are characterized by small objects such as light, food, pots, dishes, hood and vase,
resulting in an extremely localized pixel contrastive that is probably not as efficient as for other sets of classes. Figure~\ref{fig:vis_ade} presents some qualitative results for the 100-50 disjoint setting. Large classes of the old task such as sky, building and car are well recognized with the UCD method while smaller objects are penalized by the smaller set of points to perform contrastive distillation.


\par
\begin{figure*}[t]
\centering
\subfigure[Image]{
    \begin{minipage}{0.14\linewidth}
        \centering
        \includegraphics[width=0.993\textwidth,height=0.5in]{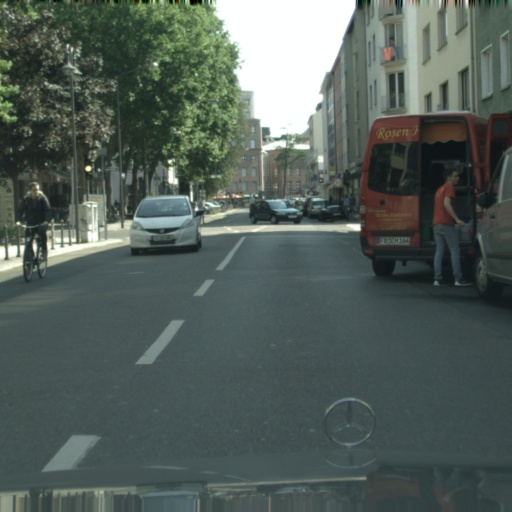}\\ 
        \includegraphics[width=0.993\textwidth,height=0.5in]{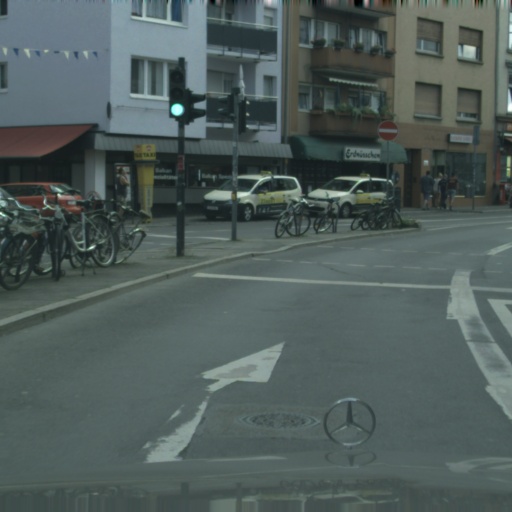}\\ 
        \includegraphics[width=0.993\textwidth,height=0.5in]{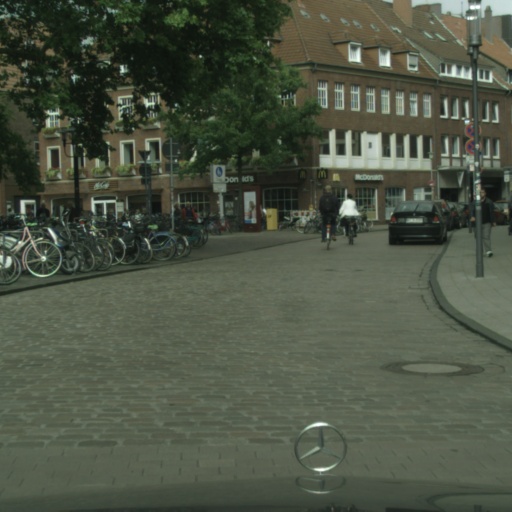}\\ 
        \includegraphics[width=0.993\textwidth,height=0.5in]{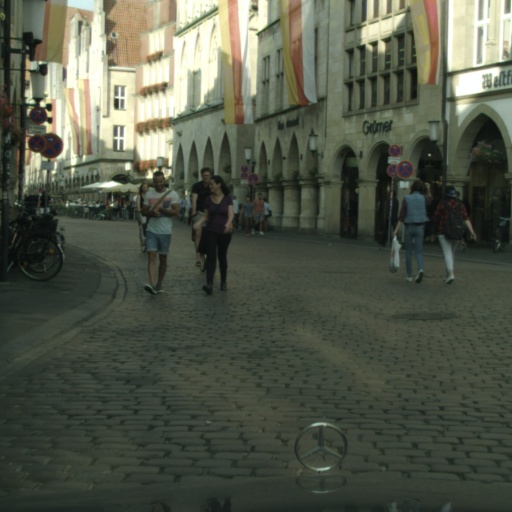}\\ 
        \includegraphics[width=0.993\textwidth,height=0.5in]{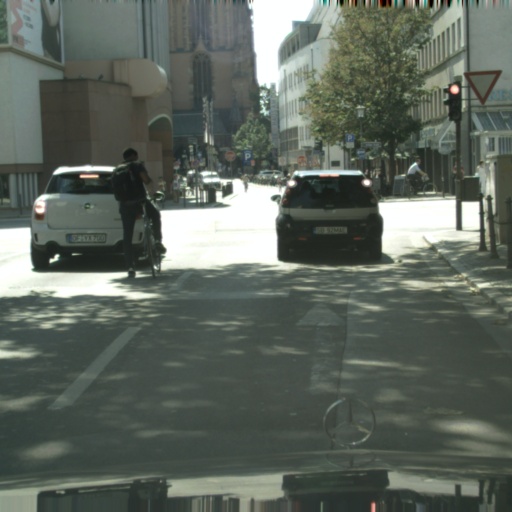}\\ 
        \vspace{2px}
    \end{minipage}%
}%
\subfigure[ILT~\cite{michieli2019incremental}]{
    \begin{minipage}{0.14\linewidth}
        \centering
        \includegraphics[width=0.993\textwidth,height=0.5in]{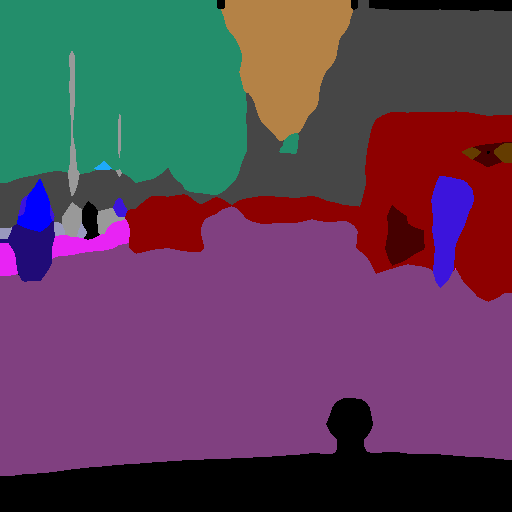}\\ 
        \includegraphics[width=0.993\textwidth,height=0.5in]{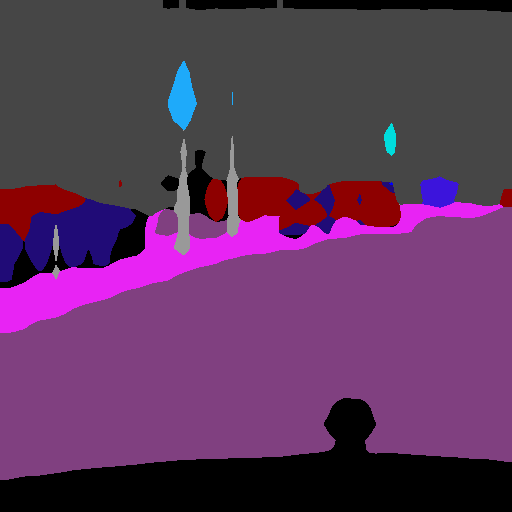}\\ 
        \includegraphics[width=0.993\textwidth,height=0.5in]{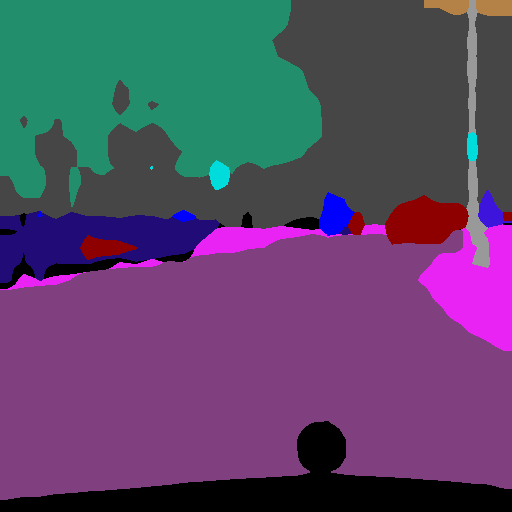}\\ 
        \includegraphics[width=0.993\textwidth,height=0.5in]{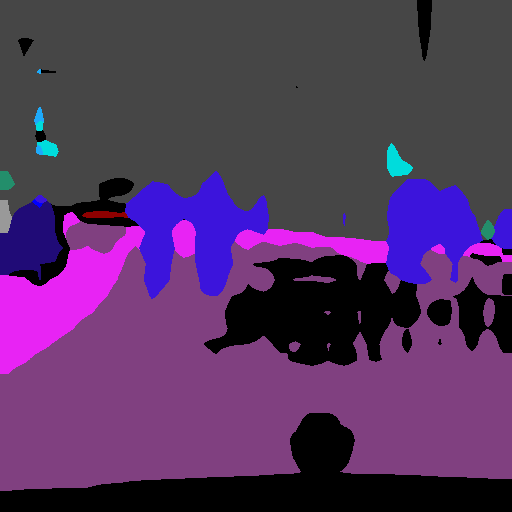}\\ 
        \includegraphics[width=0.993\textwidth,height=0.5in]{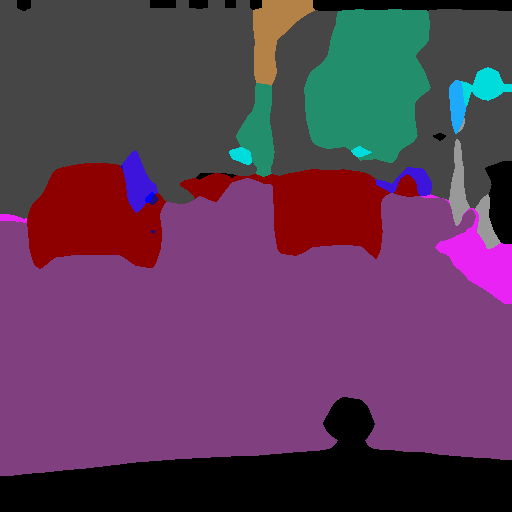}\\ 
        \vspace{2px}
    \end{minipage}%
}%
\subfigure[PLOP~\cite{douillard2020plop}]{
    \begin{minipage}{0.14\linewidth}
        \centering
        \includegraphics[width=0.993\textwidth,height=0.5in]{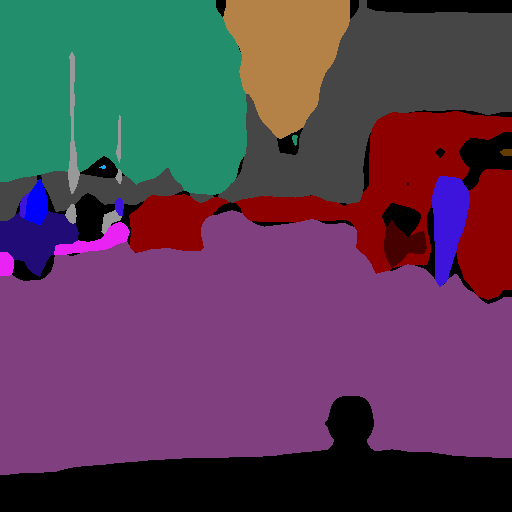}\\ 
        \includegraphics[width=0.993\textwidth,height=0.5in]{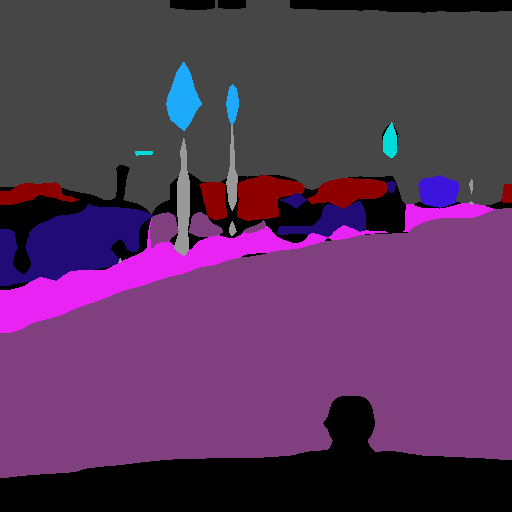}\\ 
        \includegraphics[width=0.993\textwidth,height=0.5in]{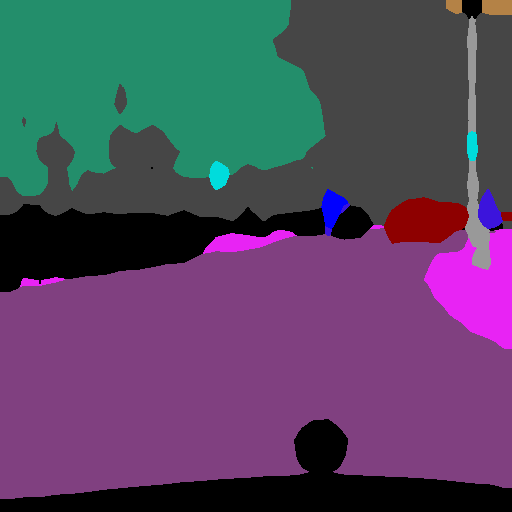}\\ 
        \includegraphics[width=0.993\textwidth,height=0.5in]{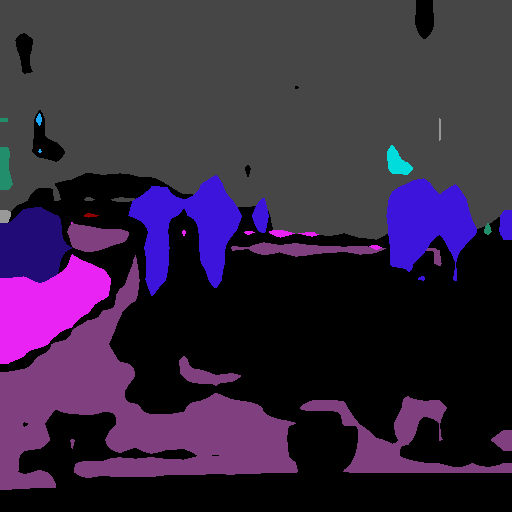}\\ 
        \includegraphics[width=0.993\textwidth,height=0.5in]{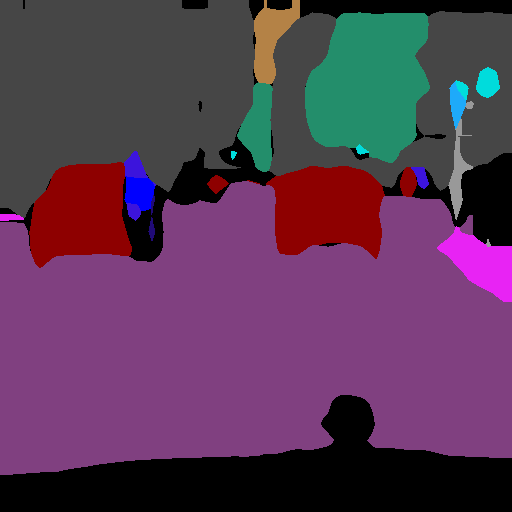}\\ 
        \vspace{2px}
    \end{minipage}%
}%
\subfigure[PLOP+UCD]{
    \begin{minipage}{0.14\linewidth}
        \centering
        \includegraphics[width=0.993\textwidth,height=0.5in]{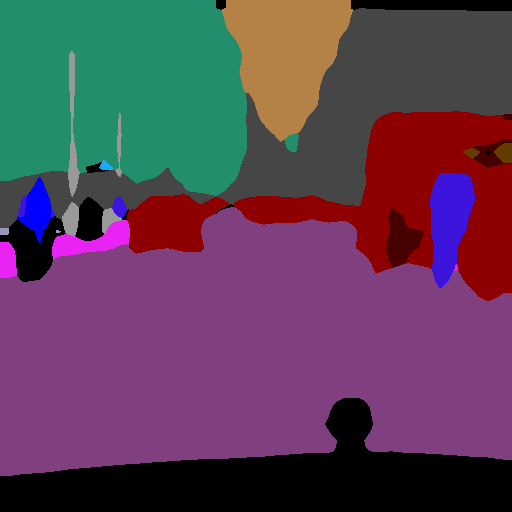}\\ 
        \includegraphics[width=0.993\textwidth,height=0.5in]{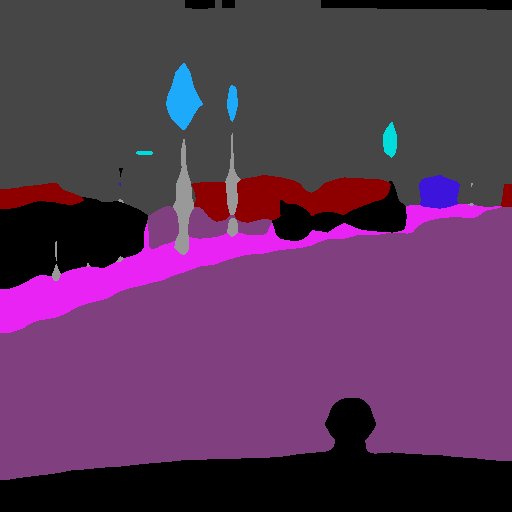}\\ 
        \includegraphics[width=0.993\textwidth,height=0.5in]{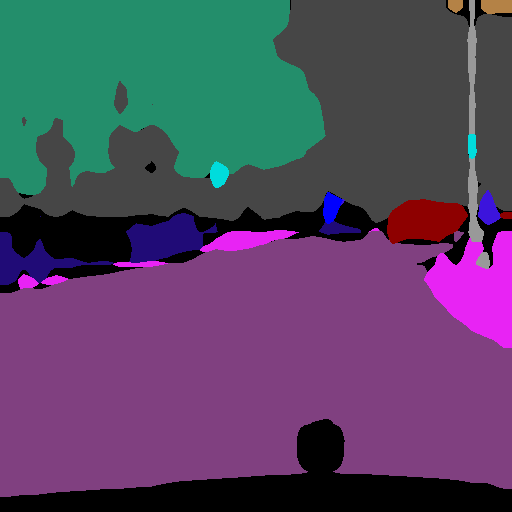}\\ 
        \includegraphics[width=0.993\textwidth,height=0.5in]{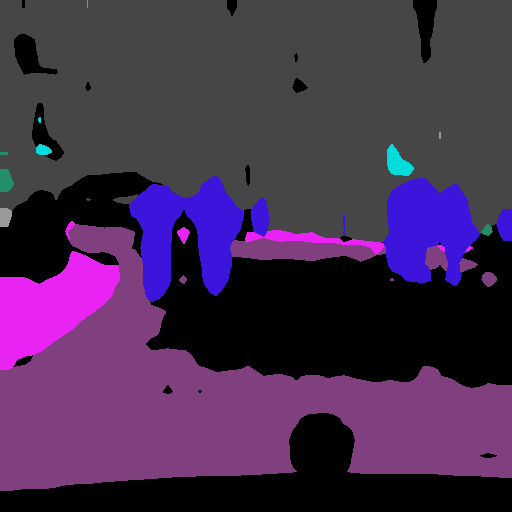}\\ 
        \includegraphics[width=0.993\textwidth,height=0.5in]{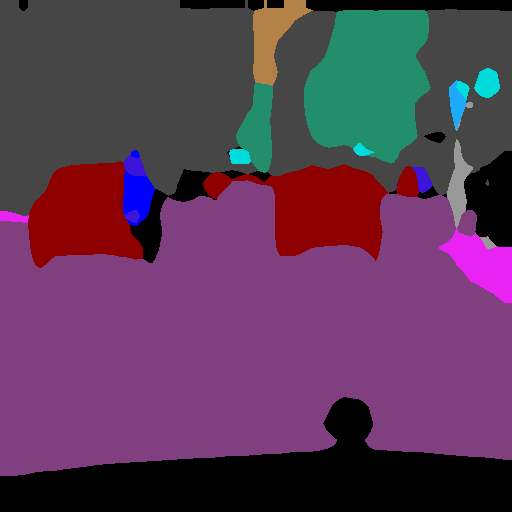}\\ 
        \vspace{2px}
    \end{minipage}%
}%
\subfigure[MiB~\cite{cermelli2020modeling}]{
    \begin{minipage}{0.14\linewidth}
        \centering
        \includegraphics[width=0.993\textwidth,height=0.5in]{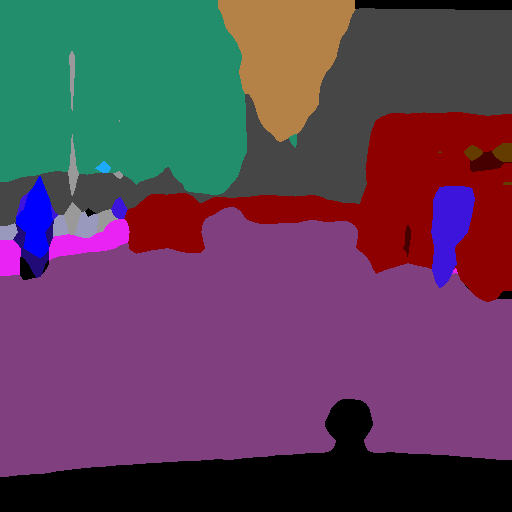}\\ 
        \includegraphics[width=0.993\textwidth,height=0.5in]{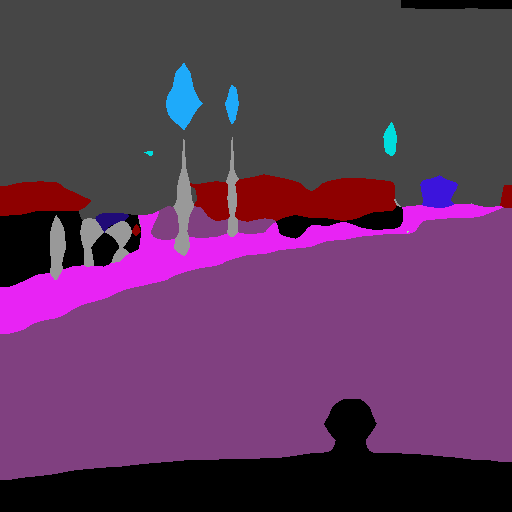}\\ 
        \includegraphics[width=0.993\textwidth,height=0.5in]{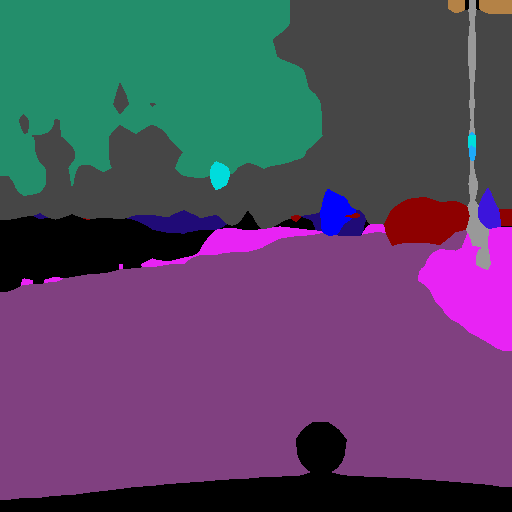}\\ 
        \includegraphics[width=0.993\textwidth,height=0.5in]{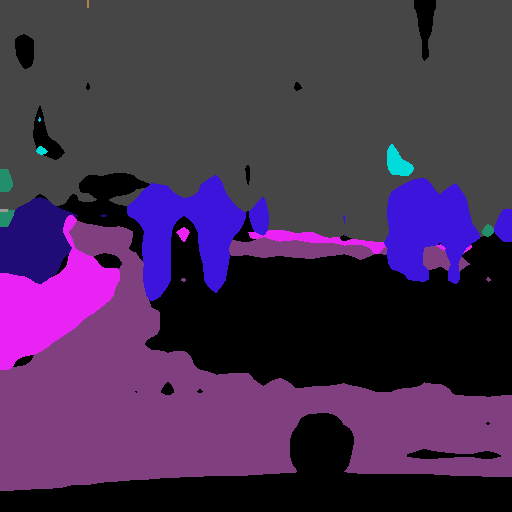}\\ 
        \includegraphics[width=0.993\textwidth,height=0.5in]{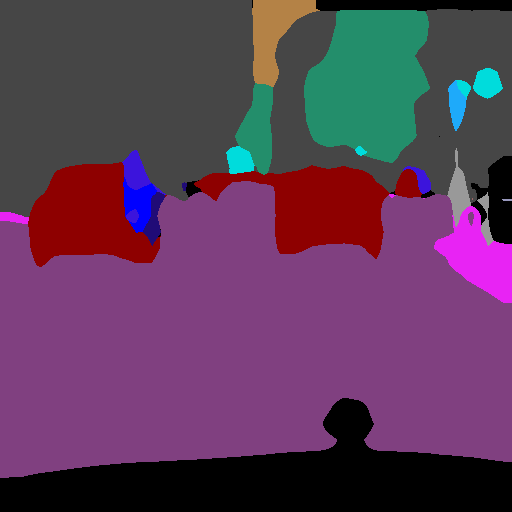}\\ 
        \vspace{2px}
    \end{minipage}%
}%
\subfigure[MiB+UCD]{
    \begin{minipage}{0.14\linewidth}
        \centering
        \includegraphics[width=0.993\textwidth,height=0.5in]{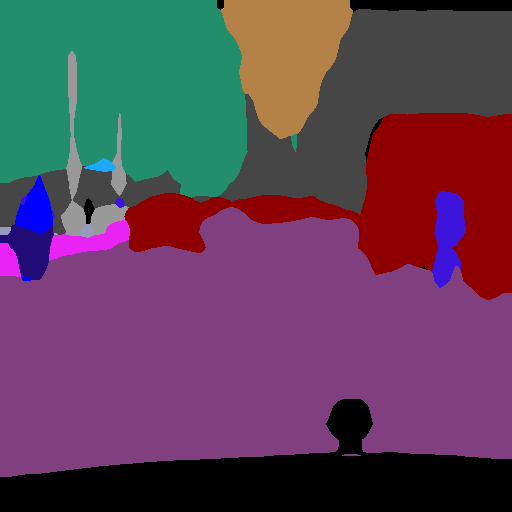}\\ 
        \includegraphics[width=0.993\textwidth,height=0.5in]{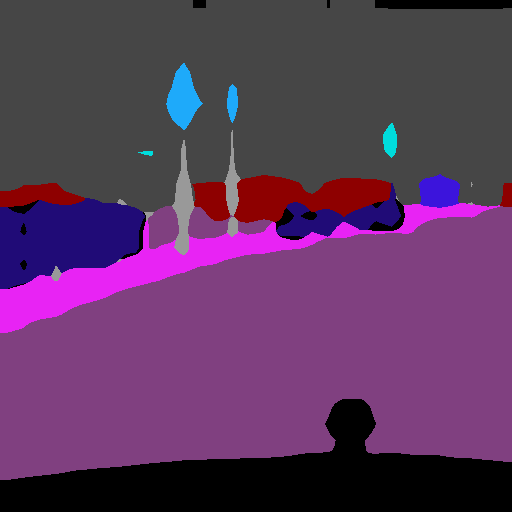}\\ 
        \includegraphics[width=0.993\textwidth,height=0.5in]{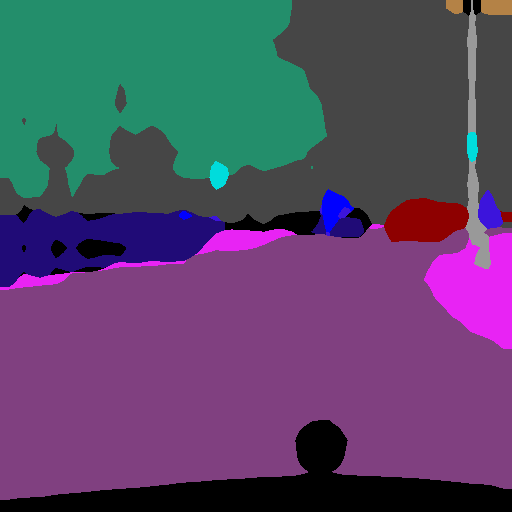}\\ 
        \includegraphics[width=0.993\textwidth,height=0.5in]{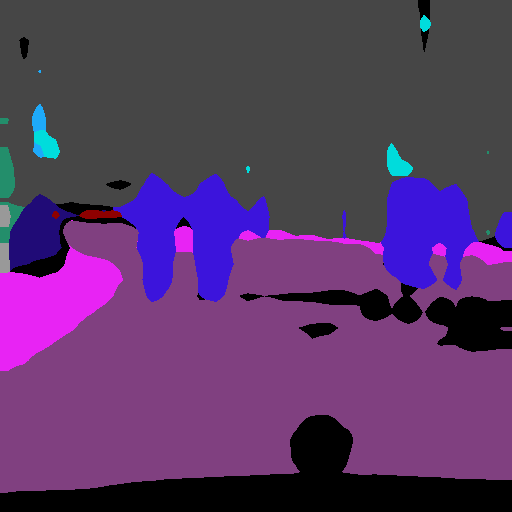}\\ 
        \includegraphics[width=0.993\textwidth,height=0.5in]{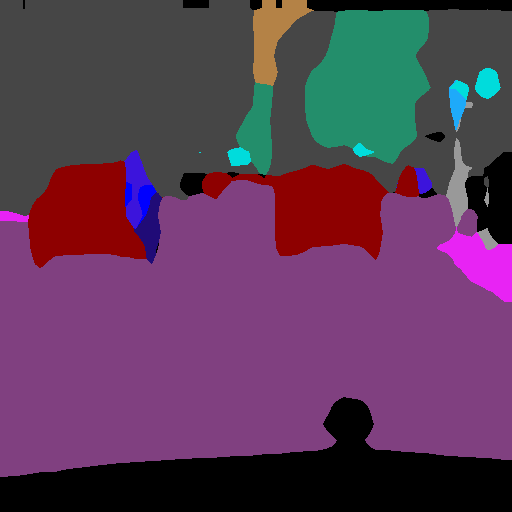}\\ 
        \vspace{2px}
    \end{minipage}%
}%
\subfigure[GT]{
    \begin{minipage}{0.14\linewidth}
        \centering
        \includegraphics[width=0.993\textwidth,height=0.5in]{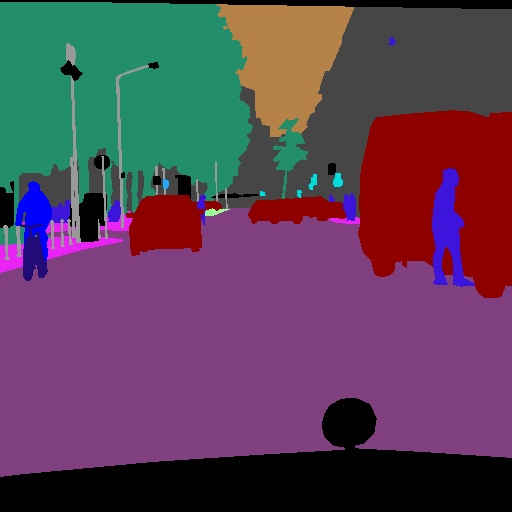}\\ 
        \includegraphics[width=0.993\textwidth,height=0.5in]{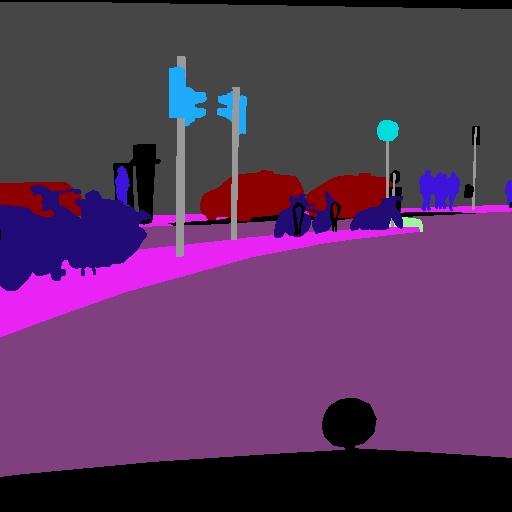}\\ 
        \includegraphics[width=0.993\textwidth,height=0.5in]{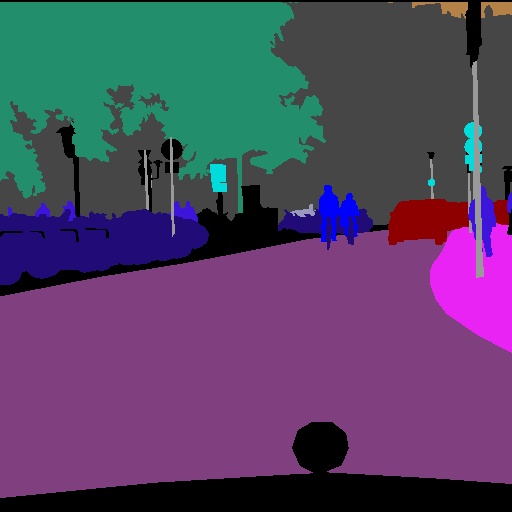}\\ 
        \includegraphics[width=0.993\textwidth,height=0.5in]{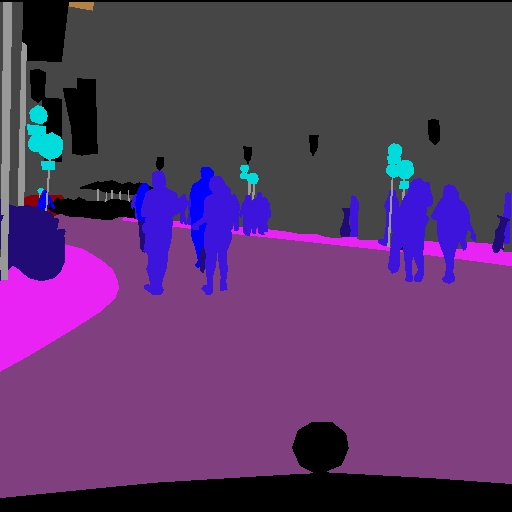}\\ 
        \includegraphics[width=0.993\textwidth,height=0.5in]{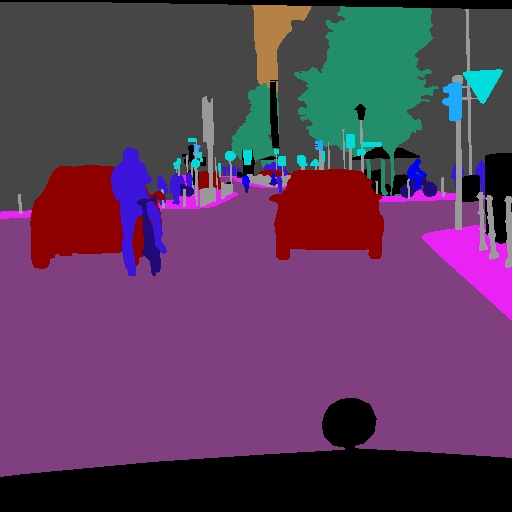}\\ 
        \vspace{2px}
    \end{minipage}%
}
\centering
\caption{Qualitative results on the Cityscapes dataset (13-6 overlapped setting).}
\label{fig:vis_city}
\end{figure*}

\noindent\textbf{Comparison on Cityscapes.}
In this work we introduce a new benchmark scenario for incremental learning for semantic segmentation based on cityscapes dataset. In the specific two settings, similar to those in previously treated datasets, have been proposed: both considering 13 classes for the old task but one testing 6 classes (\textit{i.e.}, car, truck, bus, train, motorcycle, bicycle) at once for the new task (13-6) and the other adding once class at the time (13-1) and averaging the final results over the remaining 6 classes. Cityscapes does not allow test on a \textit{disjoint} setting so we only propose the \textit{overlapped} one. 
Unlike ADE20K and VOC, the dataset shows a more coherent scenario moving the performance of base methods toward the \textit{Joint} upper bound. However, also in this ideal scenario the contribute of UDC is beneficial in almost all the settings. Some qualitative results of the (13-6) setting are presented in  Figure~\ref{fig:vis_city}. According to 2nd-4th rows in Figure~\ref{fig:vis_city}, adding UCD into MiB's and PLOP's frame works effectively solves background over-fitting caused by category forgetting.


\subsubsection{Ablation study}
\label{sec:ablation}
\begin{table}[t]
\centering
\caption{Ablation study of the proposed method on the Pascal-VOC 2012 15-5 disjoint setup and ADE 20K 100-50 disjoint setup. ``pixel contrastive'' stands for simply applying contrastive learning in the current task; ``CD'' means contrastive distillation; 
"UCD" means uncertainty-aware contrastive distillation loss.}
\label{tab:ab_pcont}
\resizebox{0.993\linewidth}{!}{%
\begin{tabular}{lccccccc}
\toprule
\multirow{2}{*}{Method} & \multicolumn{3}{c}{ADE} & \multicolumn{3}{c}{VOC} \\
\cmidrule(lr){2-4} \cmidrule(lr){5-7}
 & 1-100 & 101-150 & mIoU & 1-15 & 16-20 & mIoU \\
 \midrule
baseline & 37.5 & 27.0 & 34.0 & 65.4 & 41.4 & 59.4\\
{+ segment contrastive} & 38.1 & 27.0 & 34.4  & 67.6 & 42.6 & 61.4\\
+ pixel contrastive & 38.5 & 27.1 & 34.7 & 69.2 & 43.8 & 62.9 \\
+ CD & 39.9 & 27.7 & 35.8 & 72.6 & 43.5 & 65.3 \\
+ UCD & \textbf{40.5} & \textbf{28.1} & \textbf{36.4} & \textbf{73.0} & \textbf{46.2} & \textbf{66.3}\\
\bottomrule
\end{tabular}%
}
\end{table}
To demonstrate the contribution of each component of our model, in this section we show the results of two specific studies where these components are added one by one in UCS and evaluated accordingly.

\begin{figure}[t]
    \centering
    \includegraphics[width=0.993\linewidth]{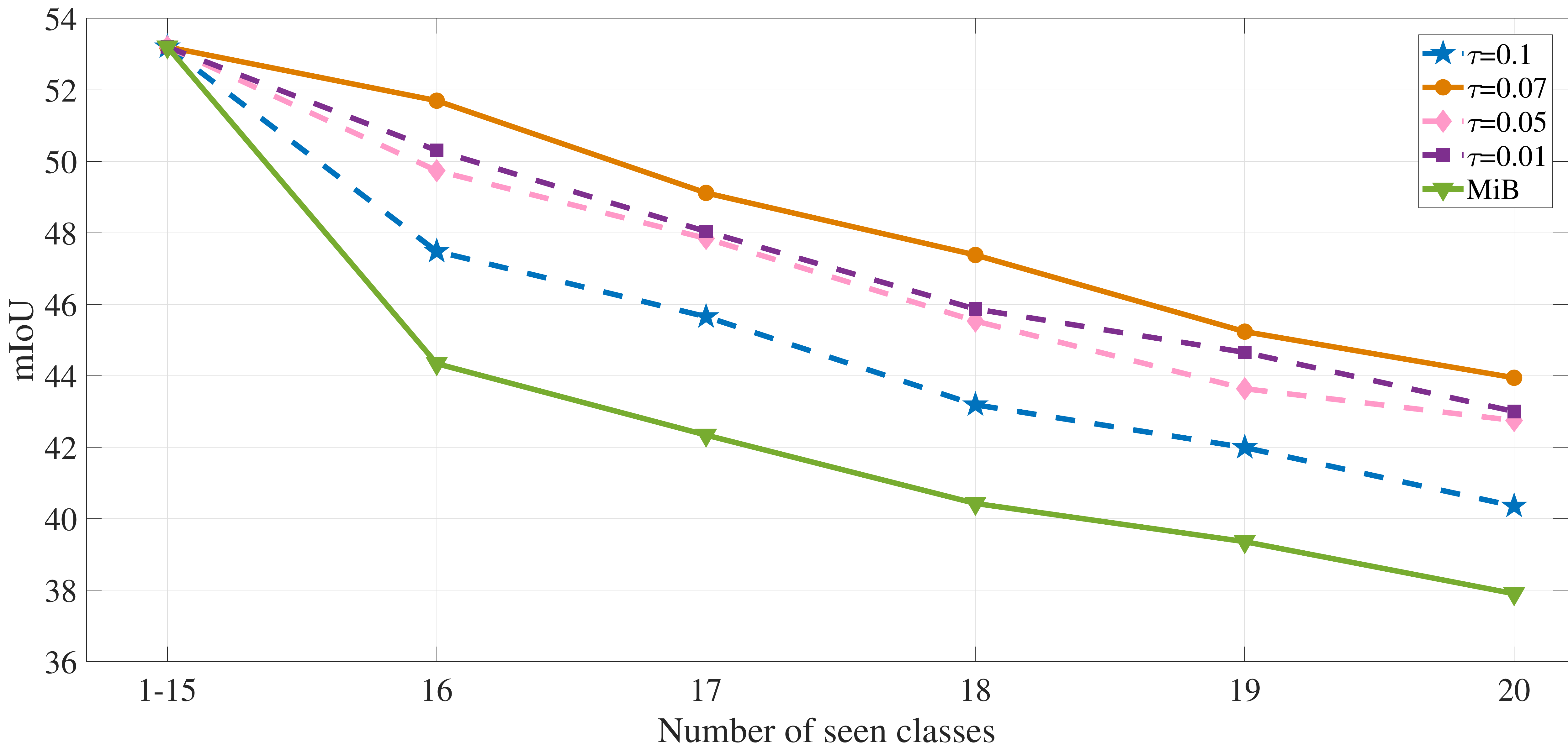}
    \caption{Mean IoU over time evolution with different temperature of contrastive loss on VOC 2012 15-1 disjoint setup. 
    }
    \label{fig:ablation cm}
\end{figure}%

\begin{table*}[t]
\centering
\caption{Ablation study of uncertainty-aware contrastive distillation loss. Old, new and bg mean the features belong to old task, new task or background. {Memory means GPU's memory consumption during training, measured by GB. Time means time consumption to complete the training, measured by minutes.} }
\label{tab:ab_ucd}
\resizebox{0.7\textwidth}{!}{%
\begin{tabular}{lccccccccccc}
\toprule
\multicolumn{3}{c}{\textbf{Anchor}} & \multicolumn{6}{c}{\textbf{Contrast}}  & \multicolumn{1}{c}{\multirow{3}[3]{*}{{\textbf{Memory}}}} & \multicolumn{1}{c}{\multirow{3}[3]{*}{{\textbf{Time}}}} & \multicolumn{1}{c}{\multirow{3}[3]{*}{\textbf{mIoU}}} \\ 
\cmidrule(lr){1-3}\cmidrule(lr){4-9}
\multicolumn{3}{c}{\textbf{New model}}  & \multicolumn{3}{c}{\textbf{New model}} & \multicolumn{3}{c}{\textbf{Old model}} & \multicolumn{1}{c}{} & \multicolumn{1}{c}{} \\
\cmidrule(lr){1-3} \cmidrule(lr){4-6} \cmidrule(lr){7-9}
\multicolumn{1}{c}{\textbf{Old}} & \multicolumn{1}{c}{\textbf{New}} & \multicolumn{1}{c}{\textbf{BG}} & \multicolumn{1}{c}{\textbf{Old}} & \multicolumn{1}{c}{\textbf{New}} & \multicolumn{1}{c}{\textbf{BG}} & \multicolumn{1}{c}{\textbf{Old}} & \multicolumn{1}{c}{\textbf{New}} & \multicolumn{1}{c}{\textbf{BG}} & \multicolumn{1}{c}{} & \multicolumn{1}{c}{}& \multicolumn{1}{c}{} \\
\midrule
$\checkmark$ & $\checkmark$ & $\checkmark$   & $\checkmark$ & $\checkmark$  & $\checkmark$ & $\checkmark$ & $\checkmark$  & $\checkmark$ & 63.4 & 147  & 66.9 \\
$\checkmark$ & $\checkmark$ & & $\checkmark$ & $\checkmark$ &  & $\checkmark$ & $\checkmark$ & & 46.5 & 108 & {66.5} \\
$\checkmark$ & $\checkmark$  & & $\checkmark$ & $\checkmark$ & &  & $\checkmark$ &  & {41.8} & 78 & 66.3 \\
\bottomrule
\end{tabular}
}
\end{table*}

\begin{table}[t]
\centering
\caption{Different temperature study of the proposed method on the Pascal-VOC 2012 15-5 disjoint setup and ADE 20K 100-50 disjoint setup. }
\label{tab:ab_t}
\resizebox{0.993\linewidth}{!}{%
\begin{tabular}{lccccccc}
\toprule
\multirow{2}{*}{Temperature} & \multicolumn{3}{c}{ADE} & \multicolumn{3}{c}{VOC} \\
\cmidrule(lr){2-4} \cmidrule(lr){5-7}
 & 1-100 & 101-150 & mIoU & 1-15 & 16-20 & mIoU \\
 \midrule
1 & 37.0 & 25.7 & 33.2 & 66.6 & 42.2 & 60.5\\
0.1 & 38.3 & 26.6 & 34.4 & 69.0 & 43.7 & 62.6\\
0.07 & \textbf{40.5} & \textbf{29.4} & \textbf{36.8} & \textbf{73.0} & \textbf{46.2} & \textbf{66.3} \\
0.05 & 40.0 & 27.7 & 35.9 & 72.0 & 45.6 & 65.4\\
0.01 & 40.0 & 27.8 & 36.0 & 72.2 & 45.7 & 65.5 \\
\bottomrule
\end{tabular}%
}
\end{table}

\begin{table}[t]
\centering
\caption{Ablation study of $\lambda_{ucd}$ on the Pascal-VOC 2012 15-5 and 19-1 disjoint setup . }
\label{tab:ab_lambda}
\resizebox{0.993\linewidth}{!}{%
\begin{tabular}{lccccccc}
\toprule
\multirow{2}{*}{$\lambda_{ucd}$} & \multicolumn{3}{c}{15-5} & \multicolumn{3}{c}{19-1} \\
\cmidrule(lr){2-4} \cmidrule(lr){5-7}
 & 1-15 & 16-20 & mIoU & 1-19 & 20 & mIoU \\
 \midrule
0 & 65.5 & 41.4 & 59.4 & 66.8 & 15.2 & 64.2\\
0.1 & 72.6 & 43.4 & 65.3 & 73.5 & 28.1 & 71.3\\
0.01 & \textbf{73.0} & \textbf{46.2} & \textbf{66.3} & \textbf{73.9} & \textbf{28.2} & \textbf{71.7} \\
0.001 & 71.7 & 42.9 & 64.5 & 73.3 & 28.0 & 71.0\\
0.0001 & 68.4 & 42.0 & 61.7 & 69.7 & 21.5 & 67.3 \\
\bottomrule
\end{tabular}%
}
\end{table}

Table~\ref{tab:ab_pcont} shows the results of our ablation on different contrastive losses. We start from a baseline model that includes only knowledge distillation at class-probability level and cross-entropy loss as in \cite{cermelli2020modeling}. In the subsequent rows, {several additional components are added on top of the baseline gradually to investigate the performance effect of each component}. 
{Specifically, as shown in the table, the pixel contrastive refers to a supervised contrastive loss on the new task features, $f^k$. Differently, the segment contrastive indicates the strategy using a contrastive loss applied by averaging the input features for each pixel within the same class. 
The CD and UCD components indicate the plugging of the proposed contrastive distillation and  uncertainty-aware contrastive distillation, respectively.
Specifically, the full contrastive distillation loss, $\mathcal{L}_{cd}$, is used for the component of {CD}. Finally, the contrastive distillation loss with uncertainty is used for the component of UCD.}
{As can be observed from Table~\ref{tab:ab_pcont}}, our results clearly demonstrate that the pixel contrastive loss outperforms the segment contrastive loss on both datasets. This {motivates} our choice {of designing} a contrastive distillation loss in UCD based on pixel contrastive loss. {In} Table~\ref{tab:ab_pcont}, CD outperforms the standard pixel-wise contrastive loss.
Moreover, the results also show that {the proposed} contrastive distillation loss {with uncertainty estimation and the mitigation of catastrophic forgetting} significantly performs better than the CD, {with an improvement on mIoU by 0.5\% on ADE 20K 100-50 and by 1.0\% on Pascal-VOC 2012 15-5, respectively. All these ablation studies concretely verifies the effectiveness of the proposed approach.} 

{After ablation on different contrastive losses, we further make ablation study on component of uncertainty-aware contrastive distillation loss in Tab.~\ref{tab:ab_ucd}. In detail, we run experiments with and without background class in the contrastive loss. According to results in Tab.~\ref{tab:ab_ucd}, We find the performance without old class to be comparable. However, without background , the number of pixels used in the contrastive loss is much smaller, resulting in GPU's memory consumption dropping significantly. Therefore we concluded that it is better to remove the background in the contrastive distillation loss. For the same reason, we also remove the old task feature of old model features in our method.}

\begin{figure}[t]
    \centering
    \includegraphics[width=0.993\linewidth]{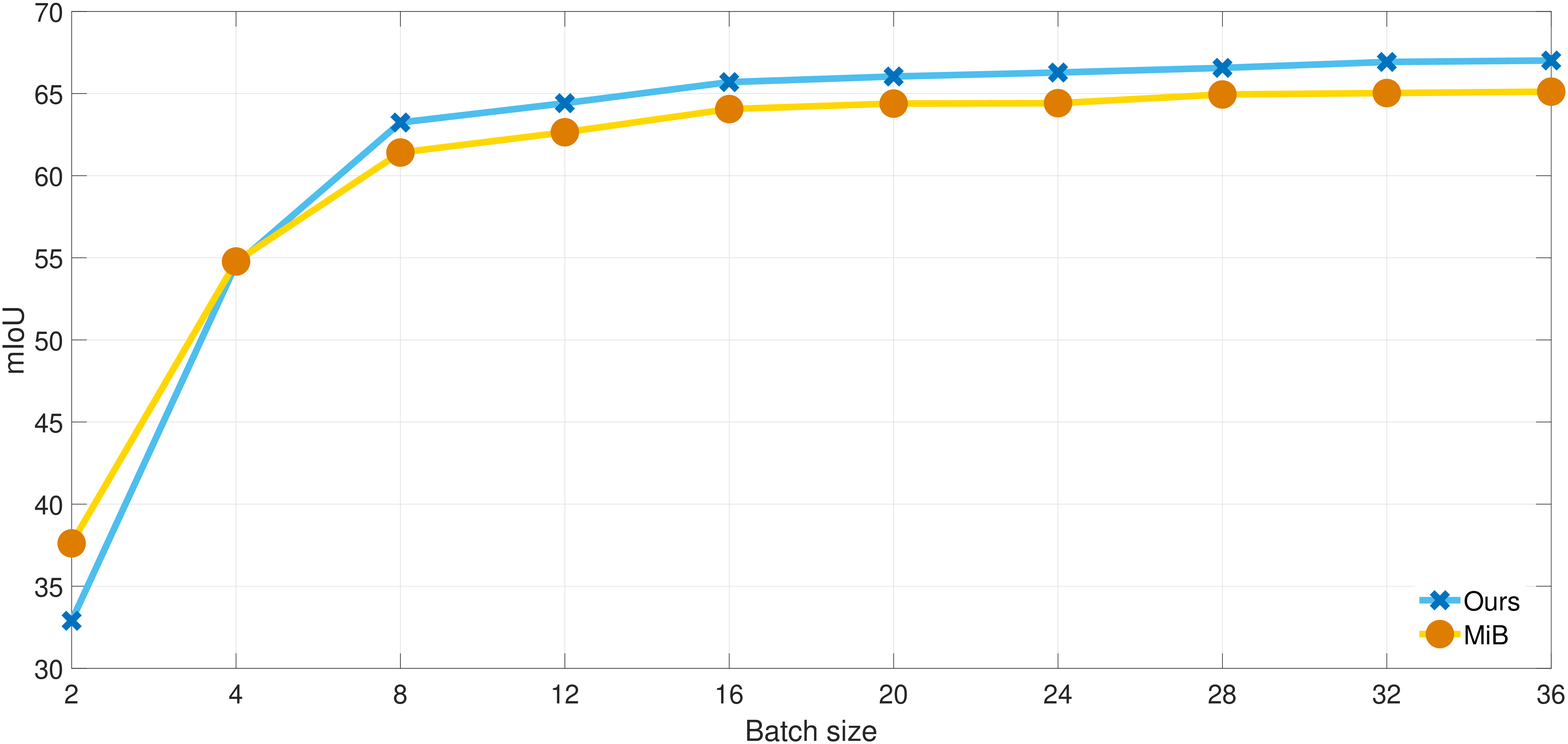}
    \caption{{Mean IoU over time evolution with different batch size on VOC 2012 15-5 disjoint setup.}}
    \label{fig:ablation bs}
\end{figure}

We also provide a sensitivity study. The {evaluation of the} performance effect of the temperature value $\tau$ {and} the number of learned classes is shown in~{Table~\ref{tab:ab_t} and Figure.~\ref{fig:ablation cm}, respectively}. It clearly {shows the importance of using a proper temperature value. Varying the value leads to certain variance in the final performance.} Nevertheless, {the performance decreasing} is consistent with the increasing of the number of seen classes.
Lastly, Fig.~\ref{fig:ablation bs} shows the mIoU performance with the increase of the batch size on Pascal-VOC 2012 15-5 disjoint setup. This study confirms what found in previous other tasks~\cite{chen2020simple,he2020momentum}, that is contrastive learning benefits the network optimization with larger batch sizes, {except for very small batch size such as 2 and 4}. {We could also observe from the figure, our method steadily outperforms the main competitor (i.e. MiB) at different batch-size values. Therefore, the effectiveness of the proposed approach is independent of the batch size setup.} {Meanwhile, performance does not drastically improve with larger batch size while increasing the batch size also increases the memory usage. For fair comparison with other methods like MiB and PLOP, the same batch size (24) has been chosen for all the compared methods.}

